\documentclass[Afour,sageh,times]{sagej}

\usepackage{moreverb,url}

\usepackage[colorlinks,bookmarksopen,bookmarksnumbered,citecolor=red,urlcolor=red]{hyperref}

\usepackage{algorithm}
\usepackage{algpseudocode}
\usepackage[title]{appendix}

\newcommand\BibTeX{{\rmfamily B\kern-.05em \textsc{i\kern-.025em b}\kern-.08em
T\kern-.1667em\lower.7ex\hbox{E}\kern-.125emX}}

\begin{document}

\runninghead{Lloyd and Lepora}

\title{A pose and shear-based tactile robotic system for object tracking, surface following and object pushing}

\author{John Lloyd\affilnum{1} and Nathan Lepora\affilnum{2}}

\affiliation{\affilnum{1}University of Bristol and Bristol Robotics Laboratory, UK\\
\affilnum{2}University of Bristol and Bristol Robotics Laboratory, UK}

\corrauth{John Lloyd, Department of Engineering Mathematics
Faculty of Engineering, University of Bristol and Bristol Robotics Laboratory,
Bristol,
BS34~8QZ, UK.}

\email{jl15313@my.bristol.ac.uk}

\begin{abstract}

Tactile perception is a crucial sensing modality in robotics, particularly in scenarios that require precise manipulation and safe interaction with other objects. Previous research in this area has focused extensively on tactile perception of contact poses as this is an important capability needed for tasks such as traversing an object's surface or edge, manipulating an object, or pushing an object along a predetermined path. Another important capability needed for tasks such as object tracking and manipulation is estimation of post-contact shear but this has received much less attention. Indeed, post-contact shear has often been considered a "nuisance variable" and is removed if possible because it can have an adverse effect on other types of tactile perception such as contact pose estimation. This paper proposes a tactile robotic system that can simultaneously estimate both the contact pose and post-contact shear, and use this information to control its interaction with other objects. Moreover, our new system is capable of interacting with other objects in a smooth and continuous manner, unlike the stepwise, position-controlled systems we have used in the past. We demonstrate the capabilities of our new system using several different controller configurations, on tasks including object tracking, surface following, single-arm object pushing, and dual-arm object pushing.

\end{abstract}


\maketitle

\section{Introduction}
\label{sec:introduction}

In robotics, tactile sensing allows robots to perceive and manipulate objects in a way that is similar to human touch, enabling them to perform tasks that would be difficult or impossible with other sensing modalities. This is crucial in scenarios that require precise manipulation and safe interaction with objects, such as in manufacturing, assembly, and healthcare applications (\cite{luo2017robotic, li2020review}). One of the main advantages of tactile sensing is that it provides robots with a more comprehensive understanding of the objects they are interacting with. Vision-based sensing systems, for example, may be limited by lighting conditions or the complexity of the object being perceived. Tactile sensing, on the other hand, allows robots to perceive details such as texture, shape, and hardness, which are not always visible to the human eye. Tactile perception is also important in scenarios where robots need to interact with delicate or fragile objects, such as in surgical applications where they must be able to manipulate delicate tissues without causing damage.

Tactile sensing has become increasingly important in the field of robotics, particularly in the technique known as \emph{tactile servoing} (\cite{li2013control, lepora2021pose}). This technique involves using tactile feedback to control and adjust the position, velocity, or force of a robot's end-effector, such as a gripper or tool. Tactile servoing is an important technique in robotics because it allows robots to manipulate objects with greater precision and accuracy and adapt to changes in their environment in real time. It also allows robots to detect and respond to obstacles or hazards during their operation.

For example, in \emph{surface following} tasks, where a robot uses its fingers, hand or end-effector to traverse the surface of an object to gather information about its texture, shape or size, tactile servoing can be used to adjust the robot's position, velocity or applied force in order to maintain appropriate contact with the surface.

Similarly, in \emph{object pushing} tasks, if the object translates or rotates while being pushed, the robot can use tactile servoing to detect this motion and maintain appropriate contact with the object. The ability to push objects over surfaces is important in several industrial and manufacturing robot applications as it allows robots to perform tasks such as object repositioning, sorting, and assembly. Robots that are capable of pushing objects can work in environments that are difficult or dangerous for humans, reducing the risk of injury to human workers. Moreover, robots with this sort of capability can manipulate delicate or fragile objects without causing damage or deformation, making them useful in applications such as electronics assembly or food packaging.

A third type of task where tactile servoing is important is \emph{object tracking}, where a robot needs to maintain a predefined contact pose with respect to a static or moving object. In this scenario, the robot can use tactile servoing to detect the motion of the object and maintain the required contact pose. In many ways, this task is similar to the surface following task, but here the object surface is typically moving and the goal is to maintain a given contact pose with respect to the moving surface rather than to move across a fixed surface. This type of operation is important in collaborative environments (robot-robot or robot-human) where a robot may need to respond to the motions or forces transmitted to it via objects being manipulated by a collaborator.

In our previous research on surface following and object pushing, we used convolutional neural networks (CNNs) to estimate the contact pose between an object surface and an optical tactile sensor mounted as an end-effector on a robot arm (\cite{lepora2020optimal, lepora2021pose, lloyd2021goal}). Since our main objective was not to estimate the post-contact shear experienced by the sensor, we assumed that the corresponding pose components were zero and trained the CNN to be invariant to this type of motion using a form of domain randomization. Another recent study from our group took a slightly different approach, which involved disentangling the contact pose and post-contact shear in the latent feature space of the CNN model during the training process (\cite{pmlr-v164-gupta22a}). However, in both of these approaches shear was treated as an undesirable "nuisance variable" that could interfere with the primary goal of contact pose estimation, rather than as a useful attribute that could aid in servoing and manipulation tasks. However, in many applications, these shear effects can be critically important. For example, when tracking a moving object (regardless of whether the object is being moved by the tracking system or by a separate system), it is essential that a robot can detect and respond to both shear and normal forces applied to the tactile sensor. A robot may also need to detect and respond to shear forces when sliding over a surface to measure its roughness or texture, or to identify anomalies such as bumps, holes, or other defects. The ability to detect shear is also useful in preventing shear-related damage to soft tactile sensors.

In this paper, we present an extension of our surface pose estimation model (see \cite{lepora2021pose}) to include post-contact shear effects. We use this model to develop a tactile robotic system that can be programmed for a diverse range of non-prehensile manipulation tasks, such as object tracking, surface following, single-arm object pushing, and dual-arm (i.e., stabilized) object pushing. While we have already demonstrated surface following and single-arm pushing in our previous work, our new system is capable of performing these tasks with a continuous smooth motion instead of using discrete, position-controlled movements. Completely new tasks, such as object tracking and dual-arm pushing, are enabled solely by the capability to anticipate shear motion.

In developing our new system, we needed to address two main challenges. The first challenge was related to the \emph{tactile aliasing} that occurs when trying to estimate shear between an object's surface and a soft tactile sensor in the presence of slippage. During data collection, a particular contact motion that induces slip between the sensor and surface can produce sensor images that closely resemble those obtained for a different contact motion. Consequently, very similar sensor images may become associated with very different pose labels, leading to an increase in the error and uncertainty in the trained model's contact pose estimates (\cite{lloyd-RSS-21}). Here, the type of uncertainty we are referring to is \emph{aleatoric uncertainty} or \emph{data uncertainty}, which is due to the variability or noise in training or test data labels. The other main type of uncertainty is known as \emph{epistemic uncertainty} or \emph{model uncertainty} and is due to insufficient training data or gaps in the data. This second type of uncertainty is less of a concern in our situation since we can ensure that data is collected sufficiently densely and uniformly over the required domain.

The second challenge centred on the requirement to produce a smooth continuous robot motion while interacting with an object or surface. In our prior work, we used discrete, position-controlled robot movements to traverse surfaces or contour features or to push objects across surfaces. However, if robots are to track objects that move smoothly, or measure surface properties such as roughness or texture in a manner similar to humans, they will need to perform these operations in a smooth continuous fashion.

\begin{figure}
	\centering
	\includegraphics[width=\columnwidth]{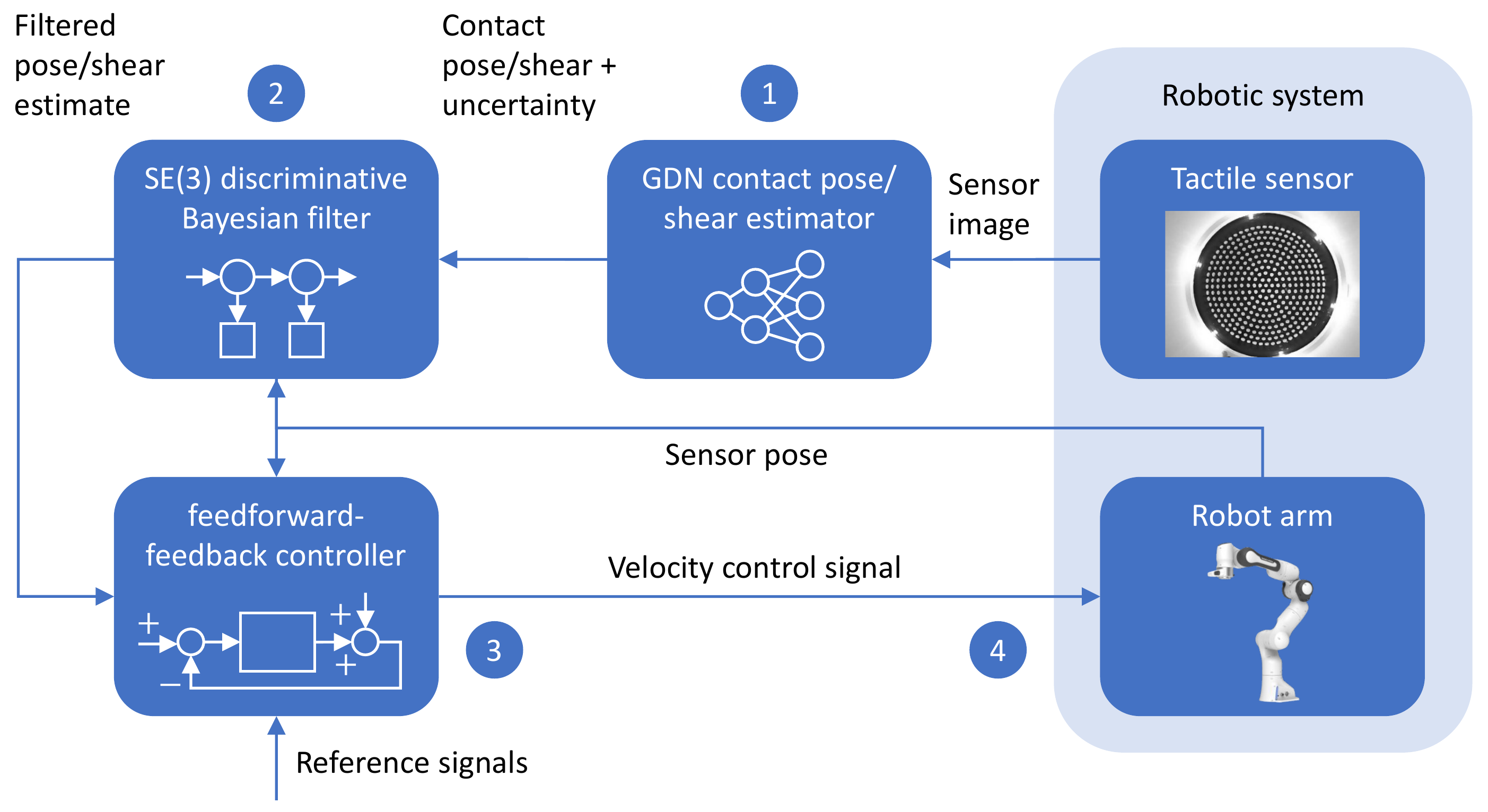}
	\caption{A pose and shear-based tactile robotic system for controlling a robot arm. The main system components are: a Gaussian density network (GDN) model for estimating the contact pose/shear and corresponding uncertainty, based on a tactile sensor image (1); an $SE(3)$ discriminative Bayesian filter for reducing the error and uncertainty of pose/shear estimates (2); a feedforward-feedback controller for controlling the robot arm (3); and a robot arm fitted with a tactile sensor as an end-effector (4).
	\label{fig:system_architecture}}
\end{figure}

In this paper, we propose a novel tactile robotic system that uses contact pose and post-contact shear information to execute a variety of non-prehensile manipulation and tactile servoing tasks while addressing the two key challenges outlined above. The system comprises four main components, including a Gaussian density network (GDN) for pose and shear estimation, a discriminative Bayesian filter that operates in the $SE(3)$ space, a feedforward-feedback controller, and a robot arm equipped with an optical tactile sensor (Figure~\ref{fig:system_architecture}). Although the pose estimator and controller may appear superficially similar to those we have used in the past, our new system incorporates several innovative extensions for predicting uncertainty in pose and shear estimates and for enabling smooth, continuous robot motion. We have also included a novel Bayesian filter component for decreasing the error and uncertainty associated with contact pose estimates. The rationale, derivation, and design of each of these components is described in the following sections.

This paper makes several capability-related and technical contributions to research in this area. In terms of tactile robot capability, we have:
\begin{itemize}
	\item Shown how to combine post-contact shear with contact pose information to extend the pose-based tactile servoing (PBTS) framework used in our previous work (\cite{lepora2021pose}).
	\item Demonstrated how this extended framework can be used to perform several object tracking and dual-arm object pushing tasks, which to the best of our knowledge have not been demonstrated before (and certainly not with optical tactile sensors).
	\item Extended two existing tactile robotic tasks, 3D surface following and single-arm object pushing, so that they can be carried out using smooth continuous motion instead of the discrete point-to-point motion we used in our previous work. This is likely to be important for robotic equivalents of the tactile exploratory procedures that humans use to investigate object properties.
\end{itemize}
In terms of technical contributions, we have:
\begin{itemize}
	\item Shown how multi-output regression convolutional neural networks (CNNs) can be adapted to estimate the uncertainty associated with predictions, and in particular for cases such as ours where some of the output variables are much noisier than others.
	\item Developed a novel Bayesian filter for combining successive pose/shear estimates to increase the predictive accuracy and reduce the corresponding uncertainty over a sequence of observations.
	\item Re-framed many of our existing tactile servoing methods in terms of Lie group theory, so that we can leverage established tools and techniques from probability and control theory and thereby achieve a wider range of capabilities.
\end{itemize}

\section{Background and related work}
\label{sec:background}

\subsection{Tactile pose and shear estimation}
\label{sec:tactile_pose_shear_estimation}

Contemporary methods for tactile pose estimation can be broadly categorised according to whether they estimate a \emph{local contact pose} or a \emph{global object pose}. The local contact pose estimation problem is somewhat easier to solve because it only depends on tactile information provided in a single contact, although accuracy can often be improved by using a sequence of observations or contacts. As such, it has been studied over a longer period of time than the second problem. The global object pose estimation problem is generally harder to solve because it involves fusing information from several different contacts, and uses this information together with some form of object model to estimate the object pose (\cite{bimbo2015global, suresh2021tactile, villalonga2021tactile, bauza2022tac2pose, kelestemur2022tactile, caddeo2023collision}). As such, work on this type of pose estimation problem has only recently gained momentum and it has largely been driven by recent progress in deep learning models. Comprehensive reviews of pose estimation in the context of robotic tactile perception can be found in \cite{luo2017robotic} and \cite{li2020review}.

In this paper, we focus on local contact pose and post-contact shear estimation using optical tactile sensors, although in principle there is no reason why our methods could not be adapted for other types of tactile sensor, assuming that the neural network pose estimator was modified accordingly (e.g., it would not use a convolutional base unless it processed images). Bicchi et al. proposed a theoretical model for estimating pose and shear information, and described a framework for designing tactile sensors that have this capability (\cite{bicchi1993contact}). In particular, their theoretical model addressed the problem of how to determine the location of a contact, the force at the interface and the moment about the contact normals.

In the context of optical tactile sensors, Yuan et al. showed that the GelSight sensor can be used to estimate the normal contact pose between the sensor and an object surface, but is somewhat limited in its contact angle range due to its rather flat sensor geometry (\cite{yuan2017gelsight}). Similarly, Lepora et al. showed that the TacTip soft optical tactile sensor could be used to predict 2D contact poses (\cite{lepora2019pixels}) and, more recently, 3D contact poses (\cite{lepora2021pose}).

The problem of estimating post-contact shear is somewhat less well-explored than contact pose estimation. Yuan et al. showed how a GelSight sensor can be used to measure translational and rotational post-contact shear by including embedded markers in the sensing surface (\cite{yuan2015measurement}). Cramphorn et al. also described a similar approach based on marker shear that can be used with the TacTip sensor (\cite{cramphorn2018voronoi}). More recently, Gupta described a deep learning approach for disentangling contact pose and post-contact shear representations in the latent feature space of a CNN model (\cite{pmlr-v164-gupta22a}). However, in that work, the primary objective was to find a data-efficient way of decoupling the "nuisance effect" of shear from the primary goal of contact pose estimation, rather than to use the shear information for any specific task.

\subsection{Tactile servoing and object pushing}
\label{sec:tactile_servoing_pushing}

Methods for robotic tactile servoing can be grouped according to whether they control attributes in the \emph{signal space} or \emph{feature space} of tactile sensor signals/features, or attributes in the \emph{task space} associated with the problem at hand. For optical tactile sensors or tactile sensors that produce a taxel "image", if control is performed in the sensor feature space it is sometimes referred to as \emph{image-based tactile servoing} (IBTS). Conversely, if control is performed in the task space and the task involves tracking a reference pose with respect to a surface feature it is sometimes referred to as \emph{pose-based tactile servoing} (PBTS) (\cite{lepora2021pose}). In principle, there is no reason why a hybrid approach could not be used, where some aspects of control are performed in the task space and some are performed in the signal or feature space. The tactile servo control methods used in this paper can be viewed as pose-based tactile servoing methods (and more generally as task-space methods) because of the way we combine the contact pose and shear motion into a single "surface contact pose", and use it in a feedback loop to control the motion of the robot arm.

Berger and Khoslar used image-based tactile feedback on the location and orientation of edges together with a feedback controller to track straight and curved edges in 2D (\cite{berger1991using}). Chen et al. used a task-space tactile servoing approach, using an "inverse tactile model" similar in concept to a pose-based tactile servoing model, to follow straight-line and curved edges in 2D (\cite{chen1995edge}). Zhang and Chen used an image-based tactile servoing approach and introduced the concept of a "tactile Jacobian" to map image feature errors to task space errors (\cite{zhang2000control}). They used their system to track straight and curved edges in 2D and to follow cylindrical and spherical surfaces in 3D. Sikka et al. drew inspiration from image-based visual servoing to develop a tactile analogy using the taxel "images" produced by a tactile sensor to control the movement of a robot arm. They applied their tactile servoing system to the task of rolling a cylindrical pin on a planar surface (\cite{sikka2005tactile}).

Later on, Li et al. used a similar tactile servoing approach to Zhang and Sikka to demonstrate a wider selection of servoing tasks including 3D object tracking and surface following (\cite{li2013control}). Lepora et al. used a soft optical tactile sensor with a bio-inspired active touch perception method and a simple proportional controller to demonstrate contour following around several complex 2D edges and ridges (\cite{lepora2017exploratory}). Sutanto et al. used a learning-from-demonstration (LFD) framework for learning a tactile servoing dynamics model and used it to demonstrate a contact point tracking task in 3D (\cite{sutanto2019learning}). Kappassov et al. developed a task-space tactile servoing system, similar to the earlier system developed by Chen and used it for 3D edge following and object co-manipulation (\cite{kappassov2020touch}). More recently, Lepora and Lloyd described a pose-based tactile servoing approach that uses a deep learning model to map from the tactile image space to the task space (in this case the task space is defined in terms of the contact poses between the tactile sensor and object surface features) (\cite{lepora2021pose}). They used this approach to demonstrate robotic surface and edge following on complex 2D and 3D surfaces.

Most current approaches for robotic object pushing also fall into two main categories: \emph{analytical physics-based approaches}, which are used in conventional robot planning and control systems, and \emph{data-driven approaches} for learning forward or inverse models of pusher-object interactions, or for directly learning control policies (e.g., using reinforcement learning). We summarise work on these two approaches in the following paragraphs. A more comprehensive survey on robotic object pushing can be found in \cite{stuber2020let}.

In the case of analytical, physics-based object pushing, Mason derived a simple rule known as the \emph{voting theorem} for determining the direction of rotation of a pushed object (\cite{mason1986mechanics}). Goyal et al. introduced the concept of a \emph{limit surface} to describe how the sliding motion of a pushed object depends on its frictional properties (\cite{goyal1989limit}). Lee and Cutkosky derived an ellipsoid approximation to the limit surface, with the aim of reducing the computational overhead of applying it in real-world applications (\cite{lee1991fixture}). Lynch et al. used the ellipsoid approximation to obtain closed-form analytical solutions for sticking and sliding pushing interactions (\cite{lynch1992manipulation}). Howe and Cutkosky explored other non-ellipsoidal geometric forms of limit surface and provided guidelines for selecting between them (\cite{howe1996practical}). Lynch and Mason analysed the mechanics, controllability and planning of object pushing and developed a planner for finding stable pushing paths between obstacles (\cite{lynch1996stable}).

In the case of data-driven approaches, Kopicki et al. used a modular data-driven approach for predicting the motion of pushed objects (\cite{kopicki2011learning}). Bauza et al. developed models that describe how an object moves in response to being pushed in different ways, and embedded these models in a model-predictive control (MPC) system (\cite{bauza2018data}). Zhou et al. developed a hybrid analytical/data-driven approach that approximated the limit surface for different objects using a parametrised model (\cite{zhou2018convex}). Other researchers have used deep learning to model the forward or inverse dynamics of pushed object motion (\cite{agrawal2016learning, byravan2017se3, li2018push}), or to learn end-to-end control policies for pushing (\cite{clavera2017policy, dengler2022learning}). In general, analytical approaches are more computationally efficient and transparent in their operation than data-driven approaches, but they may not perform well if their underlying assumptions and approximations do not hold in practice (\cite{yu2016more}).

While most object pushing methods rely on computer vision systems to track the pose and other state information of the pushed object, a few (including ours) use tactile sensors to perform this function. Lynch et al. were the first to employ tactile sensing in this way, to manipulate a rectangular object and circular disk on a moving conveyor belt (\cite{lynch1992manipulation}). Jia and Erdmann used a theoretical analysis to show that the pose and motion of a planar object with known geometry can be determined using only the tactile contact information generated during pushing (\cite{jia1999pose}). More recently, Meier et al. used a tactile-based method for pushing an object using frictional contact with its upper surface (\cite{meier2016distinguishing}).

From a control perspective, the most similar approaches to our method for single-arm robotic pushing are the ones described by Hermans (\cite{hermans2013decoupling}) and Krivic (\cite{krivic2019pushing}). The similarities and differences are described in more detail in \cite{lloyd2021goal} but the main difference is that they both used computer vision techniques to track the state of the pushed object, rather than the tactile sensing and proprioceptive feedback that we use.

\section{Methods}

\subsection{Notation and mathematical preliminaries}
\label{sec:math_prelim}

In developing and describing our tactile robotic system, we rely on some important concepts and notation from Lie group theory. In many ways, this is a natural framework to use when dealing with velocity-based control systems that produce smooth continuous motions of 3D poses because it provides the relevant abstractions for this sort of motion. However, it turns out that this framework is also useful for defining probability distributions over 3D poses, which is important for modelling uncertainty in pose estimates. In this section, we introduce some definitions and properties of matrix Lie groups and algebras, with a particular focus on the Special Euclidean Group of rotations and translations in 3D, denoted $SE(3)$. We also describe how we represent probability distributions in $SE(3)$. A more comprehensive introduction to Lie group theory can be found in \cite{stillwell2008naive} and its application to robotics in \cite{barfoot2017state} and \cite{sola2018micro}.

\begin{figure*}
	\centering
	\includegraphics[width=\textwidth]{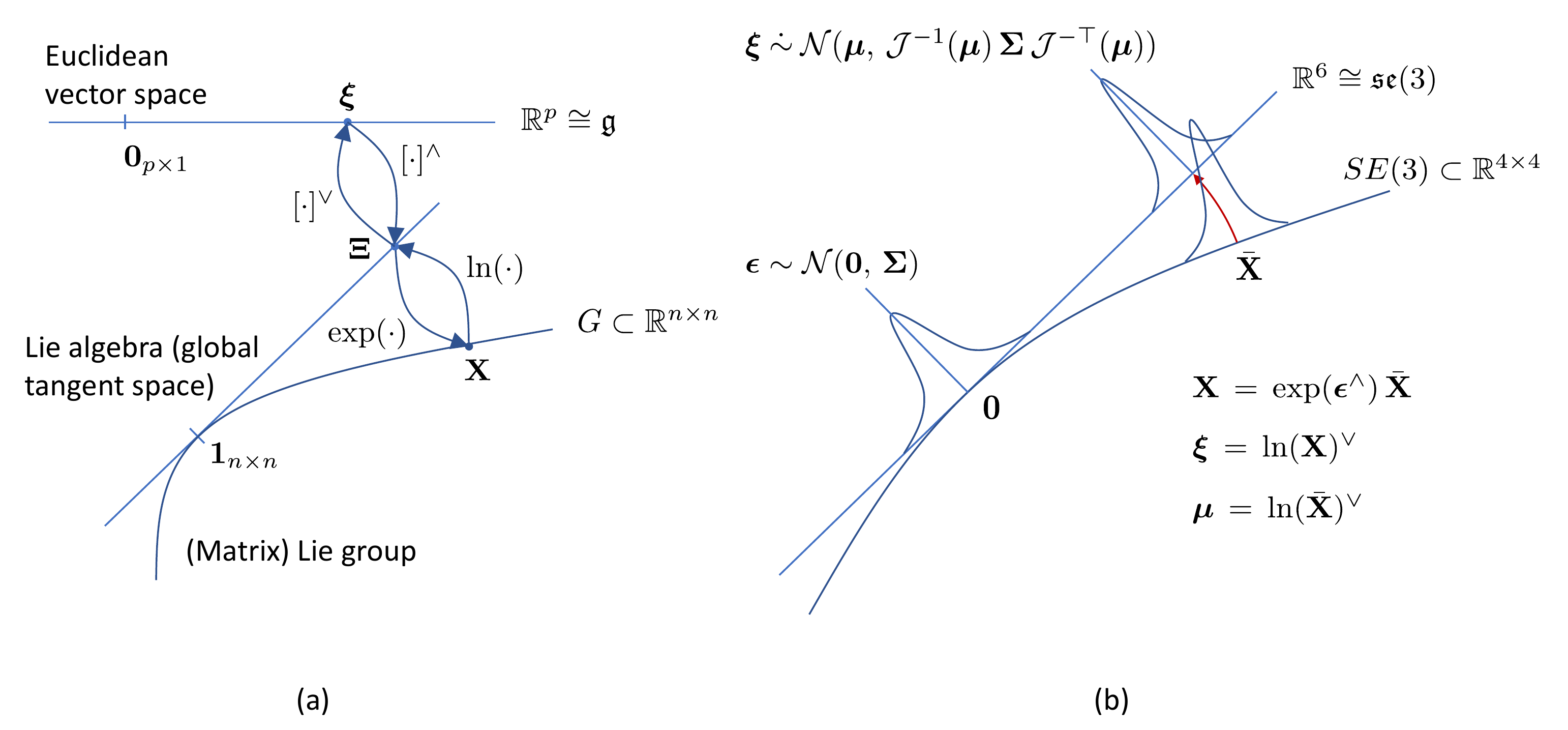}
	\caption{Matrix Lie group notation and probability distributions. (a) Matrix Lie group and Lie algebra notation, showing exponential and logarithmic maps and related "hat" and "vee" operators. (b) Perturbation-induced probability distributions in $SE(3)$ and their approximation in the global tangent space (Lie algebra) $\mathfrak{se}(3)$.
	\label{fig:matrix_lie_groups}}
\end{figure*}

A \emph{Lie group} is a group that is also a smooth, differentiable manifold (i.e., a group with some notion of distance, which is locally similar enough to a Euclidean vector space to allow one to apply calculus). Hence, the group composition and inversion operations are smooth, differentiable operations. A \emph{matrix Lie group}, $G$, is a smooth manifold in the set of $\mathbb{R}^{n \times n}$ matrices, which is closed under composition and where the composition and inversion operations are matrix multiplication and inversion, respectively. The group identity is the $n \times n$ identity matrix $\boldsymbol{\mathrm{1}}_{n \times n}$. Examples of matrix Lie groups include the Special Orthogonal Group of rotations in two or three dimensions (i.e., the $2 \times 2$ and $3 \times 3$ rotation matrices), denoted $SO(2)$ and $SO(3)$, respectively. In this paper, we focus on the Special Euclidean Group of rotations and translations in 3D, $SE(3) \subset \mathbb{R}^{4 \times 4}$ because it is commonly used to represent poses, pose transformations or changes of coordinate frame in 3D. The objects in this group are defined as the following set of $4 \times 4$ matrices:
\begin{equation}
	\label{eqn:se3_matrix_def}
	SE(3)
	\, = \,
	\{ \boldsymbol{\mathrm{X}}
	=
	\begin{bmatrix}
		\boldsymbol{\mathrm{C}} & \boldsymbol{\mathrm{r}}\\
		\boldsymbol{\mathrm{0}}^{\top} & 1
	\end{bmatrix}
	, \, \boldsymbol{\mathrm{C}} \in SO(3)
	, \, \boldsymbol{\mathrm{r}} \in \mathbb{R}^{3} \}
\end{equation}

Here, the $3 \times 3$ orthonormal matrix, $\boldsymbol{\mathrm{C}}$, represents the rotational component of the transformation, and the 3-element column vector, $\boldsymbol{\mathrm{r}}$, represents the translational component.

For any point on the smooth manifold that these matrix objects lie on, it is possible to construct a unique \emph{tangent space}, which is a $p$-dimensional vector space that is tangent to the surface at that point, whose basis, or \emph{generators}, consists of a set of $p$, $n \times n$ matrices. We sometimes refer to this as the \emph{local tangent space} at a point, to distinguish it from the \emph{global tangent space} at the group identity, which has a special significance. Since the tangent space is a vector space, its elements can be uniquely identified with elements of $\mathbb{R}^{p}$, which represent coordinates with respect to the basis. In other words the tangent space is isomorphic to the Euclidean vector space $\mathbb{R}^{p}$. For $SE(3)$, the tangent space is 6-dimensional because there are three translational and three rotational degrees of freedom (there are 6 constraints on the 9-element rotation matrix component of an $SE(3)$ object), hence in this case $p=6$. For a matrix group, $G$, the tangent space defined at the identity element in $G$ is called the \emph{Lie algebra}, denoted $\mathfrak{g}$. We denote the Lie algebra of $SE(3)$ as $\mathfrak{se}(3)$, which is isomorphic to $\mathbb{R}^{6}$, i.e., $\mathfrak{se}(3) \cong \mathbb{R}^{6}$. For the sake of brevity, in the remainder of this paper when we refer to a particular tangent space or Lie algebra of $SE(3)$ we are usually referring to the corresponding (isomorphic) Euclidean vector space, $\mathbb{R}^{6}$. This should be clear from the context.

Elements, $\boldsymbol\Xi$, of the Lie algebra, $\mathfrak{g}$, are mapped onto corresponding elements, $\boldsymbol{\mathrm{X}}$, of the Lie group, $G$, via the \emph{exponential map}:

\begin{equation}
	\label{eqn:exponential_map}
	\boldsymbol{\mathrm{X}}
	\, = \,
	\exp(\boldsymbol\Xi)	
	\, = \,
	\boldsymbol{\mathrm{1}} + \boldsymbol\Xi + \frac{\boldsymbol\Xi^{2}}{2!}
	+\frac{\boldsymbol\Xi^{3}}{3!} + \ldots
\end{equation}
where the successive powers of $\boldsymbol\Xi$ are defined recursively in terms of the group composition operation (matrix multiplication in the case of $SE(3)$). In $SE(3)$, the exponential map is surjective (i.e., every element of $SE(3)$ can be generated by many elements of $\mathfrak{se}(3)$) and it can be evaluated in closed form by grouping odd and even powers of $\boldsymbol\Xi$ in the series, and recognising that these groups can then be replaced by terms involving trigonometric functions, similar to the way that Rodrigues' formula is derived for the exponential map for $\mathfrak{so}(3)$ (see \cite{lynch2017modern}, for example).

Moving in the opposite direction, elements of the matrix group, $G$, are mapped into the Lie algebra, $\mathfrak{g}$, using the \emph{logarithmic map}, $\boldsymbol\Xi \, = \, \ln(\boldsymbol{\mathrm{X}})$, which like the exponential map, is defined in terms of an infinite series that is analogous to the corresponding scalar series. For $SE(3)$, the logarithmic map can also be evaluated in closed form by inverting the closed-form expression for the exponential map. 

We use the "hat" operator, $\left[ \cdot \right]^{\wedge}$, to map an element, $\boldsymbol\xi$, of the $p$-dimensional Euclidean vector space onto its corresponding element, $\boldsymbol\Xi$, of the Lie algebra. For $\mathfrak{se}(3)$, this operation is defined as:
\begin{equation}
	\label{eqn:se3_hat_operator}
	\boldsymbol\Xi
	\, = \,
	\boldsymbol\xi^{\wedge}
	\, = \,
	\begin{bmatrix}
		\boldsymbol\rho\\
		\boldsymbol\phi
	\end{bmatrix} ^ {\wedge}
	\, = \,
	\begin{bmatrix}
		\boldsymbol\phi^{\wedge} & \boldsymbol\rho\\
		\boldsymbol{\mathrm{0}}^{\top} & 0
	\end{bmatrix}
\end{equation}
where $\boldsymbol\rho \in \mathbb{R}^{3}$ represents the three translational components of $\boldsymbol\xi$ and $\boldsymbol\phi \in \mathbb{R}^{3}$ represents the three rotational components. In this paper, we follow the convention adopted in \cite{barfoot2017state} and \cite{murray2017mathematical} and use the first three components of $\boldsymbol\xi$ to represent $\boldsymbol\rho$ and the last three components to represent $\boldsymbol\phi$. This differs from the convention used in \cite{lynch2017modern}, where the order of these two groups of components within $\boldsymbol\xi$ is reversed. We have also overloaded the operator so that $\boldsymbol\phi^{\wedge}$ is defined as the skew-symmetric matrix:
\begin{equation}
	\label{eqn:so3_hat_operator}
	\boldsymbol\phi^{\wedge}
	\, = \,
	\begin{bmatrix}
		\phi_{1}\\
		\phi_{2}\\
		\phi_{3}
	\end{bmatrix}^{\wedge}
	\, = \,
	\begin{bmatrix}
		0 & -\phi_{3} & \phi_{2}\\
		\phi_{3} & 0 & -\phi_{1}\\
		-\phi_{2} & \phi_{1} & 0
	\end{bmatrix}
	\, \in \,
	\mathbb{R}^{3 \times 3}
\end{equation}

We use the corresponding "vee" operator, $\left[ \cdot \right]^{\vee}$, to perform the inverse mapping from $\boldsymbol\Xi$ to $\boldsymbol\xi$ in the opposite direction: $\boldsymbol\xi \, = \, \boldsymbol\Xi^{\vee}$. In the context of robot kinematics, $\boldsymbol\xi$ is often referred to as a velocity \emph{twist}, and hence its translational and angular components are expressed in units of m/s or rad/s, respectively.

Vectors in the local tangent space at $\boldsymbol{\mathrm{X}} \in G$ are mapped into the global tangent space (Lie algebra), $\mathfrak{g}$, using the \emph{adjoint representation} of $\boldsymbol{\mathrm{X}}$, which is denoted as $\mathrm{Ad} \left( \boldsymbol{\mathrm{X}} \right)$ and defined for $SE(3)$ as:
\begin{equation}
	\label{eqn:se3_adjoint}
	\begin{split}
		\mathrm{Ad} \left( \boldsymbol{\mathrm{X}} \right)
		& \, = \,
		\mathrm{Ad} 
		\left(
		\begin{bmatrix}
			\boldsymbol{\mathrm{C}} & \boldsymbol{\mathrm{r}}\\
			\boldsymbol{\mathrm{0}}^{\top} & 1
		\end{bmatrix}
		\right)\\
		& \, = \,
		\begin{bmatrix}
			\boldsymbol{\mathrm{C}} & \boldsymbol{\mathrm{r}}^{\wedge} \boldsymbol{\mathrm{C}}\\
			\boldsymbol{\mathrm{0}}_{3 \times 3} & \boldsymbol{\mathrm{C}}
		\end{bmatrix}
		\, \in \,
		\mathbb{R}^{6 \times 6}
	\end{split}
\end{equation}

Using this definition, it is straightforward to show that $\mathrm{Ad} \left( \boldsymbol{\mathrm{X}}_{1} \boldsymbol{\mathrm{X}}_{2} \right) = \mathrm{Ad} \left( \boldsymbol{\mathrm{X}}_{1} \right) \mathrm{Ad} \left( \boldsymbol{\mathrm{X}}_{2} \right)$ and $\mathrm{Ad} \left( \boldsymbol{\mathrm{X}} \right)^{-1} = \mathrm{Ad} \left( \boldsymbol{\mathrm{X^{-1}}} \right)$, and hence the adjoint representation can be used to map vectors between any pair of tangent spaces. Sometimes, we need to use the corresponding adjoint representation of $\mathfrak{g}$, which we denote as $\mathrm{ad} \left( \boldsymbol\Xi \right)$ and define for $\mathfrak{se}(3)$ as:
\begin{equation}
	\label{eqn:se3_algebra_adjoint}
	\begin{split}
		\mathrm{ad} \left( \boldsymbol\Xi \right)
		& \, = \,
		\mathrm{ad} \left( \boldsymbol\xi^{\wedge} \right)
		\, = \,
		\mathrm{ad} \left(
		\begin{bmatrix}
			\boldsymbol\rho\\
			\boldsymbol\phi
		\end{bmatrix} ^ {\wedge}
		\right)\\
		& \, = \,
		\begin{bmatrix}
			\boldsymbol\phi^{\wedge} & \boldsymbol\rho^{\wedge} \\
			\boldsymbol{\mathrm{0}} & \boldsymbol\phi^{\wedge}
		\end{bmatrix}
		\, \in \,
		\mathbb{R}^{6 \times 6}
	\end{split}
\end{equation}

In cases where we need to take the product of two exponentials in $\mathfrak{se}(3)$, this is not as straightforward as taking the product of two scalar exponentials, where we can just sum the exponents. Instead, we compute the product using the following approximation, which is based on the \emph{Baker-Campbell-Hausdorff (BCH)} formula (\cite{barfoot2017state}):
\begin{multline}
	\label{eqn:bch_approx}
	\ln(\exp(\boldsymbol\Xi_{1}) \, \exp(\boldsymbol\Xi_{2}))^{\vee}\\
	\, = \,
	\ln(\exp(\boldsymbol\xi^{\wedge}_{1}) \, \exp(\boldsymbol\xi^{\wedge}_{2}))^{\vee}\\
	\approx
	\begin{cases}
		\mathcal{J}(\boldsymbol\xi_{2})^{-1}\boldsymbol\xi_{1} + \boldsymbol\xi_{2} & \text{if $\boldsymbol\xi_{1}$ small}\\
		\boldsymbol\xi_{1} + \mathcal{J}(-\boldsymbol\xi_{1})^{-1}\boldsymbol\xi_{2} & \text{if $\boldsymbol\xi_{2}$ small}\\
	\end{cases} 
\end{multline}

Here, $\mathcal{J} \in \mathbb{R}^{6 \times 6}$ is the \emph{left Jacobian} of $SE(3)$, which is defined by the following series expansion:
\begin{equation}
	\label{eqn:left_jacobian_def}
	\mathcal{J} \left( \boldsymbol\xi \right)
	\, = \,
	\sum_{n=0}^{\infty}
	\frac {1} {\left( n+1 \right)!}
	\left( \boldsymbol\xi^{\curlywedge} \right)^{n}
\end{equation}
where the "curly hat" operator is defined as:
\begin{equation}
	\label{eqn:curly_hat_def}
	\boldsymbol\xi^{\curlywedge}
	\, = \,
	\mathrm{ad} \left( \boldsymbol\xi^{\wedge} \right)
	\, = \,
	\begin{bmatrix}
		\boldsymbol\phi^{\wedge} & \boldsymbol\rho^{\wedge} \\
		\boldsymbol{\mathrm{0}} & \boldsymbol\phi^{\wedge}
	\end{bmatrix}
	\in
	\mathbb{R}^{6 \times 6}
\end{equation}

It is also possible to define a \emph{right Jacobian} of $SE(3)$ but we will not need it here and so will just refer to $\mathcal{J}$ as the \emph{Jacobian}. Similarly, the inverse (left) Jacobian is defined using the series expansion:
\begin{equation}
	\label{eqn:inv_jacobian_def}
	\mathcal{J} \left( \boldsymbol\xi \right)^{-1}
	\, = \,
	\sum_{n=0}^{\infty}
	\frac {B_{n}} {n!}
	\left( \boldsymbol\xi^{\curlywedge} \right)^{n}
\end{equation}
where $B_{n}$ are the \emph{Bernoulli numbers}, $B_{0}=1, \, B_{1}=-\frac{1}{2}, \, B_{2}=\frac{1}{6}, \, B_{3}=0, \, B_{4}=-\frac{1}{30}, \, \ldots$. In this paper, if we need to calculate the Jacobian or its inverse, we typically truncate the corresponding series after second-order terms because we have found that this is sufficiently accurate for our purposes (this is also consistent with the findings of \cite{barfoot2014associating}).

Moving on to consider random variables in $SE(3)$, and following \cite{barfoot2014associating, barfoot2017state, bourmaud2015estimation, bourmaud2016intrinsic}, we define a random variable, $\boldsymbol{\mathrm{X}} \in SE(3)$, in terms of a small random perturbation, $\boldsymbol{\epsilon} \sim \mathcal{N}(\boldsymbol{\mathrm{0}}, \,\boldsymbol{\Sigma})$, in the global tangent space (Lie algebra) composed with a deterministic mean value, $\bar{\boldsymbol{\mathrm{X}}} \in SE(3)$:
\begin{equation}
	\label{eqn:se3_random_variable}
	\boldsymbol{\mathrm{X}}
	\, = \,
	\exp(\boldsymbol\epsilon^{\wedge}) \, \bar{\boldsymbol{\mathrm{X}}}
\end{equation}

The Gaussian perturbation, $\boldsymbol{\epsilon}$, induces a non-Gaussian probability distribution function (PDF) over $\boldsymbol{\mathrm{X}} \in SE(3)$ (\cite{barfoot2014associating}):
\begin{equation}
	\label{eqn:se3_pdf}
	p \left( \boldsymbol{\mathrm{X}} \right)
	\, = \,
	\beta \left( \boldsymbol\epsilon \right)
	\exp \left( -\frac{1}{2} \boldsymbol\epsilon^{\top}
	\boldsymbol{\mathrm{\Sigma}}^{-1} \boldsymbol\epsilon \right)
\end{equation}

Here, $ \boldsymbol\epsilon = \ln \left(\boldsymbol{\mathrm{X}} \bar{\boldsymbol{\mathrm{X}}}^{-1} \right)^{\vee}$ and $\beta \left( \boldsymbol\epsilon \right) = \frac {\eta} {\left| \mathrm{det}( \mathcal{J}(\boldsymbol\epsilon) ) \right|}$. The non-constant normalisation factor $\beta \left( \boldsymbol\epsilon \right)$ derives from the relationship between infinitesimal volume elements in $\mathfrak{se}(3)$ and $SE(3)$ (\cite{barfoot2017state}):
\begin{equation}
	\label{eqn:se3_inf_volume_element}
	d \boldsymbol{\mathrm{X}}
	\, = \,
	\left| \mathrm{det}( \mathcal{J}(\boldsymbol\epsilon) ) \right| \, d \boldsymbol\epsilon
\end{equation}

Sometimes, it is necessary to change variables and express $\boldsymbol{\mathrm{X}}$ as a random variable, $\boldsymbol\xi$, in the global tangent space, and find its PDF in that space (as opposed to only expressing the left-perturbation $\boldsymbol\epsilon$ in the global tangent space). To do this, we first map $\boldsymbol{\mathrm{X}}$ to the global tangent space using the logarithmic map and then approximate it using the BCH approximation (Equation ~\ref{eqn:bch_approx}):
\begin{equation}
	\label{eqn:glob_tangent_space_approx}
	\boldsymbol\xi
	\, = \,
	\ln(\boldsymbol{\mathrm{X}})^{\vee}
	\, \approx \,
	\boldsymbol\mu + \mathcal{J} \left( \boldsymbol\mu \right)^{-1} \boldsymbol\epsilon
\end{equation}
where $\boldsymbol\mu \, = \, \ln(\bar{\boldsymbol{\mathrm{X}}})^{\vee}$. Then, noting that the second term on the right hand side is just a linear transform of the Gaussian random variable, $\boldsymbol\epsilon$, we find that the PDF of $\boldsymbol\xi$ is also approximately Gaussian (see Appendix~\ref{app:probabilistic_transform}):
\begin{equation}
	\label{eqn:glob_tangent_space_pdf}
	\boldsymbol{\xi} \: \dot\sim \: \mathcal{N}(\boldsymbol\mu, \, \mathcal{J}^{-1}(\boldsymbol\mu) \, \boldsymbol{\Sigma} \, \mathcal{J}^{-\top}(\boldsymbol\mu))
\end{equation}

\subsection{Neural network based pose and shear estimation}
\label{sec:nn_pose_shear_estimate}

In our past work, we used convolutional neural networks (CNNs) with multi-output regression heads to estimate contact poses from tactile sensor images (\cite{lepora2020optimal, lepora2021pose, lloyd2021goal}). So, our initial approach on this project was to use a similar CNN to estimate both the contact pose and the post-contact shear from an image. However, this turned out to be problematic because of the level of (unavoidable) noise present in the training labels, particularly in the shear-related outputs. So, instead of using a CNN to produce single-point estimates, we modified the network head to predict the parameters of a \emph{distribution} over pose/shear estimates. We call this modified network a Gaussian density network (GDN) because of the assumed Gaussian distribution of estimates. We then pass the distribution estimated by the GDN to a Bayesian filter to reduce the noise-related errors over a sequence of estimates.

In this section, we begin by describing how we combine contact pose and post-contact shear information in a single \emph{surface contact pose} to reduce the complexity of the filtering and control stages. We then describe the process we use to collect and pre-process the training data, and the CNN and GDN pose/shear estimation models that we use in this work. We include the CNN network in the discussion because it shares the same convolutional base as the GDN model and we also use it as a baseline for comparing the GDN performance against (Section~\ref{sec:nn_pose_shear_estimate_exp}).

\subsubsection{Surface contact poses: combining pose and shear information}
\label{sec:surface_contact_poses}

In previous work (\cite{lepora2020optimal, lepora2021pose, lloyd2021goal}), we assumed that a surface contact pose in Euler representation can be represented by a 6-component vector, $\left( 0,0,z,\alpha,\beta,0 \right)$. Here, the $z$-component denotes the contact depth and the $\left( \alpha,\beta \right)$-components denote the two contact angles that define the orientation of the sensor with respect to the surface. The three remaining components are set to zero because we assumed that all surface contacts are invariant to $\left( x,y \right)$-translation and $\gamma$-rotation parallel to the surface. Since this is only true in situations where there is no frictional contact between the sensor and surface or there is no post-contact motion parallel to the surface, we applied a form of domain randomization during training to encourage the CNN to learn features that are covariant with the $\left( z,\alpha,\beta \right)$ variables we were trying to estimate and invariant to the $\left( x=0,y=0,\gamma=0 \right)$ variables we trying to ignore. This helped ensure that the contact depth and surface orientation angles estimated by the trained model were accurate in situations where post-contact shear distorted the tip of the sensor and the corresponding marker images (e.g., during sliding contact).

\begin{figure*}
	\centering
	\includegraphics[width=\textwidth]{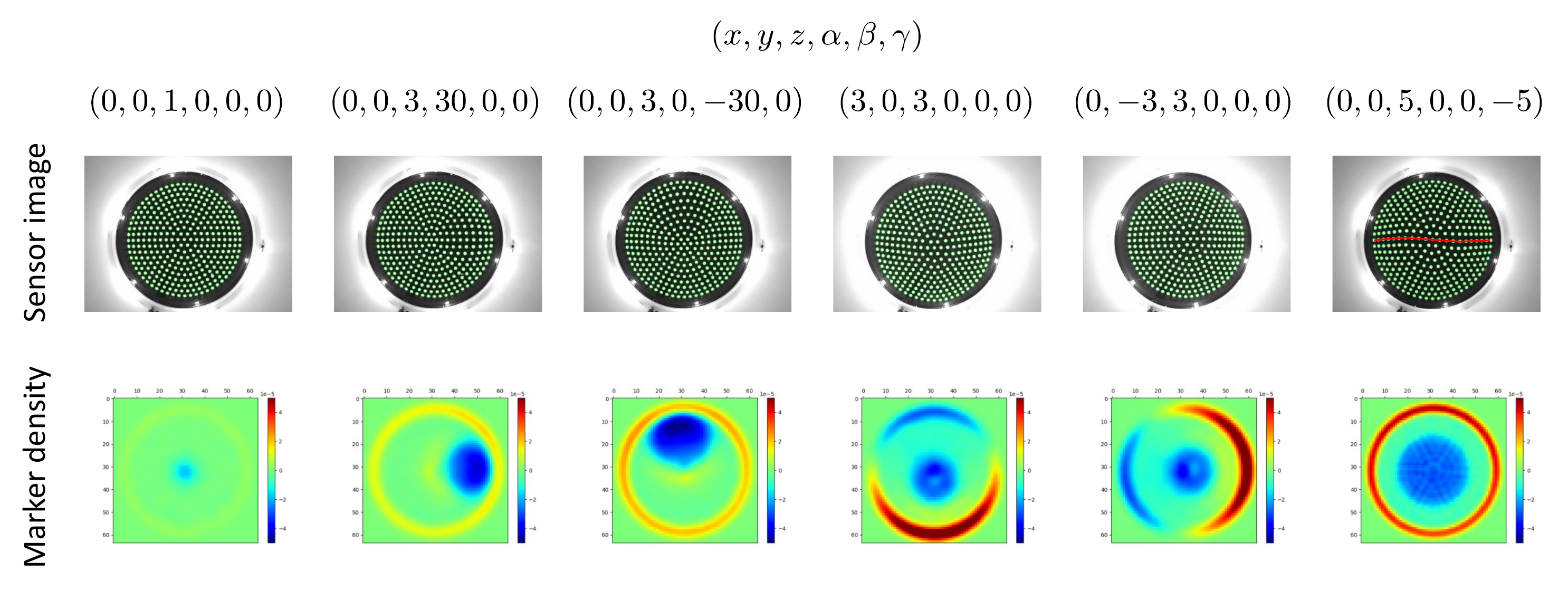}
	\caption{Tactile sensor images and the corresponding changes in marker density with respect to the quiescent, no-contact state for six different surface contact poses (annotated above sensor images). Marker densities are estimated using a kernel density model with Gaussian kernels located at marker centroids and a constant kernel width equal to the mean distance between adjacent markers. The size of the low-density blue region in the centre of the image depends on the contact depth, while its location in the image depends on the sensor orientation. Changes in marker density around the periphery of the sensor depend on the post-contact translational shear, and subtler changes in pattern within the blue contact region depend on the post-contact rotational shear.
	\label{fig:marker_density_images}}
\end{figure*}

In this paper, we remove this assumption and instead train a model to estimate all six non-zero components of a surface contact pose, $\left( x,y,z,\alpha,\beta,\gamma \right)$. Based on early exploratory work where we visualized the marker densities of sensor images using a kernel density model (see \cite{silverman2018density}, for example) we were confident that the sensor images contained enough information to produce these estimates (Figure~\ref{fig:marker_density_images}). Marker densities are a type of feature that CNNs can easily replicate if needed, by applying a sequence of convolution and pooling operations. In our revised definition of a \emph{surface contact pose}, we make the following two assumptions:
\begin{enumerate}
	\item The contacted surfaces can be locally approximated by flat surfaces.
	\item The sensor-surface contacts can be (approximately) decomposed into an equivalent normal contact motion followed by a tangential shear motion (i.e., a motion that is parallel to the contacted surface).
\end{enumerate}

The first assumption is reasonable for smooth, gently curving surfaces and it has produced consistently good results where we have implicitly used it in the past (\cite{lepora2020optimal, lepora2021pose, lloyd2021goal}). In practice, it also seems to hold for somewhat sharper, non-smooth surfaces, which we believe is due to the smoothing effect that the rubber-like sensor skin has on the mechanical-optical transduction process for sharper stimuli.

At first sight, the second assumption might appear somewhat more controversial because there are infinitely many complex trajectories that a tactile sensor can follow when making contact with a surface. However, there are two main reasons why we use this assumption in our work. Firstly, from a practical point-of-view, it is simply not tractable to generate and sample all of the possible contact trajectories when gathering training data for neural network models. There are just too many possibilities and so we need to simplify the process. Secondly, we have used this two-step, normal-contact-followed-by-shear procedure to capture training data in our earlier work and found that we could train models that produced accurate pose estimates even in situations where the assumption was not strictly valid (e.g., for arbitrary sliding and pushing contact with surfaces). In other words, while the assumption may not be strictly true in all cases, it does seem to produce good results in most practical situations we have encountered.

\begin{figure}
	\centering
	\includegraphics[width=\columnwidth]{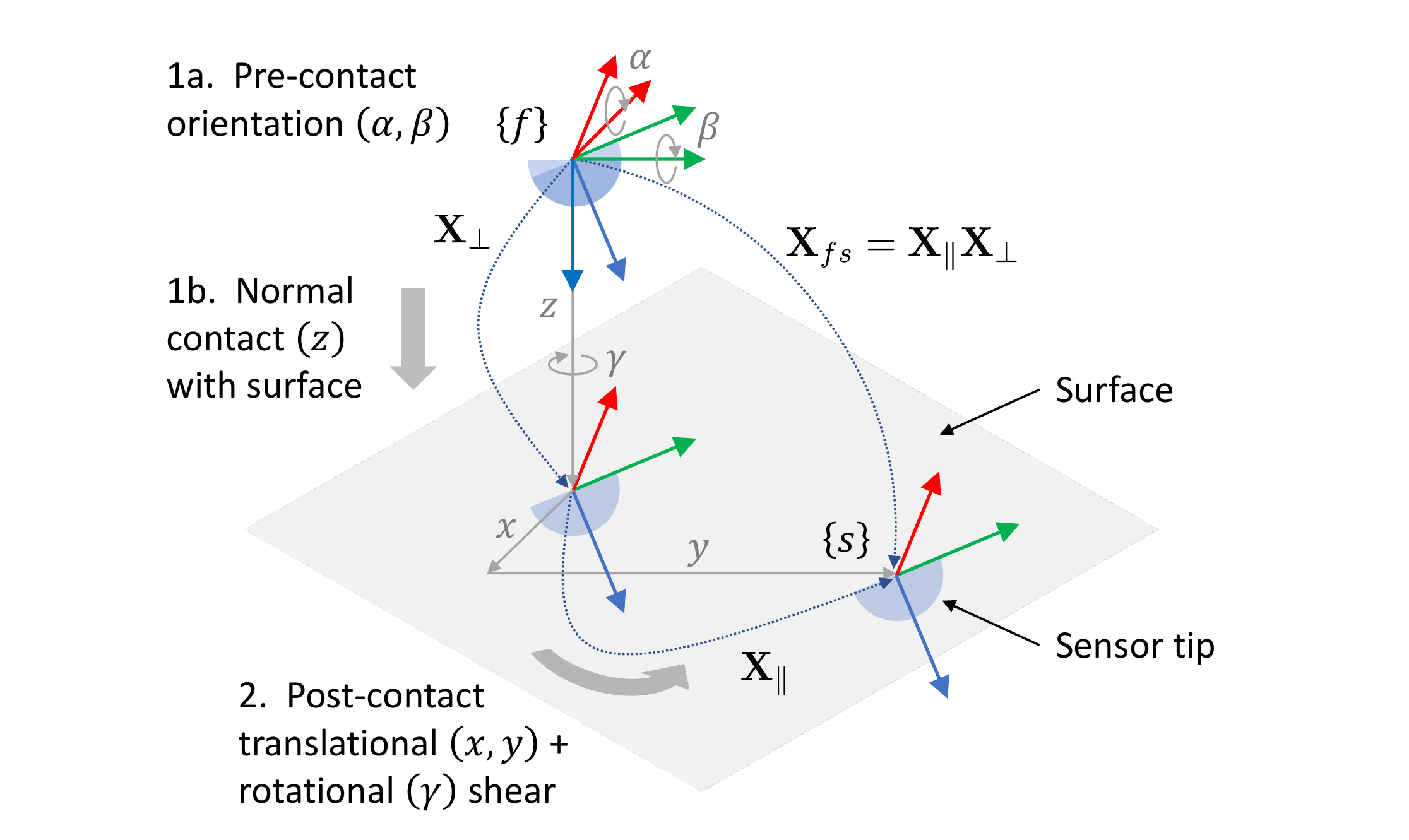}
	\caption{Definition and generation of surface contact poses using a two-step process of normal contact motion followed by translational and rotational shear. Step 1a: prior to normal contact motion, the sensor is rotated by Euler angles $(\alpha,\beta)$ with respect to the surface plane. Step 1b: the sensor is brought into normal contact with the surface through distance $z$. Step 2: the sensor is translated by $(x,y)$ parallel to the surface, while simultaneously being rotated through angle $\gamma$ about the normal contact axis.
	\label{fig:surface_contact_pose_def}}
\end{figure}

With these two assumptions in mind, we now define the surface contact poses we use to train our pose/shear estimation models (Figure~\ref{fig:surface_contact_pose_def}) and describe the process for sampling poses and recording the corresponding tactile sensor images.

We start by attaching a sensor coordinate frame, $\{s\}$, to the centre of the hemispherical sensor tip so that the $z$-axis is directed outwards from the tip of the sensor, along its radial axis. We also attach a surface feature frame, $\{f\}$, to the surface so that its $z$-axis is normal to and directed inwards towards the surface. The $\{f\}$-frame is also located so that it is aligned with the sensor frame, $\{s\}$, when the sensor is in its initial position, just out of normal contact with the surface.

As discussed above, the surface contact motion is assumed equivalent to one that is carried out in two stages: a normal contact motion followed by a post-contact, tangential shear motion. The normal contact motion, represented by an $\{f\}$-frame $SE(3)$ transform, $\boldsymbol{\mathrm{X}}_{\perp}$, rotates the sensor by Euler angles $(\alpha,\beta)$ with respect to the surface (assuming an extrinsic-$xyz$ Euler convention) and then brings it into normal contact with the surface through distance $z$. The tangential shear motion is represented by another $\{f\}$-frame $SE(3)$ transform, $\boldsymbol{\mathrm{X}}_{\parallel}$, which translates the sensor by a displacement $(x,y)$ parallel to the surface, while simultaneously rotating it about the normal contact axis through an angle $\gamma$.

Since Step 1 of the motion involves translation only along the $z$-axis and Step 2 involves rotation only about the $z$-axis, we can interchange the order of these operations and hence the composite $\{f\}$-frame motion is equivalent to a pure rotation $(\alpha,\beta,\gamma)$ followed by a pure translation $(x,y,z)$, which can be represented using a single $SE(3)$ transform: $\boldsymbol{\mathrm{X}}_{fs} = \boldsymbol{\mathrm{X}}_{\parallel} \boldsymbol{\mathrm{X}}_{\perp}$. We use this composite transform, with Euler representation $(x,y,z,\alpha,\beta,\gamma)$, to represent the surface contact pose of the sensor in the surface feature frame $\{f\}$. Combining the contact pose and post-contact shear information in this way simplifies the subsequent filtering and control stages because it avoids the need for two separate filters and two pose-based controllers.

It is important to note that here we have not explicitly assumed that the sensor does not slip when making contact with the surface. This is because in practice it is difficult to pre-specify contact poses where we can be confident this will not happen (e.g., when making light sliding contact with a surface). However, if slip does occur, the second of our two assumptions still holds if we assume that there is some \emph{equivalent} normal contact followed by tangential shear motion (without slip) that produces the same sensor image that is produced under slip conditions.

\subsubsection{Data Collection.}
\label{sec:data_collection}

We collect data for training our pose estimation models by using a robot arm to move the sensor into different contact poses with a flat surface, in a two-step motion, and recording the corresponding tactile sensor image in each case. Each data sample consists of a $640 \times 480$, 8-bit gray scale image together with the corresponding surface contact pose in extrinsic-$xyz$ Euler format. We use an Euler format during data collection because it is directly supported by the robot application programming interface (API) and is easier for humans to visualize than other pose representations such as $4 \times 4$ matrices or quaternions. However, if required, we can easily convert to other formats using the Python \emph{transforms3d} library. We sample the surface contact poses, $(x,y,z,\alpha,\beta,\gamma)$, according to the following rules:
\begin{itemize}
	\item $x$ and $y$ are sampled so that the translational shear displacements are distributed uniformly over a disk of radius $r_{\mathrm{max}}=5.0$ mm centred on the initial point of normal contact with the surface.
	\item $z$ is sampled uniformly in the range $[0.5, 6.0]$ mm.
	\item $\alpha$ and $\beta$ are sampled so that contacts with the sensor are distributed uniformly over a spherical cap of the sensor that is subtended by angle $\phi_{\mathrm{max}}=25$ degrees with respect to its central axis.
	\item $\gamma$ is sampled uniformly at random in the range $[-5.0, 5.0]$ degrees.
\end{itemize}

The sampling of $x$ and $y$ over a disk, is defined as follows:
\begin{equation}
	\label{eqn:disk_random_sampling}
	\begin{gathered}
		x = r \cos \theta, \; y = r \sin \theta\\
		r = r_{\mathrm{max}}\sqrt{r^{\prime}}\\
		r^{\prime} \sim \mathcal{U}(0, 1), \;\; \theta \sim \mathcal{U}(0, 2\pi)
	\end{gathered}
\end{equation}
where $\mathcal{U}(a, b)$ denotes a continuous uniform PDF over the interval, $\left[ a, b \right]$. The sampling of $\alpha$ and $\beta$ over a spherical cap is based on the method outlined by \cite{simon2015generating}:
\begin{equation}
	\label{eqn:sphere_random_sampling}
	\begin{gathered}
		\alpha = -\arcsin y, \;\; \beta = -\mathrm{atan2}(x, z)\\
		x = \sin \phi \cos \theta, \;\; y = \sin \phi \sin \theta, \;\; z = \cos \phi\\
		\phi = \arccos (1 - (1 - \cos \phi_{\mathrm{max}}) \phi^{\prime})\\
		\phi^{\prime} \sim \mathcal{U}(0, 1), \;\; \theta \sim \mathcal{U}(0, 2\pi)
	\end{gathered}
\end{equation}

After generating the required number of random pose samples (i.e., training labels), we collect the corresponding sensor images using a two-step procedure that mirrors the definition of the surface contact poses given in the previous section (Algorithm~\ref{alg:data_collection}).

\begin{algorithm}
	\caption{Data collection}
	\label{alg:data_collection}
	\begin{algorithmic}
	\Require Surface contact poses $\boldsymbol{\mathrm{X}}_{i}$, $i=1 \ldots N$
	\Ensure Sensor images $\boldsymbol{\mathrm{I}}_{i}$ for poses $\boldsymbol{\mathrm{X}}_{i}$, $i=1 \ldots N$
	\For{each pose $(x,y,z,\alpha,\beta,\gamma) = \boldsymbol{\mathrm{X}}_i$, }
		\State 1. Move sensor to start point above surface
		\State 2. Rotate sensor to contact angle $(\alpha, \beta)$
		\State 3. \parbox[t]{\dimexpr\linewidth-\algorithmicindent}{%
			Move sensor normal to surface to make contact at depth $z$
		}
		\State 4. \parbox[t]{\dimexpr\linewidth-\algorithmicindent}{%
			Move sensor parallel to surface through translation $(x,y)$ and rotation $\gamma$
		}
		\State 5. Capture sensor image $\boldsymbol{\mathrm{I}}_{i}$
	\EndFor
	\end{algorithmic}
\end{algorithm}

Three distinct data sets were used to develop the pose estimation models: a training set of 6000 samples, a validation set of 2000 samples for model selection and hyper-parameter tuning, and a test set of 2000 samples for independently verifying the performance of the model after training.

\subsubsection{Pre- and post-processing.}
\label{sec:pre_post_processing}

We used the following steps to pre-process the raw sensor images of the training, validation and test sets, and to pre-process images after the model is deployed:

\begin{enumerate}
	\item Crop the images to a $430 \times 430$ pixel square that encloses the circular marker region (removes regions of the image that do not contain any relevant information).
	\item Apply a $5 \times 5$ median blur to remove sensor noise (e.g., due to dust particles on the lens).
	\item Apply an adaptive threshold to convert the image to binary format (assumes that relevant information is encoded in the marker positions rather than the shading).
	\item Resize the images to $128 \times 128$ pixels.
	\item Convert the integer pixel values to floating point values and normalise them so that they lie in the range $[0, 1]$.
\end{enumerate}

We also pre-processed the pose labels of the training, validation and test sets so that the estimates produced by the trained models are in the correct format for the subsequent filtering and controller stages:

\begin{enumerate}
	\item Convert pose labels from their Euler representations to $4 \times 4$ homogeneous matrices $\boldsymbol{\mathrm{X}}_{fs} \in \mathbb{R}^{4 \times 4}$.
	\item Invert the $4 \times 4$ matrices, so that instead of representing sensor poses in the surface feature frames they now represent surface feature poses in the sensor frames: $\boldsymbol{\mathrm{X}}_{sf}=\boldsymbol{\mathrm{X}}_{fs}^{-1}$.
	\item Convert the inverted matrices to exponential coordinates: $\boldsymbol\xi_{sf}=\ln(\boldsymbol{\mathrm{X}}_{sf})^{\vee} \in \mathbb{R}^{6}$ (see Section~\ref{sec:math_prelim}).
\end{enumerate}

\subsubsection{Pose and shear estimation using convolutional neural networks.}
\label{sec:cnn_model}

\begin{figure*}
	\centering
	\includegraphics[width=\textwidth]{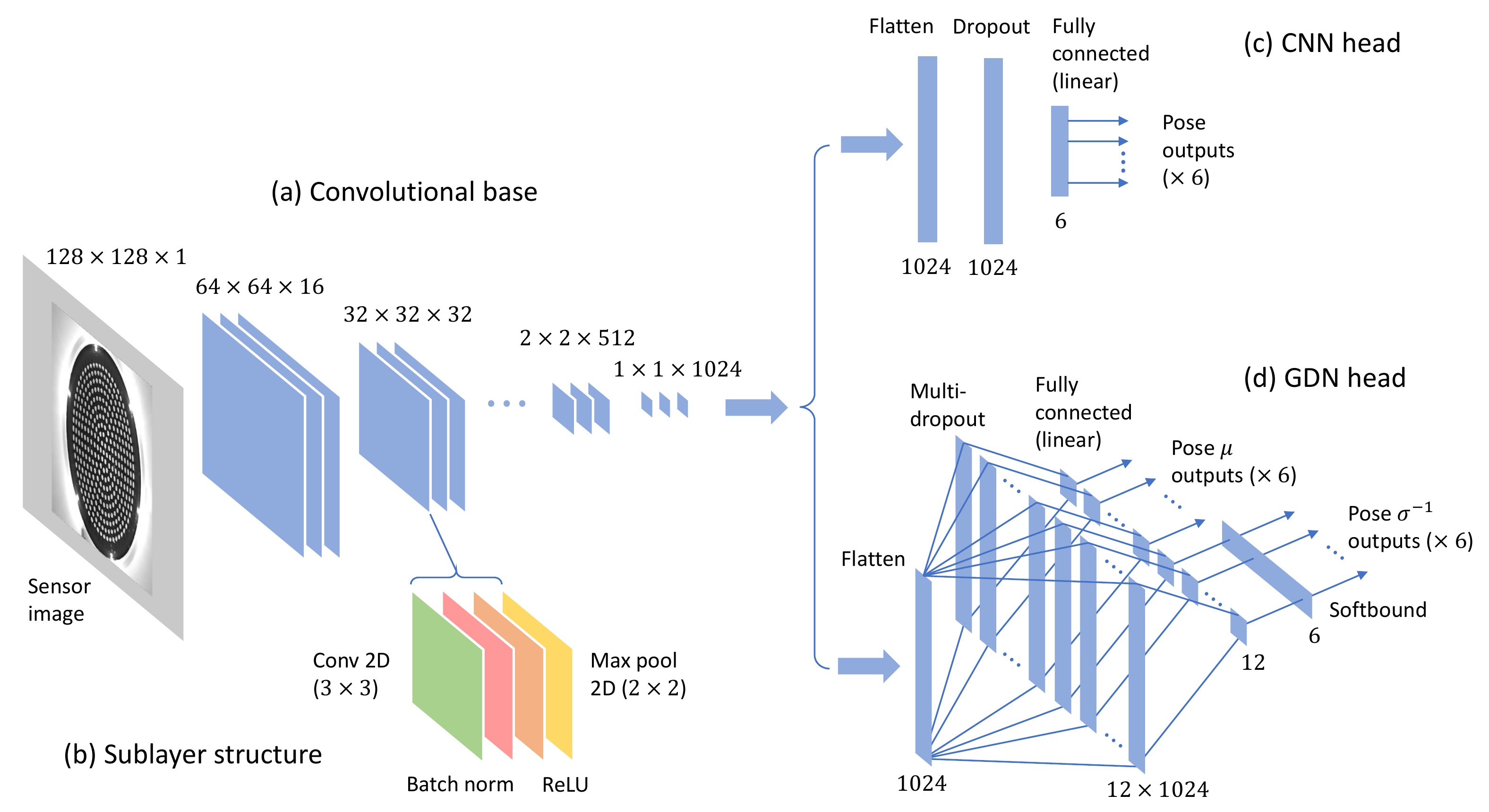}
	\caption{CNN and GDN architectures used for surface contact pose estimation. (a) Convolutional base for CNN and GDN models. (b) Convolutional block sub-layer structure. (c) CNN multi-output regression head. (d) GDN distribution parameter estimation head.
	\label{fig:cnn_gdn_architecture}}
\end{figure*}

Initially, we tried using multi-output regression CNNs similar to the ones we used in our earlier work for estimating surface contact poses. Later on, we used these CNN models as a baseline for evaluating the performance of our more capable Gaussian density network (GDN) models. These CNN models are constructed from a sequence of convolutional layer blocks, where each block is composed of a sequence of sub-layers: $3 \times 3$, 2D convolution; batch normalisation (\cite{ioffe2015batch}); rectified linear unit (ReLU) activation function; and $2 \times 2$ max-pooling. The feature map dimensions are reduced by half at each block as we move forwards through the blocks, due to the max-pooling that takes place in the final sub-layer of each block. So we balance the progressive loss of feature resolution by doubling the number of features in consecutive blocks. This is important if we want to obtain accurate pose estimates from coarse features that represent different spatial distributions of sensor markers.

The output of the convolutional base feeds into a densely-connected, multi-output regression head, composed of a flatten layer, dropout layer with dropout probability $p=0.1$ (\cite{srivastava2014dropout}), and a single fully-connected layer with a linear activation function. When a pre-processed sensor image is applied as input to the CNN, it outputs a surface contact pose estimate in exponential coordinates. If required, we convert the pose estimates, $\hat{\boldsymbol\xi}_{i}$ from exponential coordinates back to $4 \times 4$ homogeneous matrices, $\hat{\boldsymbol{\mathrm{X}}}_{i}$, using, $\hat{\boldsymbol{\mathrm{X}}}_{i} = \exp(\hat{\boldsymbol\xi}_{i}^{\wedge})$ (see Section~\ref{sec:math_prelim}).

We train the CNN by minimising the following weighted mean-squared error (MSE) loss function, which is defined over $N$ training examples and $M=6$ network outputs (three translational pose components and three rotational components):
\begin{equation}
	\label{eqn:mse_loss_def}
	\mathrm{MSE}
	\, = \,
	\frac{1}{N} \sum_{i=1}^{N} \sum_{j=1}^{M} \alpha_{j} \left(\xi_{ij} - \hat{\xi}_{ij} \right)^{2}
\end{equation}

Here, $\xi_{ij}$ is the ${j}$th component of the $i$th sample pose label (in exponential coordinates), and $\hat{\xi}_{ij}$ is the corresponding CNN output. The loss weights, $\alpha_{j}$, are hyperparameters, which can be varied to compensate for different output scales (e.g., translation outputs may be expressed in mm/s and angular outputs in rad/s) and to avoid over-fitting when some outputs are noisier than others. Through trial and error, we found a good set of weights to be, $ \boldsymbol\alpha = (\alpha_{j}) = (1,1,1,100,100,100)$.

To get a meaningful error metric in the appropriate units for each output, we also computed the mean absolute error (MAE) for each CNN output. This helps identify situations where the overall performance is dominated by a subset of outputs. The MAE for output $j$ is defined over $N$ training examples as:
\begin{equation}
	\label{eqn:mae_metric_def}
	\mathrm{MAE}_{j}
	\, = \,
	\frac{1}{N} \sum_{i=1}^{N} \left|\xi_{ij} - \hat{\xi}_{ij} \right|
\end{equation}

We trained the CNN models using the Adam optimizer, with a batch size of 16 and a linear rise, polynomial decay (LRPD) learning rate schedule. In our implementation of this schedule, we initialised the learning rate to 1e\nobreakdash-5 and linearly increased it to 1e\nobreakdash-3 over 3 epochs; we then maintained it for a further epoch before decaying it to 1e\nobreakdash-7 over $e_{\mathrm{max}} = 50$ epochs using a $\sqrt{1-{e}/{e_{\mathrm{max}}}}$ polynomial decay weighting factor. We found that a good learning rate schedule can make the training process less sensitive to a particular choice of learning rate and generally improves the performance of the trained model. We used "early stopping" to terminate the training process when the validation loss reached its minimum value over a "patience" of 25 epochs.

\subsubsection{Gaussian Density Networks: Pose and shear estimation with uncertainty.}
\label{sec:gdn_model}

In our early experiments, we found that the pose estimates produced by our trained CNN models were not as accurate as those obtained in previous work. We think this was mainly due to training samples that now included much shallower contact depths than before, so it was more likely that the sensor would slip across the surface during the shear moves that followed normal contacts. This can cause tactile aliasing where similar images become associated with very different pose labels and this increases the estimation error and uncertainty (\cite{lloyd-RSS-21}). This theory is corroborated by our results to some extent (Section~\ref{sec:nn_pose_shear_estimate_exp}), which show that the estimation errors are indeed larger for pose components associated with shear motion. We also observed that the Franka robot arms used in this study were not as accurate as the ABB Robotics and Universal Robots (UR) robot arms we used in previous work and this could be another factor behind the drop in performance.

To address this problem, we modified the CNN regression head to estimate the parameters of a Gaussian pose distribution rather than produce a single-point estimate. This allows us to estimate both the surface contact pose and its associated uncertainty (Figure~\ref{fig:cnn_gdn_architecture}). The motivation for doing this was discussed in \cite{lloyd-RSS-21}: if we know the uncertainty associated with a pose (or, even better, the full pose distribution), we can use this information to reduce the error and uncertainty using other system components such as the Bayesian filter described in the next section. We refer to this modified CNN model as a Gaussian density network (GDN) because it uses a CNN to predict the parameters of a multivariate Gaussian PDF that captures uncertainty in the pose outputs.

The GDN model can also be viewed as a degenerate (i.e., single-component) mixture density network (MDN), which performs a similar function to the GDN but uses a Gaussian mixture model to model the output distribution (\cite{bishop1994mixture, bishop2006pattern}). This is relevant because the difficulties encountered when training MDNs are well-documented, and include problems such as training instability and mode collapse (\cite{hjorth1999regularisation, makansi2019overcoming}). Although our GDN model does not suffer these problems to the same extent because it assumes a simpler form of output PDF, we found that a na\"ive approach (e.g., using a vanilla CNN to predict the parameters of a TensorFlow Probability Gaussian distribution with a \emph{softplus} layer to constrain the standard deviation outputs to positive values) was prone to instability and slow progress, particularly at the start of training. To overcome these difficulties, we incorporated several novel extensions to our architecture, which we now discuss.

Firstly, rather than use a CNN to directly estimate the component means and standard deviations of a multivariate Gaussian pose distribution (assuming a diagonal covariance matrix), we instead estimate the means and \emph{inverse standard deviations} (i.e., the square root of the inverse covariance or \emph{precision matrix}). Consequently, the estimated means and their corresponding inverse standard deviations appear as products in the mean negative log-likelihood loss function instead of quotients, as they would otherwise do if the network estimated the standard deviations. This is important because we found that using a neural network to simultaneously estimate two variables that appear as quotients in a loss function can cause instability or slow progress during training. We found this to be particularly true in the early stages of training when the outputs tend to be smaller due to smaller initial network weights, or later in the training process if the gradient is driving the quotient denominator to become very small.

Secondly, we introduce a new activation function layer, which we call a \emph{softbound} layer. We use this layer to bound the values of the (inverse) standard deviation within a pre-defined range to prevent it from becoming too large or too small, which we have also found helps speed up training and reduce instability.

Finally, we include a new dropout configuration for multi-output neural networks, which allows us to apply different dropout probabilities to different outputs - something that is not possible when a conventional dropout layer is inserted before the output layer. We call this type of dropout configuration, \emph{multi-dropout}. In our early experiments, we found that for our pose estimation task dropout is more effective than other forms of regularisation such as L2 regularization, and so we needed a way to vary the amount of dropout across different outputs. As a passing observation, we note that very little appears to have been published on how to adapt single-output or single-task neural networks to cope with problems encountered in multi-output or multi-task scenarios (e.g., situations where some output labels are much noisier than others and hence global regularization approaches are less effective than in the single-output/task case).

Our GDN architecture uses the same convolutional base as the original CNN architecture, but instead of feeding its output through a regular multi-output regression head, we feed it through a modified GDN head that includes the enhancements discussed above. Inside the GDN head, the output of the convolutional base is flattened and replicated to a set of 12 dropout layers, one for each of the 12 network outputs. Each of these dropout layers feeds into a separate single-output, fully-connected output layer with a linear activation function. The outputs of the first six output layers estimate the mean of the Gaussian PDF; the outputs of the remaining six output layers are passed through a \emph{softbound} layer to estimate the inverse standard deviations that define the diagonal (inverse) covariance matrix. Since each single-output output layer has its own dedicated dropout layer connecting back to the flatten layer, a different dropout probability can be used with each layer to control the relative amount of regularization for each output. We use a higher level of dropout to increase the regularization on the noisier shear-related outputs, and a lower level of dropout on the remaining pose-related outputs. Through trial-and-error, we found a good set of dropout probabilities to be, $\boldsymbol{p}_{\mu}=(0.7,0.7,0.1,0,0,0.4)$ and 
$\boldsymbol{p}_{\sigma^{-1}}=(0.1,0.1,0,0,0,0.05)$.

We define the \emph{softbound} function in terms of the well-known \emph{softplus} function, using:
\begin{multline}
	\label{eqn:softbound_def}
	\mathrm{softbound} \left( x \right)
	\, = \,
	x_{\mathrm{min}}
	+ \mathrm{softplus} \left( x - x_{\mathrm{min}} \right)\\
	- \mathrm{softplus} \left( x - x_{\mathrm{max}} \right)
\end{multline}
where
\begin{equation}
	\label{eqn:softplus_def}
	\mathrm{softplus} \left( x \right)
	\, = \,
	\ln \left( 1 + \exp \left( x \right) \right)
\end{equation}

Assuming that $x_{\mathrm{max}} \geq x_{\mathrm{min}}$, the softbound function implements the following approximation:
\begin{equation}
	\label{eqn:softbound_approx}
	\begin{aligned}
		\mathrm{softbound} \left( x \right)
	 	\approx
		x_{\mathrm{min}}
		+ \mathrm{max} \left(0, x - x_{\mathrm{min}} \right)\\
		- \mathrm{max} \left(0, x - x_{\mathrm{max}} \right)\\
		\, = \,
		\begin{cases}
		x_{\mathrm{min}} & \text{if $x < x_{\mathrm{min}}$}\\
		x & \text{if $x_{\mathrm{min}} \leq x \leq x_{\mathrm{max}}$}\\
		x_{\mathrm{max}} & \text{if $x > x_{\mathrm{max}}$}
		\end{cases}
	\end{aligned} 
\end{equation}

If required, the function argument, $x$, can be inversely scaled by a temperature parameter, $T$, to control the softness of the transition between the linear region and its bounds, as is sometimes done in the case of the softplus and \emph{softmax} functions, but we did not find this necessary in this project. Details of how to implement this function in a numerically stable way are included in Appendix~\ref{app:softbound implementation}. A softbound layer applies the softbound function to each of its inputs to produce a corresponding set of outputs. In the GDN architecture, we use this type of layer to bound the inverse standard deviations in the range $\sigma_{i}^{-1} \in$ [1e\nobreak-6, 1e6].

We use the GDN outputs, $\hat{\boldsymbol\mu}_{i} = (\hat{\mu}_{ij})$ and $\hat{\boldsymbol\sigma}_{i}^{-1} = (\hat{\sigma}_{ij}^{-1})$, for the $i$th input image to estimate the parameters of a multivariate Gaussian PDF over $\boldsymbol\xi_{i} = (\xi_{ij})$, $j = 1 \ldots M, \, M = 6$:
\begin{multline}
	\label{eqn:multi_gaussian_pdf}
	p \left( \boldsymbol\xi_{i} \right)
	\, = \,
	\frac{1}{\sqrt{ \left( 2 \pi \right)^{M}}}
	\prod_{j=1}^{M} \hat{\sigma}_{ij}^{-1}\\
	\times
	\exp \left( - \frac{1}{2} \sum_{j=1}^{M}
	\left({\hat{\sigma}_{ij}^{-1}} \left( \xi_{ij} - \hat{\mu}_{ij} \right) \right)^{2} \right)
\end{multline}

Here, to simplify the model and reduce the amount of data needed to train it, we have assumed a diagonal covariance matrix of the form:
\begin{equation}
	\label{eqn:multi_gaussian_cov}
	\hat{\boldsymbol\Sigma}_{i}
	\, = \,
	\begin{bmatrix}
		1 / \hat{\sigma}_{i1}^{-2} & & 0 \\
		& \ddots & \\
		0 & & 1 / \hat{\sigma}_{iM}^{-2}
	\end{bmatrix}
\end{equation}

Where necessary, we convert the (mean) pose estimates from exponential coordinates to $4 \times 4$ homogeneous matrices using $\bar{\boldsymbol{\mathrm{X}}}_{i} = \exp(\hat{\boldsymbol\mu}_{i}^{\wedge})$. However, for the GDN model an additional step is required because the covariance matrix that represents uncertainty in the pose estimates relates to a Gaussian distribution around the mean in the global tangent space, whereas our chosen method of representing random variables in $SE(3$) is based on a mean $SE(3)$ pose that is perturbed on the left by a zero-mean Gaussian random variable in the global tangent space. With reference to Section~\ref{sec:math_prelim}, we compute the required covariance of the left perturbation by inverting the expression for the global covariance in Equation~\ref{eqn:glob_tangent_space_pdf}:
\begin{equation}
	\label{eqn:left_perturbation_cov}
	\boldsymbol\Sigma_{i}
	\, = \,
	\mathcal{J}(\hat{\boldsymbol\mu}_{i}) \hat{\boldsymbol\Sigma}_{i} \mathcal{J}(\hat{\boldsymbol\mu}_{i})^{\top}
\end{equation}

We train the GDN model by minimising a mean NLL loss function, $\overline{\mathrm{NLL}}$, which we define as follows:
\begin{equation}
	\label{eqn:gdn_nll_loss_def}
	\mathrm{NLL}
	\, = \,
	- \ln \mathcal{L}
	\, = \,
	-\ln \prod_{i=1}^{N} p \left( \boldsymbol\xi_{i} \right)
	\, = \,
	-\sum_{i=1}^{N} \ln p \left( \boldsymbol\xi_{i} \right)\\
\end{equation}

\begin{multline}
	\label{eqn:gdn_nll_loss_def_explicit}
	\mathrm{NLL}
	\, = \,
	\frac{NM}{2} \ln \left( 2 \pi \right)\\
	+ \frac{1}{2} \sum_{i=1}^{N} \sum_{j=1}^{M}
	\left[ \left( \hat{\sigma}_{ij}^{-1} \left( \xi_{ij} - \hat{\mu}_{ij} \right) \right)^{2} - 2 \ln \hat{\sigma}_{ij}^{-1} \right]
\end{multline}

\begin{multline}
	\label{eqn:gdn_mean_nll_loss_def}
	\overline{\mathrm{NLL}}
	\, = \,
	\frac{M}{2} \ln \left( 2 \pi \right)\\
	+ \frac{1}{2N} \sum_{i=1}^{N} \sum_{j=1}^{M}
	\left[ \left( \hat{\sigma}_{ij}^{-1} \left( \xi_{ij} - \hat{\mu}_{ij} \right) \right)^{2} - 2 \ln \hat{\sigma}_{ij}^{-1} \right]
\end{multline}

Comparing this definition to Equation~\ref{eqn:mse_loss_def}, we can see that if $\hat{\sigma}_{ij} = 1$ for all $i, j$, minimizing $\overline{\mathrm{NLL}}$ is equivalent to minimising MSE. Moreover, the squared inverse standard deviations in Equation~\ref{eqn:gdn_mean_nll_loss_def} play a similar role to the loss weights in Equation~\ref{eqn:mse_loss_def}.

We trained the GDN model in the same way that we trained the CNN model, using the Adam optimizer, with a batch size of 16 and the same linear rise, polynomial decay (LRPD) learning rate schedule. As before, we terminated the training process when the validation loss reached its minimum value over a "patience" of 25 epochs.

As a final remark, we note that other types of neural network base (e.g., a stack of fully-connected layers) could be used instead of the convolutional base used in the CNN or GDN models, to adapt the pose estimation model for use with other types of tactile sensor.

\subsection{Reducing error and uncertainty in pose and shear estimates: Discriminative Bayesian filtering in SE(3)}
\label{sec:se3_bayes_filter}

In this section, we derive the $SE(3)$ discriminative Bayesian filter we use for reducing the error and uncertainty of GDN pose estimates. Our filter recursively combines a sequence of GDN pose estimates using proprioceptive information from the robot to transform a previous pose estimate to the current sensor coordinate frame so that it can be meaningfully combined with the current estimate to produce a more accurate combined estimate. Our filter differs from more conventional types of Bayesian filter in two important ways. Firstly, it assumes a \emph{discriminative state model} rather than the more standard generative observation model. Secondly, it uses \emph{$SE(3)$ PDFs} to represent state distributions (see Section~\ref{sec:math_prelim}) rather than the Gaussian distributions typically used in Kalman filters or their extensions, or the sample-based representations used in particle filters. We avoid using Gaussian distributions because pose distributions in $SE(3)$ tend to be more "banana-shaped" than ellipsoid, and as uncertainty increases many algorithms become inconsistent when the Gaussian assumption breaks down (\cite{long2013banana}). We avoid using particle filters because they tend to be computationally expensive and hence are less appropriate for real-time applications such as ours.

\subsubsection{A discriminative Bayesian filter.}
\label{sec:discriminative_bayes_filter}

\begin{figure*}
	\centering
	\includegraphics[width=\textwidth]{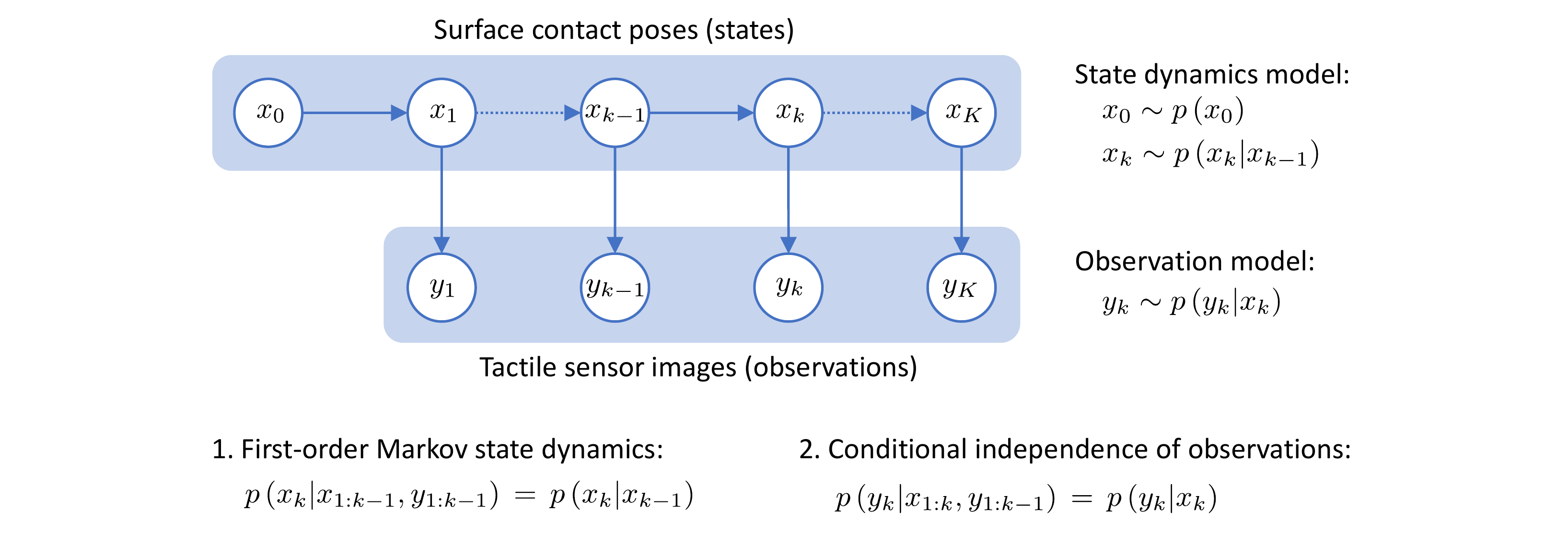}
	\caption{Probabilistic state space model that we use to describe the relationship between surface contact poses (states) and tactile sensor images (observations) in our Bayesian filter.
	\label{fig:state_space_model}}
\end{figure*}

To simplify the notation in this section, we use lower-case italic letters (e.g., $x$ or $y$) to represent continuous random variables, regardless of whether they are scalars, vectors, or $SE(3)$ objects. We model the sequential pose estimation problem using a probabilistic state-space model (Figure~\ref{fig:state_space_model}) that is defined by two interrelated sequences of conditional PDFs: the state dynamics model and the observation model. 

The \emph{state dynamics model} describes how states, $x_{k}$, are transformed between time steps, $k=1, 2, \ldots $:
\begin{equation}
	\label{eqn:state_dynamics model}
	\begin{split}
		x_{k} & \sim p(x_{k}|x_{k-1})\\
		x_{0} & \sim p(x_{0})
	\end{split}
\end{equation}

The \emph{observation model} describes how observations, $y_{k}$, are are related to the state at time step $k$:
\begin{equation}
	\label{eqn:observation_model}
	y_{k} \sim p(y_{k}|x_{k})
\end{equation}
As is conventional in this type of model, we assume first-order Markov state dynamics and conditional independence of observations:
\begin{equation}
	\label{eqn:bayes_filter_assumptions}
	\begin{split}
		p \left( x_{k} | x_{1:k-1}, y_{1:k-1} \right)
		& \, = \,
		p \left( x_{k} | x_{k-1} \right))\\
		p \left( y_{k} | x_{1:k}, y_{1:k-1} \right)
		& \, = \,
		p \left( y_{k} | x_{k} \right)
	\end{split}
\end{equation}

We infer the (conditional) PDF over states, $x_{k}$, using the following pair of recursive equations:
\begin{multline}
	\label{eqn:bayes_filter_predict}
	p(x_{k}|y_{1:k-1})\\
	\, = \,
	\int p(x_{k}|x_{k-1}) \, p(x_{k-1}|y_{1:k-1}) \, dx_{k-1}
\end{multline}
and
\begin{equation}	
	\label{eqn:bayes_filter_correct}
	p(x_{k}|y_{1:k})
	\, = \,
	\frac{1}{Z_{k}} \,
	p(y_{k}|x_{k}) \, p(x_{k}|y_{1:k-1})
\end{equation}
where the normalisation coefficient $Z_{k}$ is given by:
\begin{equation}
	\label{eqn:bayes_filter_correct_normaliser}
	Z_{k}
	\, = \,
	\int p(y_{k}|x_{k}) \, p(x_{k}|y_{1:k-1}) \, dx_{k}
\end{equation}

The first equation (Equation~\ref{eqn:bayes_filter_predict}) is known as the \emph{prediction step} or the \emph{Chapman-Kolmogorov} equation and it computes an interim PDF over states $x_{k}$ at time step $k$, given observations up to and including time step $k-1$. Since the integral marginalises over the state distribution at the previous time step, it can be viewed as computing the PDF of the \emph{probabilistic transformation} of the previous state using the state dynamics model. The second equation (Equation~\ref{eqn:bayes_filter_correct}) is known as the \emph{correction step} and it uses Bayes' rule to compute the PDF over states at time step $k$, given observations up to and including time step $k$. This step can be viewed as \emph{probabilistic fusion} of the current observation with the interim state computed in the prediction step (Equation~\ref{eqn:bayes_filter_predict}).

This type of state-space model and inference equations form the basis of many Bayesian filtering algorithms, including the Kalman filter (\cite{kalman1960new, kalman1961new}), extended Kalman filter (EKF) (see \cite{gelb1974applied}), unscented Kalman filter (UKF) (\cite{julier1995new, julier1997new}) and particle filters (see \cite{sarkka2013bayesian}).

The standard observation model defined in Equation~\ref{eqn:observation_model} is a \emph{generative} model because it specifies how to generate observations, $y_{k}$, given a state, $x_{k}$. However, as pointed out in \cite{burkhart2020discriminative}, we do not always have access to such a model but instead have a \emph{discriminative} model of the form $x_{k} \sim p(x_{k}|y_{k})$. This alternative type of model corresponds to the situation we are dealing with here, where the GDN model estimates a PDF over states (poses), given an observation (sensor image). To use this type of model in the Bayesian filter equations, we must first invert the original observation model using a second application of Bayes' rule and then substitute the result back in the original correction step(Equation~\ref{eqn:bayes_filter_correct}) to give a modified correction step:
\begin{equation}
	\label{eqn:bayes_filter_prob_fusion}
	p(x_{k}|y_{1:k})
	\, = \,
	\frac{1}{Z_{k}^{\prime}} \,
	\frac{p(x_{k}|y_{k}) \, p(x_{k}|y_{1:k-1})}{p(x_{k})}
\end{equation}

Here, the $p(y_{k})$ term has been absorbed in the modified normalisation constant $Z_{k}^{\prime}$. If we also assume a constant, (i.e., uninformative and possibly improper) prior, $p(x_{k})$, we can further simplify this modified correction step to a normalised product of PDFs:
\begin{equation}
	\label{eqn:bayes_filter_prob_fusion_approx}
	p(x_{k}|y_{1:k})
	\, = \,
	\frac{1}{Z_{k}^{\prime\prime}} \,
	p(x_{k}|y_{k}) \, p(x_{k}|y_{1:k-1})
\end{equation}
where the constant prior, $p(x_{k})$ has been absorbed in the modified normalisation constant $Z_{k}^{\prime \prime}$.

A similar approach was used to derive a pair of discriminative variations of the Kalman filter, referred to as the Discriminative Kalman Filter (DKF) and robust DKF (\cite{burkhart2020discriminative}). However, in that work the authors modified the inference equations \emph{after} specialising the state-space model to a linear-Gaussian model. We were not able to follow that approach here because we are not using Gaussian PDFs to represent the state distributions. It is nevertheless reassuring to know that if we had assumed a linear-Gaussian model with our more general equations (Equation~\ref{eqn:bayes_filter_predict} together with Equation~\ref{eqn:bayes_filter_prob_fusion} or Equation~\ref{eqn:bayes_filter_prob_fusion_approx}), we would obtain the same filter equations as these authors obtained in their work (see Appendix~\ref{app:probabilistic_transform_fusion}).

Having modified the correction step of the Bayesian filter to use a discriminative observation model, we now show how the prediction step (probabilistic transformation) and correction step (probabilistic fusion) can be implemented using $SE(3)$ PDFs of the form described in Section~\ref{sec:math_prelim}.

\subsubsection{Probabilistic transformation in SE(3).}
\label{sec:se3_probabilistic_transform}

It is well-known that the PDF of a linearly-transformed Gaussian random variable with added Gaussian noise is itself Gaussian, and its parameters can be computed analytically without having to explicitly evaluate a marginalisation integral like the one in Equation~\ref{eqn:bayes_filter_predict} (see Appendix~\ref{app:probabilistic_transform}). So, if we were using a linear-Gaussian state dynamics model with our Bayesian filter we could efficiently compute the prediction step specified in Equation~\ref{eqn:bayes_filter_predict} in closed form. In this section, we will derive an analogous simplification for $SE(3)$ random variables of the form discussed in Section~\ref{sec:math_prelim}.

We begin by assuming we are dealing with random variables (surface contact poses) of the form:
\begin{equation}
	\label{eqn:se3_random_variable_2}
	\boldsymbol{\mathrm{X}}
	\, = \,
	\exp({\boldsymbol\epsilon}^{\wedge}) \bar{\boldsymbol{\mathrm{X}}}
\end{equation}
where $\boldsymbol{\epsilon} \sim \mathcal{N}(\boldsymbol{\mathrm{0}}, \,\boldsymbol{\Sigma})$ and $ \bar{\boldsymbol{\mathrm{X}}} \in SE(3)$. We first transform the random variable, $\boldsymbol{\mathrm{X}}$, by composing it on the left with a deterministic $SE(3)$ transformation, $\boldsymbol{\mathrm{T}}$, to get another random variable of the same form (\cite{barfoot2014associating, barfoot2017state}):
\begin{equation}
	\label{eqn:se3_rv_deterministic_transform}
	\boldsymbol{\mathrm{T}} \boldsymbol{\mathrm{X}}
	\, = \,
	\boldsymbol{\mathrm{T}}
	\exp \left( \boldsymbol\epsilon^{\wedge} \right) \bar{\boldsymbol{\mathrm{X}}}
	\, = \,
	\exp \left( \left( \boldsymbol{\mathcal{T}} \boldsymbol\epsilon \right)^{\wedge} \right) \boldsymbol{\mathrm{T}} \bar{\boldsymbol{\mathrm{X}}}
\end{equation}

Here, $\boldsymbol{\mathcal{T}} = \mathrm{Ad}(\boldsymbol{\mathrm{T}})$ is the adjoint representation of $\boldsymbol{\mathrm{T}}$. We complete the probabilistic transformation to the new random variable, $\boldsymbol{\mathrm{X}}^{\prime}$, by adding some Gaussian noise, $\boldsymbol{\phi} \sim \mathcal{N}(\boldsymbol{\mathrm{0}}, \, \boldsymbol{\Sigma}_{\boldsymbol{\phi}})$ to the transformed perturbation, $\boldsymbol{\mathcal{T}} \boldsymbol\epsilon$, in the global tangent space:
\begin{equation}
	\label{eqn:se3_rv_probabilistic_transform}
	\boldsymbol{\mathrm{X}}^{\prime}
	\, = \,
	\exp \left( \left( \boldsymbol{\mathcal{T}} \boldsymbol\epsilon + \boldsymbol\phi \right)^{\wedge} \right) \boldsymbol{\mathrm{T}} \bar{\boldsymbol{\mathrm{X}}}
\end{equation}

The added Gaussian noise represents uncertainty in the transformation in much the same way as additive Gaussian noise represents uncertainty in the linear-Gaussian model. Notice that the transformed random variable, $\boldsymbol{\mathrm{X}}^{\prime}$, also has the same form as the original random variable, $\boldsymbol{\mathrm{X}}$:
\begin{equation}
	\label{eqn:se3_rv_probabilistic_transform_2}
	\boldsymbol{\mathrm{X}}^{\prime}
	\, = \,
	\exp \left( {\boldsymbol\epsilon}^{\prime \wedge} \right) \bar{\boldsymbol{\mathrm{X}}}^{\prime}
\end{equation}
where the \emph{nominal component}, $\bar{\boldsymbol{\mathrm{X}}}^{\prime}$, and \emph{perturbation component}, $\boldsymbol\epsilon^{\prime}$, of the transformation are given by:
\begin{equation}
	\label{eqn:se3_rv_prob_transform_components}
	\begin{gathered}
		\bar{\boldsymbol{\mathrm{X}}}^{\prime}
		\, = \,
		\boldsymbol{\mathrm{T}} \bar{\boldsymbol{\mathrm{X}}}\\
		\boldsymbol\epsilon^{\prime}
		\, = \,
		\boldsymbol{\mathcal{T}} \, \boldsymbol\epsilon + \boldsymbol\phi
	\end{gathered}
\end{equation}

Since $\boldsymbol{\epsilon}^{\prime} = \boldsymbol{\mathcal{T}} \boldsymbol\epsilon + \boldsymbol\phi$
represents a linear transformation of a Gaussian random variable, $\boldsymbol\epsilon$, with added Gaussian noise, $\boldsymbol\phi$, it also has a Gaussian distribution of the form (see Appendix~\ref{app:probabilistic_transform}):
\begin{equation}
	\label{eqn:se3_prob_transform_perturbation_pdf}
	{\boldsymbol{\epsilon}}^{\prime} \sim \mathcal{N}(\boldsymbol{\mathrm{0}}, \, \boldsymbol{\mathcal{T}} \boldsymbol{\Sigma} \boldsymbol{\mathcal{T}}^{\top} + \boldsymbol{\Sigma}_{\boldsymbol{\phi}})
\end{equation}

When, the random perturbations, $\boldsymbol\epsilon$ and $\boldsymbol\phi$, are both small and the transformation, $\boldsymbol{\mathrm{T}}$, is also small so that $\boldsymbol{\mathrm{T}} \approx \boldsymbol{\mathrm{1}}$, our probabilistic transformation (Equation~\ref{eqn:se3_rv_probabilistic_transform}) approximates the action of a deterministic $SE(3)$ transformation, $\boldsymbol{\mathrm{T}}$, with a zero-mean Gaussian noise perturbation, $\boldsymbol{\phi} \sim \mathcal{N}(\boldsymbol{\mathrm{0}}, \, \boldsymbol{\Sigma}_{\boldsymbol{\phi}})$, applied as a left-perturbation in the global tangent space (i.e., the action of a random $SE(3)$ transformation, $\boldsymbol{\mathrm{T}}^{\prime} = \exp ( {\boldsymbol\phi}^{\wedge} ) \boldsymbol{\mathrm{T}}$ on the $SE(3)$ random variable, $\boldsymbol{\mathrm{X}}$):
\begin{equation}
	\label{eqn:se3_probabilistic_transform_approx}
    \begin{split}
	\boldsymbol{\mathrm{X}}^{\prime}
	& \, = \,
	\exp \left( \left( \boldsymbol{\mathcal{T}} \boldsymbol\epsilon + \boldsymbol\phi \right)^{\wedge} \right) \boldsymbol{\mathrm{T}} \bar{\boldsymbol{\mathrm{X}}}\\
	& \, \approx \,
	\exp \left( {\boldsymbol\phi}^{\wedge} \right)
	\exp \left( \left( \boldsymbol{\mathcal{T}} \boldsymbol\epsilon \right)^{\wedge} \right) \boldsymbol{\mathrm{T}} \bar{\boldsymbol{\mathrm{X}}}\\
	& \, = \,
	\exp \left( {\boldsymbol\phi}^{\wedge} \right) \boldsymbol{\mathrm{T}} \boldsymbol{\mathrm{X}}\\
	& \, = \,
	\boldsymbol{\mathrm{T}}^{\prime} \, \boldsymbol{\mathrm{X}}\\		
    \end{split}
\end{equation}

Here, we have used the approximation, $ 
\exp \left( {\boldsymbol\xi_{1}^{\wedge}} \right) \exp \left( {\boldsymbol\xi_{2}^{\wedge}} \right) \, \approx \, \exp \left( \left( {\boldsymbol\xi_{1} + \boldsymbol\xi_{2}} \right)^{\wedge} \right)$, which applies when $\boldsymbol\xi_{1}$ and $\boldsymbol\xi_{2}$ are both small. Hence, our probabilistic transformation, $\boldsymbol{\mathrm{X}} \rightarrow \boldsymbol{\mathrm{X}}^{\prime}$, is (approximately) analogous to the linear-Gaussian case where a Gaussian random vector is multiplied by a deterministic linear transformation matrix, and then perturbed by adding a zero-mean Gaussian noise vector that represents uncertainty in the transformation.

So, for $SE(3)$ state variables (poses), we can use Equations~\ref{eqn:se3_rv_probabilistic_transform_2}-\ref{eqn:se3_prob_transform_perturbation_pdf} to efficiently compute the probabilistic transformation specified in the prediction step of our Bayesian filter without having to resort to expensive Monte Carlo integration.

\subsubsection{Probabilistic data fusion in SE(3).}
\label{sec:se3_probabilistic_fusion}

In the correction step of our discriminative Bayesian filter, we need to fuse two PDFs by computing their normalised product (Equation~\ref{eqn:bayes_filter_prob_fusion_approx}). For the multivariate Gaussian case, we could do this using the well-known expression for computing the normalised product of two Gaussian PDFs (see Appendix~\ref{app:probabilistic_fusion}). However, in $SE(3)$, it is not so easy because the PDFs are defined over a curved manifold instead of a Euclidean vector space. So, we need to adopt an iterative approach to solve the problem (\cite{barfoot2014associating, bourmaud2016intrinsic, smith2003computing}).

We start by assuming that our state (pose) PDFs are Lie group PDFs of the form discussed in Section~\ref{sec:math_prelim}:
\begin{equation}
	\label{eqn:se3_pdf_2}
	p \left( \boldsymbol{\mathrm{X}} \right)
	\, = \,
	\beta \left( \boldsymbol\epsilon \right)
	\exp \left( -\frac{1}{2}
	\boldsymbol\epsilon^{\top}
	\boldsymbol{\mathrm{\Sigma}}^{-1}
	\boldsymbol\epsilon \right)
\end{equation}
where $ \boldsymbol\epsilon = \ln \left(\boldsymbol{\mathrm{X}} \bar{\boldsymbol{\mathrm{X}}}^{-1} \right)^{\vee}$ and $\beta \left( \boldsymbol\epsilon \right) = \frac {\eta} {\left| \mathrm{det}( \mathcal{J}(\boldsymbol\epsilon) ) \right|}$. From Equation~\ref{eqn:left_jacobian_def}, $\mathcal{J}(\boldsymbol\epsilon)
\, = \, \boldsymbol{\mathrm{1}} + \frac{1}{2}\epsilon^{\curlywedge} + \dots
\approx \boldsymbol{\mathrm{1}}$ if the perturbation covariance matrix is sufficiently "small" (i.e., the maximum eigenvalue is sufficiently small), and if this is the case most of the probability mass will be concentrated around the mean and we can approximate the PDF as (\cite{bourmaud2016intrinsic}):
\begin{equation}
	\label{eqn:se3_pdf_approx}
	p \left( \boldsymbol{\mathrm{X}} \right)
	\approx
	\eta \exp \left( -\frac{1}{2}
	\ln \left(\boldsymbol{\mathrm{X}} \bar{\boldsymbol{\mathrm{X}}}^{-1} \right)^{\vee\top}
	\boldsymbol{\mathrm{\Sigma}}^{-1}
	\ln \left(\boldsymbol{\mathrm{X}} \bar{\boldsymbol{\mathrm{X}}}^{-1} \right)^{\vee} \right)
\end{equation}

Here, we have replaced the variable normalisation factor $\beta \left( \boldsymbol\epsilon \right)$ with the constant normalisation factor $\eta$. This \emph{concentrated Gaussian on a Lie group} assumption is made explicitly in \cite{bourmaud2015estimation} and \cite{bourmaud2016intrinsic}, and is implicit in the data fusion algorithm described in \cite{barfoot2014associating} and \cite{barfoot2017state}. In the latter case, the assumption is implied by the authors' choice of Mahalanobis cost function that they minimise to solve the data fusion problem. The reason this assumption is required for probabilistic fusion in $SE(3)$ is that a normalised product of Lie group PDFs is not in general equal to another Lie group PDF (in contrast to the normalised product of Gaussian PDFs in a Euclidean vector space). In fact, the normalised product of Lie group PDFs can only be \emph{approximated} by another PDF of the same form, but this approximation improves as the distributions become more concentrated.

Using the concentrated Gaussian on a Lie group assumption, we approximate the normalised product of two $SE(3)$ PDFs as:
\begin{multline}
\label{eqn:se3_normalised_prod}
	p_{*} \left( \boldsymbol{\mathrm{X}} \right)
	\, = \,
	\alpha p_{1} \left( \boldsymbol{\mathrm{X}} \right) p_{2} \left( \boldsymbol{\mathrm{X}} \right)\\
	\approx
	\alpha \eta_{1} \eta_{2}
	\exp 
	\left(
	-\frac{1}{2}
	\ln \left(\boldsymbol{\mathrm{X}} \bar{\boldsymbol{\mathrm{X}}}_{1}^{-1} \right)^{\vee\top}
	\boldsymbol{\mathrm{\Sigma}}_{1}^{-1}
	\ln \left(\boldsymbol{\mathrm{X}} \bar{\boldsymbol{\mathrm{X}}}_{1}^{-1} \right)^{\vee}
	\right)\\
	\times
	\exp 
	\left(
	-\frac{1}{2}
	\ln \left(\boldsymbol{\mathrm{X}} \bar{\boldsymbol{\mathrm{X}}}_{2}^{-1} \right)^{\vee\top}
	\boldsymbol{\mathrm{\Sigma}}_{2}^{-1}
	\ln \left(\boldsymbol{\mathrm{X}} \bar{\boldsymbol{\mathrm{X}}}_{2}^{-1} \right)^{\vee}
	\right)
\end{multline}
where $\alpha$ is a normalisation factor that ensures that the product PDF integrates to $1$ over its support. We now make a change of (random) variable, $\boldsymbol{\mathrm{X}} \rightarrow \boldsymbol\epsilon$, around an operating point $\bar{\boldsymbol{\mathrm{X}}}$, using:
\begin{equation}
	\begin{gathered}
		\label{eqn:se3_norm_prod_var_change}
		\boldsymbol{\mathrm{X}}
		\, = \,
		\exp(\boldsymbol\epsilon^{\wedge}) \, \bar{\boldsymbol{\mathrm{X}}}\\
		\boldsymbol\epsilon
		\, = \,
		\ln \left(\boldsymbol{\mathrm{X}} \bar{\boldsymbol{\mathrm{X}}}^{-1} \right)^{\vee}
	\end{gathered}
 \end{equation}
 Under this change of variables, the term, $\boldsymbol{\mathrm{X}} \bar{\boldsymbol{\mathrm{X}}}_{k}^{-1}, \; k=1,2$, becomes:
\begin{equation}
	\label{eqn:se3_norm_prod_factor_error}
	\boldsymbol{\mathrm{X}} \bar{\boldsymbol{\mathrm{X}}}_{k}^{-1}
	\, = \,
	\exp(\boldsymbol\epsilon^{\wedge}) \, \bar{\boldsymbol{\mathrm{X}}} \bar{\boldsymbol{\mathrm{X}}}_{k}^{-1}
	\, = \,
	\exp(\boldsymbol\epsilon^{\wedge}) \, \exp(\boldsymbol\xi_{k}^{\wedge})
\end{equation}
where $\boldsymbol\xi_{k} = \ln \left(\bar{\boldsymbol{\mathrm{X}}} \bar{\boldsymbol{\mathrm{X}}}_{k}^{-1} \right)^{\vee}$. Noting that $\boldsymbol\epsilon$ is small because of the concentrated Gaussian on a Lie group assumption, we can use the BCH approximation (Equation~\ref{eqn:bch_approx}) to write:
\begin{multline}
	\label{eqn:se3_norm_prod_factor_error_approx}
	\ln \left( \boldsymbol{\mathrm{X}} \bar{\boldsymbol{\mathrm{X}}}_{k}^{-1} \right)^{\vee}
	\, \approx \,
	\mathcal{J} \left( \boldsymbol\xi_{k} \right)^{-1} \boldsymbol\epsilon + \boldsymbol\xi_{k}\\
	\, = \,
	\mathcal{J} \left( \boldsymbol\xi_{k} \right)^{-1} \left( \boldsymbol\epsilon + \mathcal{J} \left( \boldsymbol\xi_{k} \right) \boldsymbol\xi_{k} \right)
\end{multline}

Substituting this expression in Equation~\ref{eqn:se3_normalised_prod} and rearranging terms, we get:
\begin{multline}
	\label{eqn:se3_normalised_prod_approx}
	p_{*} \left( \boldsymbol{\mathrm{X}} \right)
	\approx
	\alpha \eta_{1} \eta_{2}
	\exp
	\left(
	-\frac{1}{2}
	\left(
	\boldsymbol\epsilon - \boldsymbol\mu_{1}^{\prime} \right)^{\top}
	{\boldsymbol{\mathrm{\Sigma}}}_{1}^{\prime -1}
	\left( \boldsymbol\epsilon - \boldsymbol\mu_{1}^{\prime} \right)
	\right)\\
	\times
	\exp
	\left(
	-\frac{1}{2}
	\left(
	\boldsymbol\epsilon - \boldsymbol\mu_{2}^{\prime} \right)^{\top}
	{\boldsymbol{\mathrm{\Sigma}}}_{2}^{\prime -1}
	\left( \boldsymbol\epsilon - \boldsymbol\mu_{2}^{\prime} \right)
	\right)
\end{multline}
where, for $k=1,2$:
\begin{equation}
\label{eqn:se3_norm_prod_approx_factor_params}
	\begin{gathered}
		\boldsymbol\mu_{k}^{\prime} = -\mathcal{J}\left( \boldsymbol\xi_{k} \right) \boldsymbol\xi_{k}\\
		\boldsymbol{\mathrm{\Sigma}}_{k}^{\prime}
		\, = \,
		\mathcal{J} \left( \boldsymbol\xi_{k} \right)
		\boldsymbol{\mathrm{\Sigma}}_{k}
		\mathcal{J} \left( \boldsymbol\xi_{k} \right)^{\top}
	\end{gathered}
\end{equation}

For variables $\boldsymbol{\mathrm{X}}$ and $\boldsymbol\epsilon$, the infinitesimal volume elements, $d \boldsymbol{\mathrm{X}}$ and $d \boldsymbol\epsilon$, are related by $d \boldsymbol{\mathrm{X}} = \left| \mathrm{det}( \mathcal{J}(\boldsymbol\epsilon) ) \right| \, d \boldsymbol\epsilon$ (Equation~\ref{eqn:se3_inf_volume_element}), and using the concentrated Gaussian on a Lie group assumption, $\left| \mathrm{det}( \mathcal{J}(\boldsymbol\epsilon) ) \right| \approx \boldsymbol{\mathrm{1}} $. Hence, we can complete the change of variables in Equation~\ref{eqn:se3_normalised_prod_approx} to obtain:
\begin{multline}
	\label{eqn:se3_normalised_prod_approx_2}
	p_{*} \left( \boldsymbol\epsilon \right)
	\approx
	\alpha \eta_{1} \eta_{2}
	\exp
	\left(
	-\frac{1}{2}
	\left(
	\boldsymbol\epsilon - \boldsymbol\mu_{1}^{\prime} \right)^{\top}
	{\boldsymbol{\mathrm{\Sigma}}}_{1}^{\prime -1}
	\left( \boldsymbol\epsilon - \boldsymbol\mu_{1}^{\prime} \right)
	\right)\\
	\times
	\exp
	\left(
	-\frac{1}{2}
	\left(
	\boldsymbol\epsilon - \boldsymbol\mu_{2}^{\prime} \right)^{\top}
	{\boldsymbol{\mathrm{\Sigma}}}_{2}^{\prime -1}
	\left( \boldsymbol\epsilon - \boldsymbol\mu_{2}^{\prime} \right)
	\right)
\end{multline}

This is immediately recognizable as a normalized product of Gaussian PDFs, which is itself equal to a Gaussian PDF (\cite{bromiley2003products, petersen2008matrix}):
\begin{equation}
\label{eqn:se3_normalised_prod_approx_3}
	p_{*} \left( \boldsymbol\epsilon \right)
	\approx
	\eta_{*}
	\exp
	\left(
	-\frac{1}{2}
	\left(
	\boldsymbol\epsilon - \boldsymbol\mu_{*}^{\prime} \right)^{\top}
	{\boldsymbol{\mathrm{\Sigma}}}_{*}^{\prime -1}
	\left( \boldsymbol\epsilon - \boldsymbol\mu_{*}^{\prime} \right)
	\right)
\end{equation}
where the covariance, mean and normalisation factor are given by
\begin{equation}
\label{eqn:se3_normalised_prod_params}
	\begin{gathered}
		{\boldsymbol{\mathrm{\Sigma}}}_{*}^{\prime}
		=
		\left(
		{\boldsymbol{\mathrm{\Sigma}}}_{1}^{\prime -1}
		+
		{\boldsymbol{\mathrm{\Sigma}}}_{2}^{\prime -1}
		\right)^{-1}\\
		{\boldsymbol{\mathrm{\mu}}}_{*}^{\prime}
		=
		{\boldsymbol{\mathrm{\Sigma}}}_{*}^{\prime}
		\left(
		{\boldsymbol{\mathrm{\Sigma}}}_{1}^{\prime -1} {\boldsymbol{\mathrm{\mu}}}_{1}^{\prime}
		+
		{\boldsymbol{\mathrm{\Sigma}}}_{2}^{\prime -1} {\boldsymbol{\mathrm{\mu}}}_{2}^{\prime}
		\right)\\
		\eta_{*} = \alpha \eta_{1} \eta_{2} \gamma
	\end{gathered}
\end{equation}

This normalized product of Gaussian PDFs represents a \emph{non-zero-mean} Gaussian perturbation in the global tangent space, of the mean value, $\bar{\boldsymbol{\mathrm{X}}} \in SE(3)$. However, since our $SE(3)$ random variables require zero-mean Gaussian perturbations, we must now convert this PDF to one that represents a \emph{zero-mean} Gaussian perturbation in the global tangent space, of some other mean value $\bar{\boldsymbol{\mathrm{X}}}_{*} \in SE(3)$. We do this by reversing the steps we used to make the initial change of variables, starting by substituting the following analogous expressions to those of Equation~\ref{eqn:se3_norm_prod_approx_factor_params}, into Equation~\ref{eqn:se3_normalised_prod_approx_3}:
\begin{equation}
\label{eqn:se3_norm_prod_params_rev}
	\begin{gathered}
		\boldsymbol\mu_{*}^{\prime} = -\mathcal{J}\left( \boldsymbol\xi_{*} \right) \boldsymbol\xi_{*}\\
		\boldsymbol{\mathrm{\Sigma}}_{*}^{\prime}
		\, = \,
		\mathcal{J} \left( \boldsymbol\xi_{*} \right)
		\boldsymbol{\mathrm{\Sigma}}_{*}
		\mathcal{J} \left( \boldsymbol\xi_{*} \right)^{\top}
	\end{gathered}
\end{equation}
where
\begin{equation}
	\label{eqn:se3_norm_prod_error_rev}
	\boldsymbol\xi_{*}
	\, = \,
	\ln \left(\bar{\boldsymbol{\mathrm{X}}} \bar{\boldsymbol{\mathrm{X}}}_{*}^{-1} \right)^{\vee}
\end{equation}

Under the concentrated Gaussian on a Lie group assumption, the infinitesimal volume elements are related by $d \boldsymbol{\mathrm{X}} \approx d \boldsymbol\epsilon$, and so using an analogous BCH approximation to Equation~\ref{eqn:se3_norm_prod_factor_error_approx},
\begin{multline}
	\label{eqn:se3_norm_prod_factor_error_approx_rev}
	\ln \left( \boldsymbol{\mathrm{X}} \bar{\boldsymbol{\mathrm{X}}}_{*}^{-1} \right)^{\vee}
	\, \approx \,
	\mathcal{J} \left( \boldsymbol\xi_{*} \right)^{-1} \boldsymbol\epsilon + \boldsymbol\xi_{*}\\
	=
	\mathcal{J} \left( \boldsymbol\xi_{*} \right)^{-1} \left( \boldsymbol\epsilon + \mathcal{J} \left( \boldsymbol\xi_{*} \right) \boldsymbol\xi_{*} \right)
\end{multline}
we obtain the following approximation for the normalised product PDF in $SE(3)$:
\begin{equation}
	\label{eqn:se3_normalised_prod_rev}
	p_{*} \left( \boldsymbol{\mathrm{X}} \right)
	\approx
	\eta_{*}
	\exp 
	\left(
	-\frac{1}{2}
	\ln \left(\boldsymbol{\mathrm{X}} \bar{\boldsymbol{\mathrm{X}}}_{*}^{-1} \right)^{\vee\top}
	\boldsymbol{\mathrm{\Sigma}}_{*}^{-1}
	\ln \left(\boldsymbol{\mathrm{X}} \bar{\boldsymbol{\mathrm{X}}}_{*}^{-1} \right)^{\vee}
	\right)
\end{equation}
where, from Equations~\ref{eqn:se3_norm_prod_params_rev} and \ref{eqn:se3_norm_prod_error_rev}, we obtain:
\begin{equation}
	\label{eqn:se3_norm_prod_rev_params}
	\begin{gathered}
		\boldsymbol{\mathrm{\Sigma}}_{*}
		\, = \,
		\mathcal{J} \left( \boldsymbol\xi_{*} \right)^{-1}
		\boldsymbol{\mathrm{\Sigma}}_{*}^{\prime}
		\mathcal{J} \left( \boldsymbol\xi_{*} \right)^{-\top}\\
		\bar{\boldsymbol{\mathrm{X}}}_{*}
		\, = \,
		\exp \left( \left( -\boldsymbol\xi_{*} \right)^{\wedge} \right) \bar{\boldsymbol{\mathrm{X}}}
	\end{gathered}
\end{equation}

Since we cannot compute $\bar{\boldsymbol{\mathrm{X}}}_{*}$ in closed form because this would involve finding the solution to a non-linear equation in $\boldsymbol\xi_{*}$ and $\boldsymbol\mu_{*}^{\prime}$ (Equation~\ref{eqn:se3_norm_prod_params_rev}), we instead approximate the Jacobian in this equation using $\mathcal{J} \left( \boldsymbol\xi_{*} \right) \approx \boldsymbol{\mathrm{1}}$ (which becomes more accurate as $\boldsymbol\xi_{*}$ becomes smaller) and then substitute for $\boldsymbol\xi_{*}$ in Equation~\ref{eqn:se3_norm_prod_rev_params} to obtain an approximate solution:
\begin{equation}
\label{eqn:se3_norm_prod_rev_params_approx}
	\begin{gathered}
		\bar{\boldsymbol{\mathrm{X}}}_{*}
		\, \approx \,
		\exp \left( \boldsymbol\mu_{*}^{\prime \wedge} \right) \bar{\boldsymbol{\mathrm{X}}}\\
		\boldsymbol{\mathrm{\Sigma}}_{*}
		\, \approx \,
		\boldsymbol{\mathrm{\Sigma}}_{*}^{\prime}
	\end{gathered}
\end{equation}

We solve for $\boldsymbol\mu_{*}^{\prime}$ by updating the operating point, $\bar{\boldsymbol{\mathrm{X}}}$, to make it equal to our approximation, $\bar{\boldsymbol{\mathrm{X}}}_{*}$, and then iterate to convergence:
\begin{equation}
	\label{eqn:se3_normalised_prod_update}
	\bar{\boldsymbol{\mathrm{X}}}
	\, \leftarrow \,
	\exp \left( \boldsymbol\mu_{*}^{\prime \wedge} \right) \bar{\boldsymbol{\mathrm{X}}}
\end{equation}

At convergence, $\boldsymbol\mu_{*}^{\prime} \rightarrow \boldsymbol{\mathrm{0}}$, and we obtain the mean and covariance parameters of the normalised product of $SE(3)$ PDFs from Equation~\ref{eqn:se3_norm_prod_rev_params_approx}, as $\bar{\boldsymbol{\mathrm{X}}}_{*} \, = \,
 \bar{\boldsymbol{\mathrm{X}}}$ and $\boldsymbol{\mathrm{\Sigma}}_{*} \, = \, \boldsymbol{\mathrm{\Sigma}}_{*}^{\prime}$.

From a qualitative perspective, this method works because perturbing the operating point $\bar{\boldsymbol{\mathrm{X}}}$ on the left by $\boldsymbol\mu_{*}^{\prime} $ reduces the magnitude of the error between the operating point and the solution $\bar{\boldsymbol{\mathrm{X}}}_{*}$ (even if there is some degree of undershoot or overshoot at each iteration). This, in turn, increases the accuracy of the approximation, $\mathcal{J} \left( \boldsymbol\xi_{*} \right) \approx \boldsymbol{\mathrm{1}}$, which decreases the error at the next iteration, and so on. This continual reduction in error proceeds until the solution converges, and hence it can be viewed as a fixed-point iteration method.

We combine the various steps described above into an iterative algorithm (Algorithm~\ref{alg:se3_data_fusion}), which we have found from experience typically converges in 3-4 steps (consistent with the findings presented in \cite{barfoot2014associating}. Since we are operating in a real-time context where we want our code to run with predictable timing, we chose to perform a fixed number of iterations (5 iterations) rather than run the algorithm until convergence. To simplify the implementation, we expanded the solution in Equation~\ref{eqn:se3_normalised_prod_params} in terms of the original factor PDF means and covariances, and in terms of the inverse Jacobian rather than the Jacobian. We compute the inverse Jacobian up to second-order terms in its series expansion (Equation~\ref{eqn:inv_jacobian_def}), which we have found is sufficiently accurate in this context (again, this is consistent with findings presented in \cite{barfoot2014associating}). We also initialised the operating point (i.e., the interim solution), $\bar{\boldsymbol{\mathrm{X}}}$, to equal the mean of the first (left-hand) factor PDF of the product.

\begin{algorithm}
	\caption{$SE(3)$ data fusion}
	\label{alg:se3_data_fusion}
	\begin{algorithmic}
	\Require $SE(3)$ factor PDF parameters $(\bar{\boldsymbol{\mathrm{X}}}_{k}, \; \boldsymbol{\mathrm{\Sigma}}_{k})$; $k=1, 2$
	\Ensure Normalised product PDF parameters $(\bar{\boldsymbol{\mathrm{X}}}_{*}, \; \boldsymbol{\mathrm{\Sigma}}_{*})$
	\State \textit{Initialise trial solution (operating point):}
	\State $\bar{\boldsymbol{\mathrm{X}}} = \bar{\boldsymbol{\mathrm{X}}}_{1}$
	\While {not converged}
		\State $\boldsymbol\xi_{k} = \ln \left(\bar{\boldsymbol{\mathrm{X}}} \bar{\boldsymbol{\mathrm{X}}}_{k}^{-1} \right)^{\vee}; k=1, 2$
		\State $\mathcal{J}_{k}^{-1} \equiv \mathcal{J}(\boldsymbol\xi_{k})^{-1}
	\, \approx \, \boldsymbol{\mathrm{1}} - \frac{1}{2}\boldsymbol\xi_{k}^{\curlywedge} + \frac{1}{12}(\boldsymbol\xi_{k}^{\curlywedge})^{2}; k=1, 2$
		\State ${\boldsymbol{\mathrm{\Sigma}}}_{*}^{\prime}
			=
			\left(
			\mathcal{J}_{1}^{-\top}
			\boldsymbol{\mathrm{\Sigma}}_{1}^{-1}
			\mathcal{J}_{1}^{-1}
			+
			\mathcal{J}_{2}^{-\top}
			\boldsymbol{\mathrm{\Sigma}}_{2}^{-1}
			\mathcal{J}_{2}^{-1}
			\right)^{-1}$
		\State $\boldsymbol{\mathrm{\mu}}_{*}^{\prime}
			=
			-\boldsymbol{\mathrm{\Sigma}}_{*}^{\prime}
			\left(
			\mathcal{J}_{1}^{-\top}
			\boldsymbol{\mathrm{\Sigma}}_{1}^{-1} \boldsymbol{\mathrm{\xi}}_{1}
			+
			\mathcal{J}_{2}^{-\top}
			\boldsymbol{\mathrm{\Sigma}}_{2}^{-1} \boldsymbol{\mathrm{\xi}}_{2}
			\right)$
		\State \textit{Update trial solution (operating point):}
		\State $\bar{\boldsymbol{\mathrm{X}}}
		\, \leftarrow \,
		\exp \left( \boldsymbol\mu_{*}^{\prime \wedge} \right) \bar{\boldsymbol{\mathrm{X}}}$
	\EndWhile
	\State \textit{Return solution at convergence:}
	\State $(\bar{\boldsymbol{\mathrm{X}}}_{*}, {\boldsymbol{\mathrm{\Sigma}}}_{*}) = (\bar{\boldsymbol{\mathrm{X}}}, {\boldsymbol{\mathrm{\Sigma}}}_{*}^{\prime} )$
	\end{algorithmic}
\end{algorithm}

\begin{figure*}
	\centering
	\includegraphics[width=\textwidth]{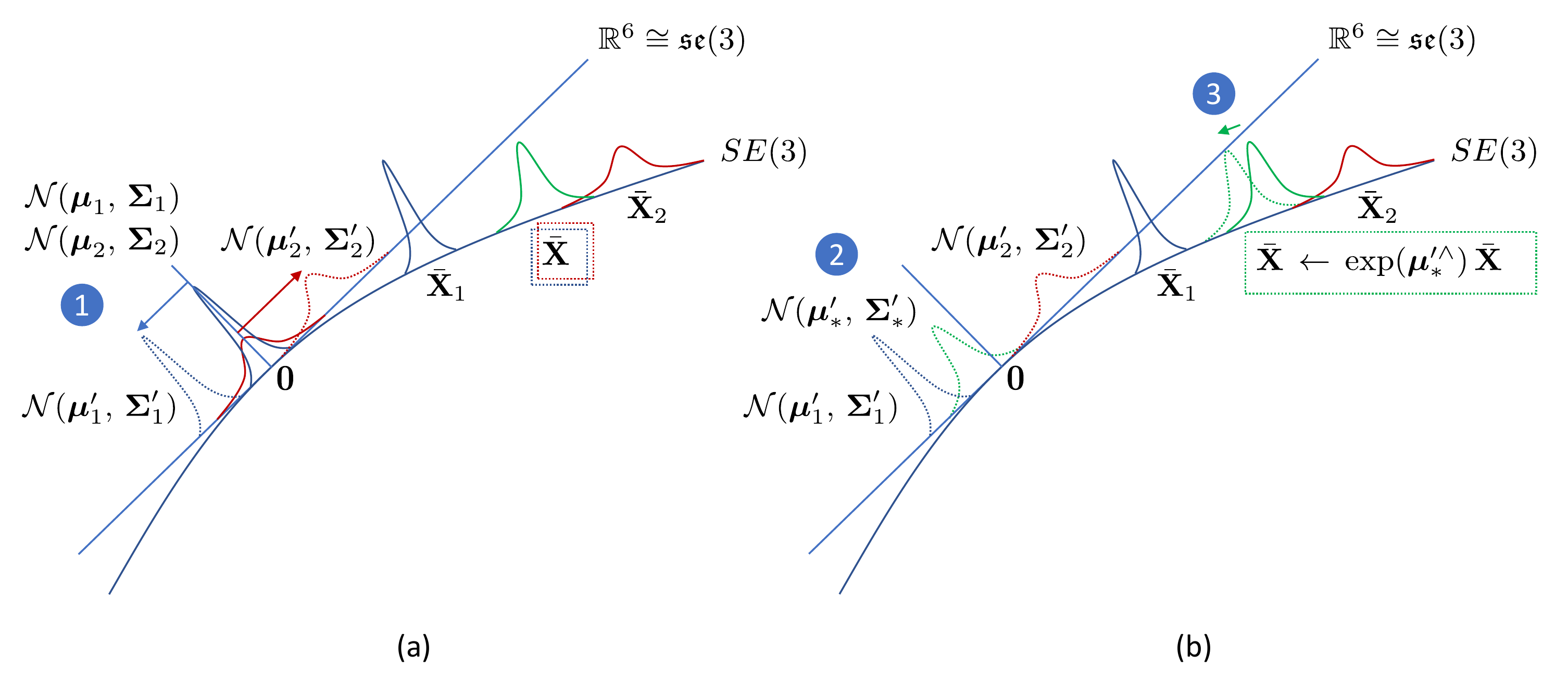}
	\caption{Visual depiction of a single iteration of our $SE(3)$ data fusion algorithm. (a) Step 1: Transform zero-mean Gaussian perturbations of two different $SE(3)$ means to non-zero-mean Gaussian perturbations of a single, common $SE(3)$ mean. (b) Step 2: Fuse the transformed non-zero-mean PDFs by computing their normalised product in the global tangent space; Step 3: Zero the mean of the fused Gaussian distribution by perturbing the associated (common) $SE(3)$ mean.
	\label{fig:data_fusion_visualisation}}
\end{figure*}

We visually depict a single iteration of our data fusion algorithm in Figure~\ref{fig:data_fusion_visualisation}. The iteration begins with the two factor PDFs of the normalised product, with $SE(3)$ means, $\bar{\boldsymbol{\mathrm{X}}}_{1}$ and $\bar{\boldsymbol{\mathrm{X}}}_{2}$, and zero-mean Gaussian left-perturbations in the global tangent space, $\boldsymbol\epsilon_{1} \sim \mathcal{N}(\boldsymbol{\mathrm{0}}, \boldsymbol\Sigma_{1})$ and $\boldsymbol\epsilon_{2} \sim \mathcal{N}(\boldsymbol{\mathrm{0}}, \boldsymbol\Sigma_{2})$. Under the change of variables (Equation~\ref{eqn:se3_norm_prod_var_change}), these PDFs are transformed so that they have the same $SE(3)$ mean, $\bar{\boldsymbol{\mathrm{X}}}$, but different Gaussian perturbations with modified non-zero means and covariances: $\boldsymbol\epsilon_{1}^{\prime} \sim \mathcal{N}(\boldsymbol\mu_{1}^{\prime}, \boldsymbol\Sigma_{1}^{\prime})$ and $\boldsymbol\epsilon_{2}^{\prime} \sim \mathcal{N}(\boldsymbol\mu_{2}^{\prime}, \boldsymbol\Sigma_{2}^{\prime})$ in the global tangent space. These two PDFs are then combined and replaced by a single non-zero-mean Gaussian PDF (the scaled product), $\boldsymbol\epsilon_{*}^{\prime} \sim \mathcal{N}(\boldsymbol\mu_{*}^{\prime}, \boldsymbol\Sigma_{*}^{\prime})$. Finally, this single combined Gaussian PDF is converted back to the required zero-mean form, by perturbing the corresponding $SE(3)$ mean, $\bar{\boldsymbol{\mathrm{X}}}$, on the left by the mean of the Gaussian perturbation, which approximately returns it to zero: $\bar{\boldsymbol{\mathrm{X}}} \, \leftarrow \, \exp \left( \boldsymbol\mu_{*}^{\prime \wedge} \right) \bar{\boldsymbol{\mathrm{X}}}$. The cycle is then repeated until convergence.

While we recognise that our data fusion algorithm is essentially identical to the algorithm proposed in \cite{barfoot2014associating}, our method of deriving it is very different. In our case, we follow an algebraic approach, where the iteration arises in finding an approximate solution to a non-linear equation. This contrasts with the original derivation, which is based on iterative Gauss-Newton minimization of a Mahalanobis cost function. We believe that our derivation offers some new insights that are not accessible using the original derivation. For example, in \cite{barfoot2014associating}, the authors only refer to their method qualitatively as a "data fusion" method, but do not show that it finds a solution to the normalised product of $SE(3)$ PDFs, which is essential for implementing the correction step of our $SE(3)$ discriminative Bayesian filter (Equation~\ref{eqn:bayes_filter_prob_fusion_approx}). 

\subsubsection{SE(3) Bayesian filtering algorithm.}
\label{sec:se3_discriminative_bayes_filter}

During each iteration of our $SE(3)$ discriminative Bayesian filter (Algorithm~\ref{alg:se3_bayes_filter}), we use the $SE(3)$ probabilistic transformation described in Section~\ref{sec:se3_probabilistic_transform} (Equations~\ref{eqn:se3_rv_prob_transform_components} and \ref{eqn:se3_prob_transform_perturbation_pdf}) to compute the prediction step (Equation~\ref{eqn:bayes_filter_predict}), and the probabilistic fusion algorithm described in Section~\ref{sec:se3_probabilistic_fusion} (Algorithm~\ref{alg:se3_data_fusion}) to compute the normalised product of $SE(3)$ PDFs in the correction step (Equation~\ref{eqn:bayes_filter_prob_fusion_approx}).

In Algorithm~\ref{alg:se3_bayes_filter}, the deterministic component, $\boldsymbol{\mathrm{T}}_{k} \in SE(3)$, of the probabilistic transformation represents the change in sensor pose between time steps $k-1$ and $k$, which approximates the change in object pose between these time steps: $\boldsymbol{\mathrm{T}}_{k} = (\boldsymbol{\mathrm{X}}_{k}^{(\mathrm{sens})})^{- 1} \boldsymbol{\mathrm{X}}_{k - 1}^{(\mathrm{sens})}$. The constant, zero-mean Gaussian noise term, $\boldsymbol\phi$ (with covariance, $\boldsymbol{\Sigma}_{\boldsymbol{\phi}}$) in Equation~\ref{eqn:se3_rv_prob_transform_components} represents the uncertainty in the change in object pose between time steps (e.g., due to motion of the object or surface relative to the sensor).

Unless specified otherwise, we use the following state dynamics noise covariance in the Bayesian filter algorithm for the experiments and demonstrations in this paper:
\begin{equation}
	\label{eqn:state_dynamics_noise_cov}
	\boldsymbol\Sigma_{\boldsymbol\phi}
	=
	\sigma_{\phi}^{2}
	\begin{bmatrix}
		1 & & & & \cdots & 0 \\
		& 1 & & & & \vdots \\
		& & 1 & & & \\
		& & & \left(\frac{\pi}{180}\right)^{2} & & \\
		\vdots & & & & \left(\frac{\pi}{180}\right)^{2} & \\
		0 & \cdots & & & & \left(\frac{\pi}{180}\right)^{2}
	\end{bmatrix}
\end{equation}
where $\sigma_{\phi}=0.5$.

\begin{algorithm}
	\caption{$SE(3)$ discriminative Bayesian filter}
	\label{alg:se3_bayes_filter}
	\begin{algorithmic}
	\Require A sequence of sensor-surface pose distribution estimates $(\bar{\boldsymbol{\mathrm{X}}}_{k}^{(\mathrm{obs})}, \boldsymbol\Sigma_{k}^{(\mathrm{obs})})$ and sensor poses $\boldsymbol{\mathrm{X}}_{k}^{(\mathrm{sens})}$; $k = 0, 1, 2, \ldots $
	\Ensure A sequence of filtered sensor-surface pose distribution estimates $(\bar{\boldsymbol{\mathrm{X}}}_{k}^{(\mathrm{fil})}, \boldsymbol\Sigma_{k}^{(\mathrm{fil})})$; $k = 0, 1, 2, \ldots $
	\State \textit{Initialisation:} $p(x_{0}^{(\mathrm{fil})}) = p(x_{0}^{(\mathrm{bel})}) = p(x_{0}^{(\mathrm{obs})})$
	\State\hspace{\algorithmicindent} $(\bar{\boldsymbol{\mathrm{X}}}_{0}^{(\mathrm{fil})}, \boldsymbol\Sigma_{0}^{(\mathrm{fil})}) = (\bar{\boldsymbol{\mathrm{X}}}_{0}^{(\mathrm{bel})}, \boldsymbol\Sigma_{0}^{(\mathrm{bel})}) = (\bar{\boldsymbol{\mathrm{X}}}_{0}^{(\mathrm{obs})}, \boldsymbol\Sigma_{0}^{(\mathrm{obs})})$
	\For{each $(\bar{\boldsymbol{\mathrm{X}}}_{k}^{(\mathrm{obs})}, \boldsymbol\Sigma_{k}^{(\mathrm{obs})})$}
		\State \textit{Prediction step:}
		\State \parbox[t]{\dimexpr\linewidth-\algorithmicindent}{%
		Compute $p(x_{k}^{(\mathrm{bel})}) = \int p(x_{k}^{(\mathrm{bel})} | x_{k - 1}^{(\mathrm{fil})}) p(x_{k - 1}^{(\mathrm{fil})}) dx_{k - 1}^{(\mathrm{fil})}$ using:
		}
		\State\hspace{\algorithmicindent} $\bar{\boldsymbol{\mathrm{X}}}_{k}^{(\mathrm{bel})} = \boldsymbol{\mathrm{T}}_{k} \bar{\boldsymbol{\mathrm{X}}}_{k - 1}^{(\mathrm{fil})}$
		\State\hspace{\algorithmicindent} $\boldsymbol\Sigma_{k}^{(\mathrm{bel})} = \boldsymbol{\mathcal{T}}_{k} \boldsymbol\Sigma_{k - 1}^{(\mathrm{fil})}  \boldsymbol{\mathcal{T}}_{k}^{\top} + \boldsymbol{\Sigma}_{\boldsymbol{\phi}}$
		\State\hspace{\algorithmicindent} where $\boldsymbol{\mathcal{T}}_{k} = \mathrm{Ad}(\boldsymbol{\mathrm{T}}_{k})$
		\State\hspace{\algorithmicindent} and $\boldsymbol{\mathrm{T}}_{k} = (\boldsymbol{\mathrm{X}}_{k}^{(\mathrm{sens})})^{- 1} \boldsymbol{\mathrm{X}}_{k - 1}^{(\mathrm{sens})}$
		\State \textit{Correction step:}
		\State \parbox[t]{\dimexpr\linewidth-\algorithmicindent}{%
	  	Compute $p(x_{k}^{(\mathrm{fil})}) = \frac{1}{Z}p(x_{k}^{(\mathrm{obs})}) \, p(x_{k}^{(\mathrm{bel})})$
	  	using the $SE(3)$ data fusion algorithm (Algorithm~\ref{alg:se3_data_fusion}) to find
		}
			\State $(\bar{\boldsymbol{\mathrm{X}}}_{k}^{(\mathrm{fil})}, \boldsymbol\Sigma_{k}^{(\mathrm{fil})})$ from $(\bar{\boldsymbol{\mathrm{X}}}_{k}^{(\mathrm{obs})}, \boldsymbol\Sigma_{k}^{(\mathrm{obs})}), (\bar{\boldsymbol{\mathrm{X}}}_{k}^{(\mathrm{bel})}, \boldsymbol\Sigma_{k}^{(\mathrm{bel})})$
	\EndFor
	\end{algorithmic}
\end{algorithm}

\subsection{Pose and shear-based control systems: Feedforward-feedback control in SE(3)}
\label{sec:se3_feedforward_feedback_control}

In this section, we describe the feedforward-feedback controllers we use for servoing and manipulation tasks such as object tracking, surface following and object pushing. These controllers are similar to ones we have used in the past, but for this project we have modified them to produce velocity-based control signals so that the robot arms move with a smooth continuous motion. Another important difference is that we now frame their operation in terms of errors and control signals computed in local or global tangent spaces, which as we will explain later in this section places them on a more principled footing than before. This is also a more natural perspective to adopt for velocity-based control because the tangent spaces contain the velocity twist objects used to represent the control signals.

\subsubsection{Feedforward-feedback control in local and global tangent spaces.}
\label{sec:tangent_space_control}

In our previous work, we made extensive use of feedback control systems. A feedback control system computes the error between an observed system variable (process variable) and a reference value (set point) and uses this error to compute a control signal that is fed back to the system in order to reduce the error over time. Since our observed system variables are typically poses in $SE(3)$, two immediate problems arise. Firstly, how can we compute the error in $SE(3)$, given that we cannot subtract $SE(3)$ objects in the way we do for scalars and vectors in a Euclidean space? Secondly, how can we use this $SE(3)$ error to compute a control signal in a meaningful way? For example, when using state feedback control in a Euclidean vector space, we usually compute a control signal as a linear transform of the error. However, in general, a linear transform of an $SE(3)$ object is no longer a valid $SE(3)$ object unless the transform is itself an $SE(3)$ object (i.e., the transformed error does not lie on the Lie group manifold). Hence, we cannot compute a control signal as a simple scalar multiple of an $SE(3)$ error, as we might do in proportional control (a scalar multiple of the rotation sub-matrix is no longer a valid rotation sub-matrix, in general). So we need to use an alternative representation for poses where this sort of operation can be carried out in a meaningful way.

\begin{figure*}
	\centering
	\includegraphics[width=\textwidth]{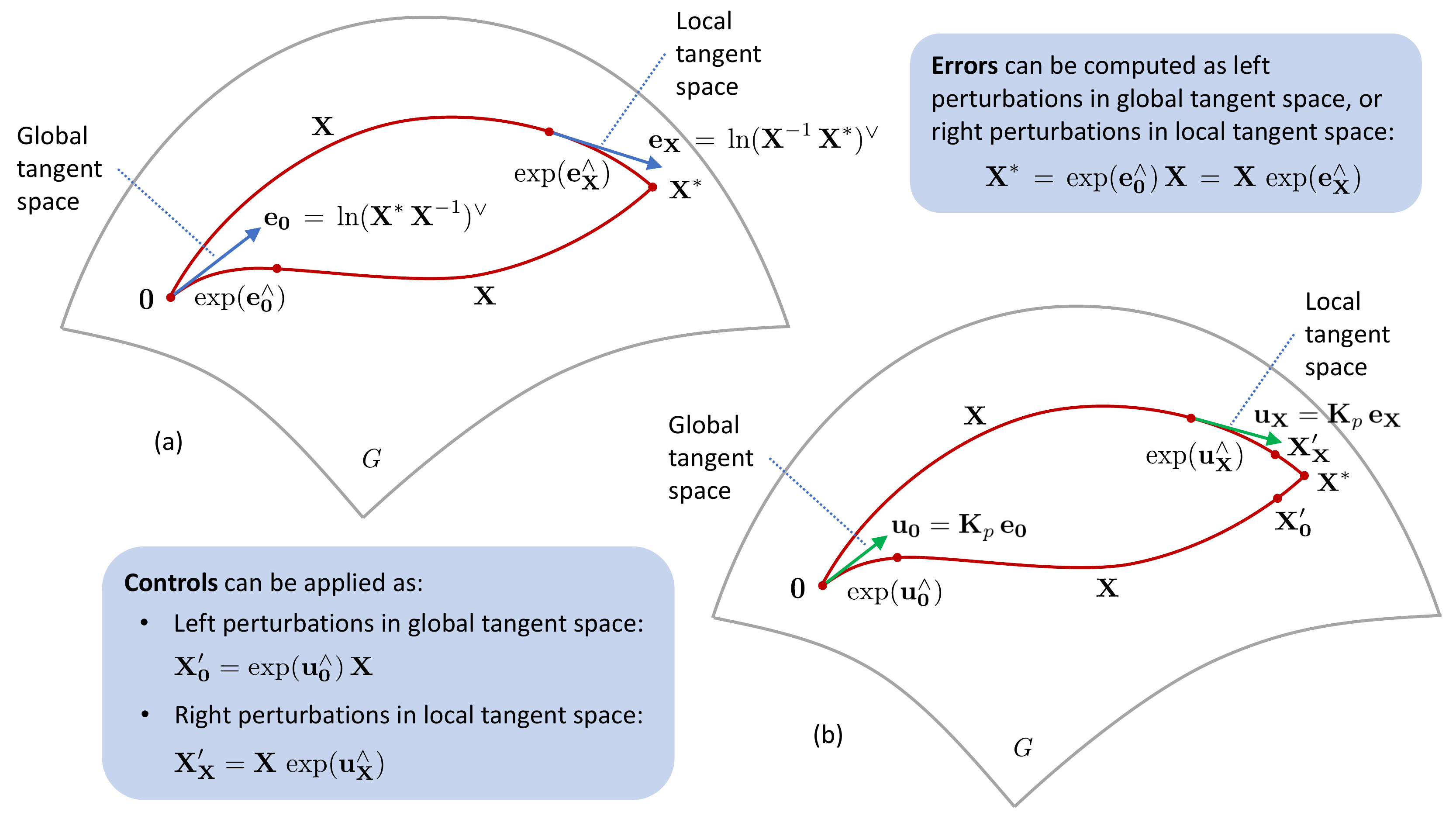}
	\caption{Projecting pose errors and computing control signals in local and global tangent spaces. (a) Computing pose errors in $SE(3)$ and projecting them into local and global tangent spaces. (b) Computing control signals in local and global tangent spaces, using proportional control as an example.
	\label{fig:tangent_space_control}}
\end{figure*}

We address the first problem of how to compute a meaningful error in $SE(3)$ by defining an equivalent (and meaningful) "subtraction" operation in terms of the group composition and inverse operations. This subtraction operation can be defined in the local (body) coordinate frame or the global (fixed) frame. In the \emph{local} coordinate frame, we define the error between the observed value $\boldsymbol{\mathrm{X}}$ (e.g., a pose) and the reference value $\boldsymbol{\mathrm{X}}^{*}$ as:
\begin{equation}
	\label{eqn:local_coord_frame_error}
	\boldsymbol{\mathrm{E}}_{\boldsymbol{\mathrm{X}}}
	\, = \,
	\boldsymbol{\mathrm{X}}^{-1} \, \boldsymbol{\mathrm{X}}^{*}
\end{equation}

This error can be viewed as the $SE(3)$ transformation that transforms $\boldsymbol{\mathrm{X}}$ to $\boldsymbol{\mathrm{X}}^{*}$, when expressed in the local coordinate frame. In the global coordinate frame, we define the error as:
\begin{equation}
	\label{eqn:global_coord_frame_error}
	\boldsymbol{\mathrm{E}}_{\boldsymbol{\mathrm{0}}}
	\, = \,
	\boldsymbol{\mathrm{X}}^{*} \, \boldsymbol{\mathrm{X}}^{-1}
\end{equation}

Similarly, this is the $SE(3)$ transformation that transforms $\boldsymbol{\mathrm{X}}$ to $\boldsymbol{\mathrm{X}}^{*}$, when expressed in the global coordinate frame.

We address the second problem of how to represent $SE(3)$ errors so that the control operations can be defined in a meaningful way, by projecting the errors into the corresponding local or global tangent spaces (which space we use depends on the coordinate frame the errors have been defined in) and computing the control signals in those spaces (Figure~\ref{fig:tangent_space_control}). Since these spaces are Euclidean vector spaces, we can employ all of the control frameworks that have been developed for such spaces (e.g., state feedback). Another advantage of using these representations is that for velocity-based control we can use the control signals generated in these spaces to directly control the robot (depending on the API). For position-based control, we simply map the tangent space control signals back to $SE(3)$ using the exponential map.

Hence, in the local tangent space, we project the corresponding local $SE(3)$ error as:
\begin{equation}
	\label{eqn:local_tangent_space_error}
	\boldsymbol{\mathrm{e}}_{\boldsymbol{\mathrm{X}}}
	\, = \,
	\ln(\boldsymbol{\mathrm{X}}^{-1} \, \boldsymbol{\mathrm{X}}^{*})^{\vee}
\end{equation}

The corresponding error projection in the global tangent space is:
\begin{equation}
	\label{eqn:global_tangent_space_error}
	\boldsymbol{\mathrm{e}}_{\boldsymbol{\mathrm{0}}}
	\, = \,
	\ln(\boldsymbol{\mathrm{X}}^{*} \, \boldsymbol{\mathrm{X}}^{-1})^{\vee}
\end{equation}

When we project the error into one or other tangent space in this way, we can view it as the right- or left-perturbation of the current $SE(3)$ pose, $\boldsymbol{\mathrm{X}}$, needed to transform it to the target reference value, $\boldsymbol{\mathrm{X}}^{*}$. In other words:
\begin{equation}
	\boldsymbol{\mathrm{X}}^{*}
	\, = \,
	\exp(\boldsymbol{\mathrm{e}}_{\boldsymbol{0}}^{\wedge}) \, \boldsymbol{\mathrm{X}}
	\, = \,
	\boldsymbol{\mathrm{X}} \, \exp(\boldsymbol{\mathrm{e}}_{\boldsymbol{\mathrm{X}}}^{\wedge})
\end{equation}

With the errors represented as tangent space perturbations, we can compute the control signals in a meaningful way using methods that are designed to operate in Euclidean vector spaces. We can then use the control signals to directly control the robot for velocity-based control, or treat them as right- or left-perturbation of the current $SE(3)$ pose for position-based control.

So, for example, in the case of MIMO proportional control in the local tangent space, we use Equation~\ref{eqn:local_tangent_space_error} to project the $SE(3)$ error into the tangent space as $\boldsymbol{\mathrm{e}}$, and then compute the control signal using:
\begin{equation}
	\label{eqn:proportional_control}
	\boldsymbol{\mathrm{u}}
	\, = \,
	\boldsymbol{\mathrm{K}}_{p} \, \boldsymbol{\mathrm{e}}
\end{equation}
where $\boldsymbol{\mathrm{K}}_{p}$ is a $6 \times 6$ diagonal gain matrix that contains the corresponding proportional gain coefficients. For full MIMO proportional-integral-derivative (PID) control we use:
\begin{equation}
	\label{eqn:pid_control}
	\boldsymbol{\mathrm{u}} = \boldsymbol{\mathrm{K}}_{p} \, \boldsymbol{\mathrm{e}}
	+ \boldsymbol{\mathrm{K}}_{i} \, \int_{0}^{t} \boldsymbol{\mathrm{e}} \, dt^{\prime}
	+ \boldsymbol{\mathrm{K}}_{d} \, \frac{d \boldsymbol{\mathrm{e}}}{dt}
\end{equation}
where $\boldsymbol{\mathrm{K}}_{i}$ and $\boldsymbol{\mathrm{K}}_{d}$ are the $6 \times 6$ diagonal gain matrices associated with the integral and derivative errors. For this type of controller, we often include a feedforward term, $\boldsymbol{\mathrm{v}}$, (specified in the relevant tangent space) which can generate a control signal in the absence of any error:
\begin{equation}
	\label{eqn:ff_fb_pid_control}
	\boldsymbol{\mathrm{u}}
	= \boldsymbol{\mathrm{v}}
	+ \boldsymbol{\mathrm{K}}_{p} \, \boldsymbol{\mathrm{e}}
	+ \boldsymbol{\mathrm{K}}_{i} \, \int_{0}^{t} \boldsymbol{\mathrm{e}} \, dt^{\prime}
	+ \boldsymbol{\mathrm{K}}_{d} \, \frac{d \boldsymbol{\mathrm{e}}}{dt}
\end{equation}

This can be useful in surface following tasks, where we want the robot end-effector to move tangentially to the surface while remaining normal to the surface at a fixed contact depth. The same is also true for object pushing tasks, where we want the robot end-effector to move the object towards the target position, while remaining normal to the contacted surface of the pushed object.

For the servoing and manipulation tasks we discuss in this paper, we use the local variation of these tangent space control methods, partly because our velocity-based robot API accepts control signals specified in the local tangent space, and partly because the feedforward signals we use for these tasks are more easily specified in the local tangent space. If we were to perform control in the global tangent space, we would need to map the feed-forward signal from the local to global tangent space using the adjoint representation $\mathrm{Ad}(\cdot)$ and then map the control signal back to the local tangent space before sending it to the robot. However, we have described both approaches here, for the sake of completeness and because in some situations it might be more appropriate to compute the control signals in the global tangent space.

Since our system operates in discrete time, we use simple backward-Euler approximations to compute the integral and derivative errors. To reduce noise in the error signal before computing the derivative, we smooth the error using an exponentially-weighted moving average (EWMA) filter with a decay coefficient of 0.5. We also clip the integral error between pre-defined limits to mitigate integral wind-up problems, and clip the output to limit the range of the control signal. Details of gain coefficients, error or output clipping ranges, feedback reference poses and feedforward reference velocities (velocity twists) are provided where we describe the controller configurations for specific tasks.

\subsubsection{Tactile servoing controller.}
\label{sec:servoing_controller}

\begin{figure*}
	\centering
	\includegraphics[width=\textwidth]{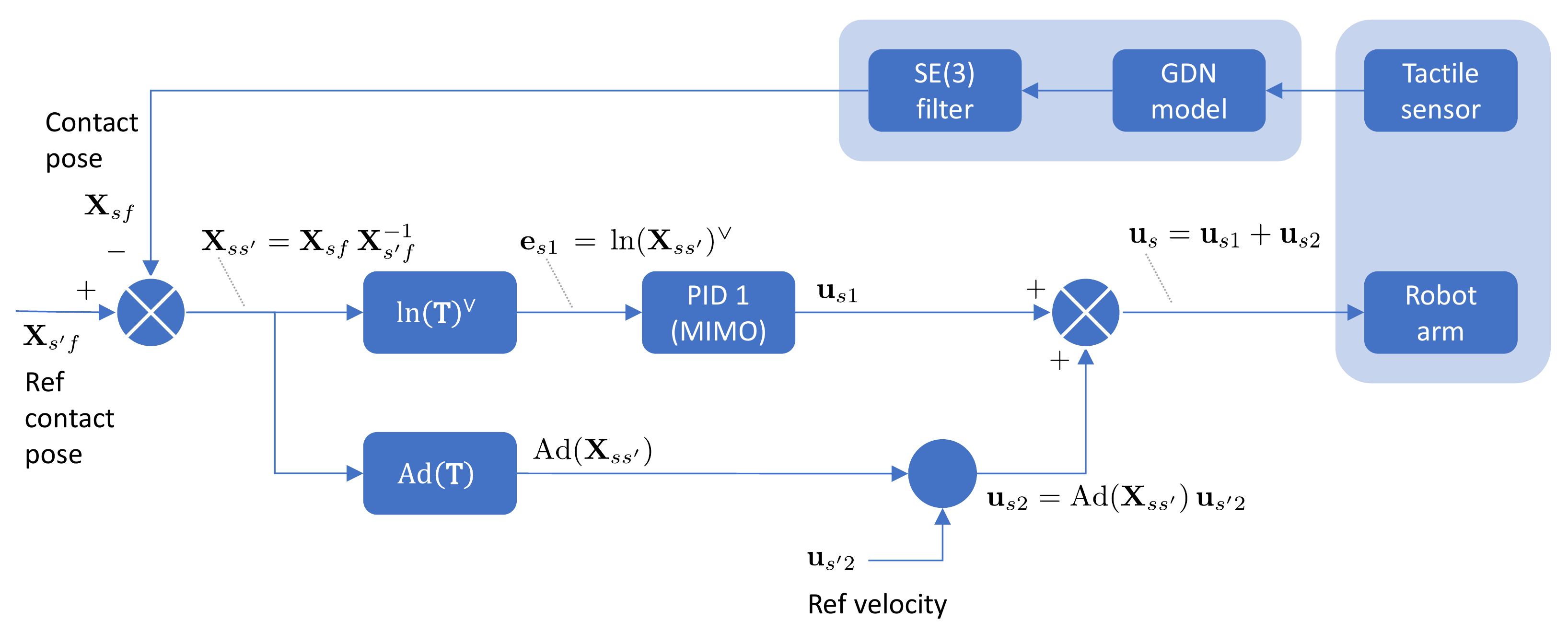}
	\caption{Tactile servoing controller used for object tracking and surface following tasks.
	\label{fig:servoing_controller}}
\end{figure*}

For object tracking and surface following, we use a tactile servoing controller (Figure~\ref{fig:servoing_controller}) that performs MIMO feedforward-feedback PID control in the local tangent space, as described in the previous section (see Equation~\ref{eqn:ff_fb_pid_control}). The goal of this controller is to align the sensor with a reference contact pose in the surface feature frame, while moving with a reference velocity in the reference pose frame. The reference contact pose is usually specified so that the sensor is normal to the surface or object, at a fixed contact depth. For object tracking, the reference velocity is  usually set to zero. For surface following, the reference velocity is specified so that the sensor moves tangentially to the normal contact (reference) pose. In Figure~\ref{fig:servoing_controller}, the observed and reference contact poses are specified in inverse form, $\boldsymbol{\mathrm{X}}_{sf}$ and $\boldsymbol{\mathrm{X}}_{s^{\prime}f}$, because this representation is more convenient for the previous filtering stage.

In each control cycle, we start by computing the $SE(3)$ error in the local (sensor) coordinate frame using:
\begin{equation}
	\label{eqn:se3_tactile_servo_error}
	\boldsymbol{\mathrm{X}}_{ss^{\prime}}
	\, = \,
	\boldsymbol{\mathrm{X}}_{sf} \, \boldsymbol{\mathrm{X}}_{s^{\prime}f}^{-1}
\end{equation}
where $\boldsymbol{\mathrm{X}}_{sf}$ is the observed feature pose (i.e., the surface contact pose) in the current sensor frame, which is predicted by the GDN model and subsequently filtered by the Bayesian filter; $\boldsymbol{\mathrm{X}}_{s^{\prime}f}$ is the target feature pose in the reference sensor frame. We then project this error into the local tangent space of the sensor frame using the logarithmic map (Equation~\ref{eqn:local_tangent_space_error}) and forward the projected error to a 6-channel MIMO PID controller. Finally, we add the PID control signal to the feedforward reference velocity. For surface following, we specify the reference sensor frame so that its $z$-axis is normal to and pointing towards the surface and the reference velocity so that it lies in the $xy$-plane of that frame (i.e., it is tangential to the surface). We use the adjoint representation of the $SE(3)$ error to map the reference velocity back into the tangent space associated with the observed sensor frame before adding it to the feedback signal because we cannot directly add signals that are specified in different tangent spaces. Finally, we use the resulting control signal to update the robot end-effector velocity during each control cycle.

\subsubsection{Tactile pushing controller.}
\label{sec:pushing_controller}

\begin{figure*}
	\centering
	\includegraphics[width=\textwidth]{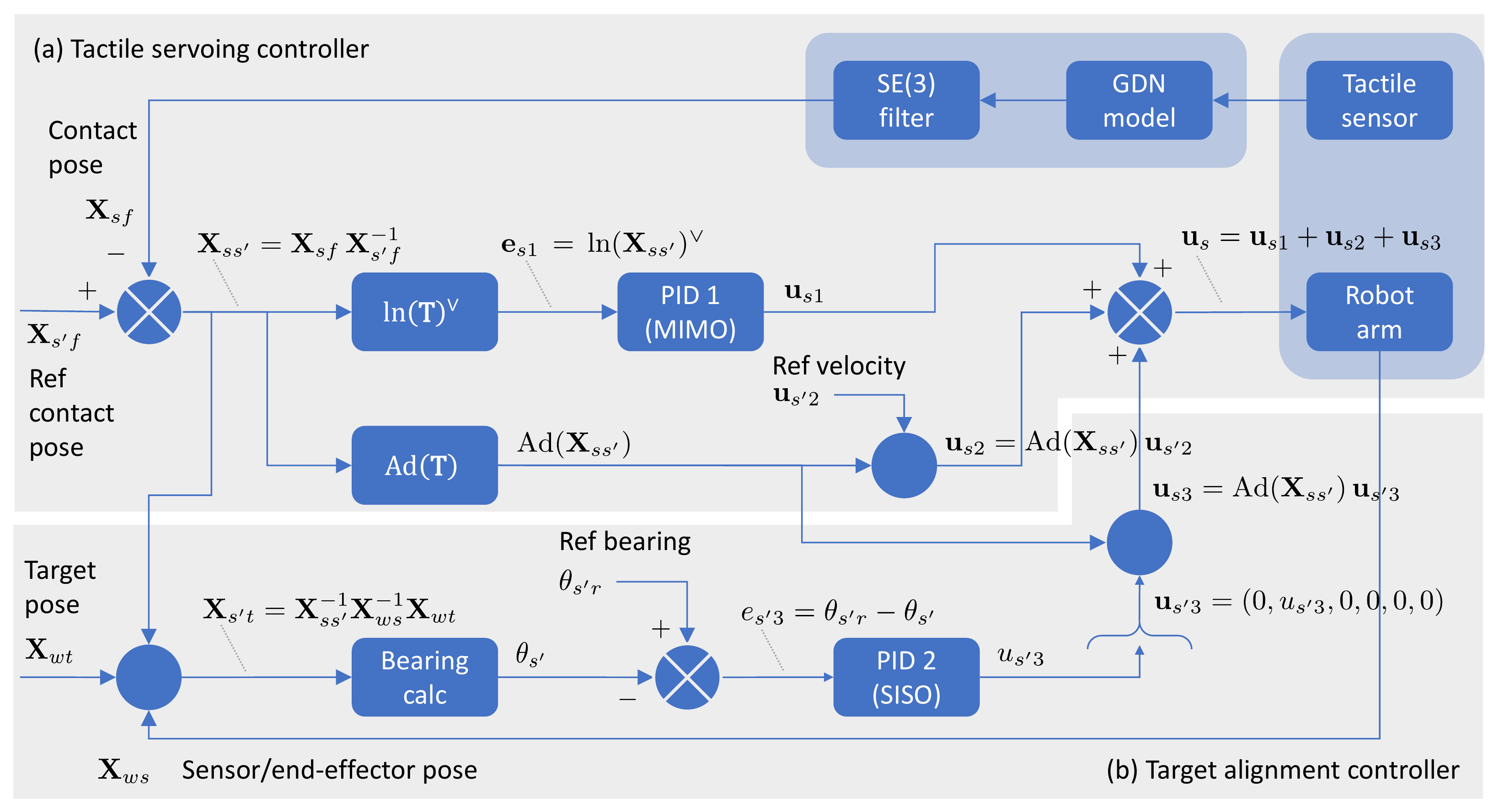}
	\caption{Tactile pushing controller used for single-arm and dual-arm object pushing, with its two main components: (a) tactile servoing controller, (b) target alignment controller.
	\label{fig:pushing_controller}}
\end{figure*}

For pushing objects across a surface towards a target, we augment the tactile servoing controller with an additional feedback control element that we refer to as the \emph{target alignment controller} (Figure~\ref{fig:pushing_controller}). The target alignment controller tries to steer the object towards the target as it is pushed forward, using sideways tangential movements while the sensor remains in frictional contact with the object. In this configuration, the controller reference velocity is specified normal to and into the object surface rather than tangential to the surface as it is for surface following. The combined effect of the tactile servoing and target alignment controllers is to get the sensor to push the object towards the target reference point while maintaining (approximate) normal contact with the surface of the pushed object. The control signal produced by the target alignment controller is applied tangentially to the sensor-object contact so that it applies a turning moment under frictional contact, steering the object one way or the other depending on the polarity of the signal. Since the tactile serviong controller has already been discussed in the previous section, we only describe its integration with the target alignment controller here.

The object pushing target is specified as a target pose, $\boldsymbol{\mathrm{X}}_{wt}$, in the robot work frame, $\{w\}$. The target pose is transformed to the reference sensor frame using the sensor pose, $\boldsymbol{\mathrm{X}}_{ws}$, obtained from the robot (i.e., proprioceptive information), and the sensor error, $\boldsymbol{\mathrm{X}}_{ss^{\prime}}$, computed by the tactile servoing controller in Equation~\ref{eqn:se3_tactile_servo_error}:
\begin{equation}
	\label{eqn:transformed_target_pose}
	\boldsymbol{\mathrm{X}}_{s^{\prime}t}
	=
	\boldsymbol{\mathrm{X}}_{ss^{\prime}}^{-1} \boldsymbol{\mathrm{X}}_{ws}^{-1} \boldsymbol{\mathrm{X}}_{wt}
\end{equation}
	
The target bearing and distance are then computed in the reference sensor frame using:
\begin{equation}
	\label{eqn:target_bearing_distance}
	\begin{gathered}
		\theta_{s^{\prime}} = \mathrm{atan2} \left( y, z \right)\\
		r_{s^{\prime}} = \sqrt{ y^{2} + z^{2} }
	\end{gathered}
\end{equation}
where $y$ and $z$ are the target pose translation components extracted from the matrix $\boldsymbol{\mathrm{X}}_{s^{\prime}t}$. The target bearing is then subtracted from the reference bearing, which is zero in our case, $\theta_{s^{\prime}r} = 0$, to obtain the bearing error in the reference sensor frame. This is forwarded to a single-input, single-output (SISO) PID controller, which generates a scalar control signal that overwrites the $y$-component of a zero, 6-vector, velocity control signal in the reference sensor frame. The $y$-component of the reference sensor frame lies parallel to the surface on which the object is pushed and tangential to the sensor-object contact between sensor and pushed object. Since the tactile servoing controller generates a control signal in the tangent space of the current sensor pose, the target alignment control signal must be transformed from the tangent space of the reference sensor frame to that associated with the current sensor frame. We do this using the adjoint representation of the $SE(3)$ error, in the same way that we transform the reference velocity signal in the tactile servoing controller. The transformed target alignment control signal is then added to the tactile servoing control signal. Finally, we use the resulting control signal to update the robot end-effector velocity during each control cycle. 

As in our previous work on pushing (\cite{lloyd2021goal}), we switch off the target alignment controller (i.e., zero its output) when the sensor is less than some pre-defined distance, $\rho^{*}$, away from the target to maintain stability close to the target. In this work, we used $\rho^{*}=120$ mm for our experimental demonstrations. After this point, only the tactile servoing controller remains active. The technical details of why we need to do this are discussed in Appendix~\ref{app:pushing_differential_analysis}. The pushing sequence is terminated when the centre of the sensor tip (radius = 20 mm) is closer than 20 mm from the target. This ensures that the sensor-object contact point is moved as close as possible to the target without moving too far past it.

\subsubsection{Single-arm and dual-arm control configurations.}
\label{sec:single_dual_arm_controllers}

The tactile servoing and object pushing controllers described in this section can either be used in isolation to control a single robot arm for object tracking, surface following or single-arm pushing tasks, or they can be used in combination to control multiple robot arms as demonstrated in our dual-arm pushing task. In the dual-arm pushing task, one arm is controlled by an active/leader pushing controller, while the second arm is controlled using a passive/follower object tracking controller. This dual-arm configuration allows the active pushing arm to control the movement of the object towards the target, while the second passive arm helps to stabilise the object along the way to prevent it from toppling over.

Another way of viewing the operation of these multi-arm configurations is that each robot arm is attempting to follow a control signal injected via the feedforward path, while simultaneously trying to satisfy the constraints imposed by the reference contact pose specified in the feedback path. In this scenario, the feedforward control signals can either be generated separately for each arm in a decentralised approach or they can be generated in a centralised, more coordinated manner. The "leader-follower" configuration we use in our dual-arm pushing task is an example of the decentralised approach.

\subsection{Tactile robot experimental platform}
\label{sec:experimental_platform}

In this section, we describe the robot arms, tactile sensors and software components of our tactile robotic system.

\subsubsection{Dual-arm robot platform.}
\label{sec:dual_arm_robot_platform}

\begin{figure*}
	\centering
	\includegraphics[width=\textwidth]{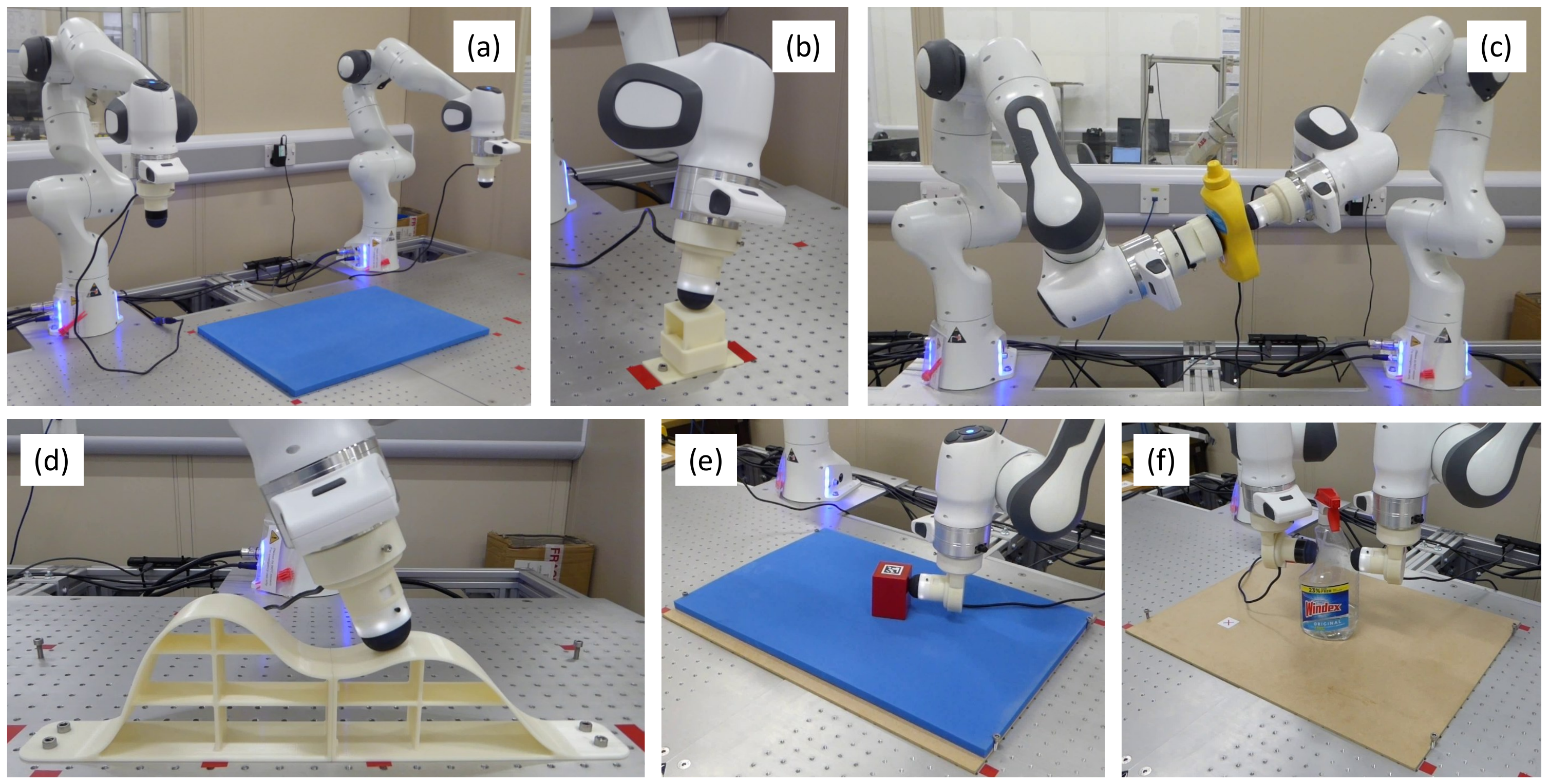}
	\caption{Tactile robotic experimental platform configured for different tasks: (a) Dual-arm robot system. (b) Data collection of surface contact poses. (c) Dual-arm object tracking. (d) Surface following. (e) Single-arm object pushing. (f) Dual-arm object pushing.
	\label{fig:exp_platform_configurations}}
\end{figure*}

For our experiments and demonstrations, we use a dual robot arm system with two Franka Emika Panda, 7 degree-of-freedom (DoF) robot arms. The robot arms are mounted on aluminium trolleys with base plates, which are bolted together so that the arms are separated by a distance of 1.0 m at the base and can be used individually or together for collaborative tasks (Figure~\ref{fig:exp_platform_configurations}). Depending on the task, the robots can either be fitted with a TacTip optical tactile sensor or a stimulus adaptor as an end-effector. Where a tactile sensor is fitted, this can be mounted in a standard, downwards-pointing configuration, or at right angles, using a special adaptor.

\subsubsection{Tactile sensor.}
\label{sec:tactile_sensor}

\begin{figure}
	\centering
	\includegraphics[width=\columnwidth]{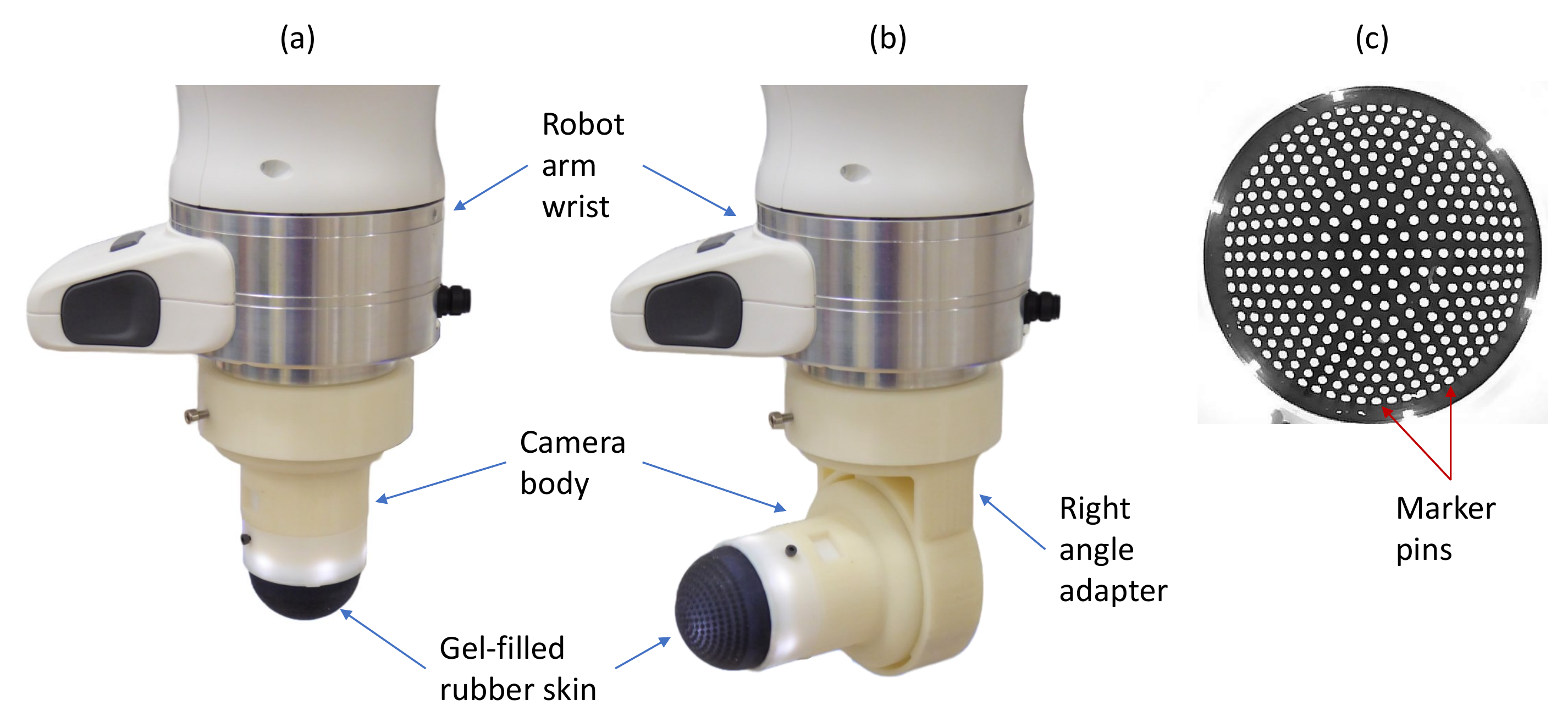}
	\caption{TacTip optical tactile sensor. (a) Standard sensor mount. (b) Right-angle sensor mount. (c) Sensor camera image showing marker pins.
	\label{fig:tactile_sensor}}
\end{figure}

The TacTip optical tactile sensor (Figure~\ref{fig:tactile_sensor}) was developed by researchers at Bristol Robotics Laboratory (BRL) in 2009 (\cite{chorley2009development}) and has since been used in a wide variety of robotic touch applications (\cite{ward2018tactip, lepora2021soft}). The 3D-printed sensor tip consists of a black rubber-like skin with an internal array of pins, which are capped with white markers. The tip is filled with an optically-clear gel that helps maintain its shape under contact. In this project, we used a TacTip sensor with 331 pins arranged in a circular array. The internal pins amplify any changes in the marker positions, caused by deformation of the external surface of the skin. The marker positions are recorded by a fast (up to 120 frames per second) USB camera, which is mounted inside the camera body. In this project, we used raw sensor images with minimal preprocessing as input to a CNN or GDN model, but it is also possible to detect the marker positions using standard image processing techniques (e.g., blob detection) and  use the marker coordinates as input to more conventional types of machine learning system.

\subsubsection{Software.}
\label{sec:software}

We control the robot arms using a layered software API, built on top of the \emph{libfranka} C++ library component (version 0.8.0) of the Franka Control Interface (\cite{franka2015fci}). This library provides several software "hooks", which allow users to specify callback functions that are called in a 1 kHz real-time control loop. On top of this, we have a developed an internal library, called \emph{pyfranka}, which provides smooth trajectory generation for position- and velocity-based control, so that velocity, acceleration and jerk constraints are not violated; it also handles any background threads needed for velocity-based control, and provides a python wrapper via \emph{pybind11}. The pyfranka library sits underneath the \emph{Common Robot Interface} (CRI) python library that we have used in most of our recent work within the group. Since the only critical functionality needed for our experiments and demonstrations is the ability to perform Cartesian (task space) position/velocity control and query the state of the robot, it should be possible to replace the Franka robot arms and API with any 6-DOF or 7-DOF robot arm that supports this functionality.

We use the OpenCV library (version 4.5.2, python interface) to capture and process the camera images produced by the tactile sensor, and TensorFlow (version 2.4) with the included Keras API to develop the neural network models we use to process the images produced by the sensor. We also use the \emph{transforms3d} python library and the software provided with Lynch and Park's book, Modern Robotics (\cite{lynch2017modern}), to manipulate 3D poses, transforms, and velocity twists.

We run all of the software components in a Pyro5 distributed object environment on an Ubuntu 18.04 desktop PC. The Pyro5 environment allows us to run several communicating python processes in parallel (and if necessary on separate PCs, although that was not required in this work), to help ensure that real-time performance constraints are met. Using this approach, we were able to run the low-level 1 kHz control loops, image capture, neural network inference (but not training) and high-level control loops for both robot arms and tactile sensors on a single PC. Since the Robot Operating System (ROS) provides a similar type of distributed processing environment to Pyro5, we think it should be relatively easy to implement our system using ROS, if desired.

\section{Experiments and demonstrations}
\label{sec:experiments_demos}

In this section, we describe and present results for the experiments and demonstrations we use to evaluate our tactile robotic system. We start by comparing the performance of our GDN pose estimation model against a baseline CNN model. We then evaluate the impact of our $SE(3)$ Bayesian filter on reducing the error and uncertainty associated with the GDN model estimates. Finally, we show how our tactile robotic system can be used to perform several useful tactile servoing and non-prehensile manipulation tasks, including object tracking, surface following, single-arm object pushing, and dual-arm object pushing.

\subsection{Neural network-based pose and shear estimation}
\label{sec:nn_pose_shear_estimate_exp}

In this experiment, we compare the performance of our GDN pose estimation model against a baseline CNN model. To ensure a fair comparison, both models were developed using the same three data sets (6000 training set samples, 2000 validation set samples and 2000 test set samples), which were collected as described in Section~\ref{sec:data_collection} and pre-processed as described in Section~\ref{sec:pre_post_processing}. We trained the CNN model as described in Section~\ref{sec:cnn_model}, and the GDN models as described in Section~\ref{sec:gdn_model}. To avoid statistical anomalies, we trained 10 CNN models and 10 GDN models from different random weight initializations and then computed the mean and standard deviation loss (MSE loss for the CNN model and mean NLL for the GDN model) and component MAEs for all models on the test data set (Table~\ref{tab:nn_pose_shear_estimate_results}).

\begin{table}[h]
	\small\sf\centering
	\caption{Overall MSE / mean NLL loss and pose component MAEs for 10 GDN and 10 CNN models (mean values $\pm$ standard deviation across 10 models). The lowest mean MAE values for each component are highlighted in bold.
	\label{tab:nn_pose_shear_estimate_results}}
	\begin{tabular}{ccc}
	\toprule
	Metric & CNN & GDN\\
	\midrule
	MSE/NLL loss & 1.4336 $\pm$ 0.0204 & -7.2035 $\pm$ 0.2536\\
	\midrule
	MAE $v_{x}$ (mm/s) & 0.4480 $\pm$ 0.0055 & \textbf{0.4259} $\pm$ 0.0054\\	
	MAE $v_{y}$ (mm/s) & 0.4514 $\pm$ 0.0038 & \textbf{0.4224} $\pm$ 0.0051\\	
	MAE $v_{z}$ (mm/s) & 0.1636 $\pm$ 0.0023 & \textbf{0.1230} $\pm$ 0.0015\\	
	MAE $\omega_{x}$ (rad/s) & 0.0153 $\pm$ 0.0005 & \textbf{0.0087} $\pm$ 0.0002\\	
	MAE $\omega_{y}$ (rad/s) & 0.0181 $\pm$ 0.0003 & \textbf{0.0111} $\pm$ 0.0002\\	
	MAE $\omega_{z}$ (rad/s) & 0.0252 $\pm$ 0.0004 & \textbf{0.0203} $\pm$ 0.0003\\
	\bottomrule
	\end{tabular}
\end{table}

\begin{figure*}
	\centering
	\includegraphics[width=\textwidth]{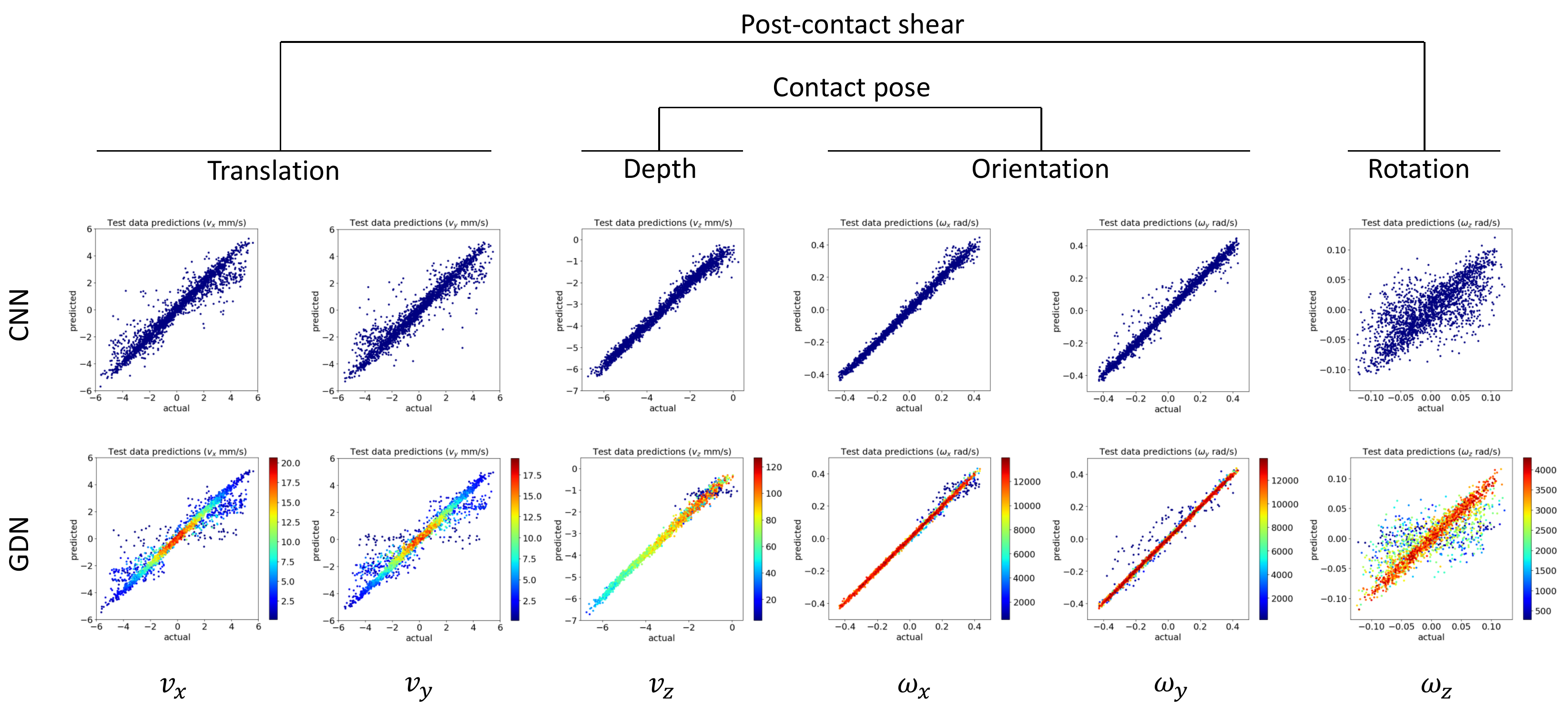}
	\caption{Distribution of test errors for best-performing (lowest loss) CNN and GDN models. Estimated (predicted) pose components are plotted against actual pose components for both models. GDN pose estimates are coloured according to their precision (i.e., reciprocal variance) estimated by the model (see attached colour bars).
	\label{fig:nn_pose_shear_estimate_performance}}
\end{figure*}

The results show that the GDN model produces lower component MAEs than the CNN model when evaluated on the test data set. We think this is because the mean NLL loss function used to train the GDN model directly incorporates the estimated uncertainty for each pose component, effectively increasing the error weighting for more confident estimates and decreasing the error weighting for less confident ones (see discussion in Section~\ref{sec:gdn_model}). This contrasts with the MSE loss function used to train the CNN model, which implicitly assumes a fixed uncertainty for each pose component (i.e., a standard deviation of $1$), and is therefore unable to adaptively weight the errors in the same way.

We visualise the distribution of test set errors by plotting the estimated pose components against the actual pose components (i.e., test set labels) for the best-performing model of each type (Figure~\ref{fig:nn_pose_shear_estimate_performance}. In the case of the GDN model, we also colour each point according to the precision (inverse variance) estimated by the model for the corresponding pose component. Points coloured in red denote high precision (low uncertainty) estimates and points coloured in blue denote low precision (high uncertainty) estimates. In the ideal case, all points would lie on an imaginary ground-truth line stretching from the bottom left corner of each plot to the top right corner and there would be no scattering about this line.

With reference to these plots, we make the following observations. Firstly, while the errors are quite large across all pose components estimated by both models, they are larger for the shear-related components than the normal contact components. We think this is probably due to aliasing effects, which are more prevalent during shear motion than normal contact motion. Secondly, the GDN estimates appear more accurate than the CNN estimates (the distribution of predicted values is more concentrated around the true values for the GDN model than the CNN model), which is consistent with the statistical results presented in Table~\ref{tab:nn_pose_shear_estimate_results}. Finally, the precision (uncertainty) values estimated by the GDN model are strongly correlated with the errors, in the sense that the red points tend to lie closer to the imaginary ground-truth line than the blue points. In the following section, we consider the impact of our $SE(3)$ Bayesian filter on reducing these estimation errors and the associated uncertainty.

\subsection{Error and uncertainty reduction using a Bayesian filter}
\label{sec:se3_bayes_filter_exp}

To evaluate the impact of our $SE(3)$ Bayesian filter on the pose estimates and uncertainty values produced by the GDN model, we re-used the original test data set that we used to compare the performance of the GDN and CNN models. However, for this experiment, we treated the data set as a \emph{sequence} of sensor images and corresponding pose estimates, which were generated as the robot arm moves between consecutive random contact poses. Since the pose changes between consecutive sensor contacts can be computed from the test data labels, we can compute the state dynamics transformation, $\boldsymbol{\mathrm{T}}$, in Equation~\ref{eqn:se3_rv_probabilistic_transform}, for time step $k$ as:
\begin{equation}
	\label{eqn:se3_filter_exp_state_trans_model}
	\boldsymbol{\mathrm{T}}_{k}
	\, = \,
	\exp \left( \boldsymbol\psi_{k}^{\wedge} \right)
	\boldsymbol{\mathrm{X}}_{k} \boldsymbol{\mathrm{X}}_{k-1}^{-1}
\end{equation}

Here, $\boldsymbol{\mathrm{X}}_{k}$ and $\boldsymbol{\mathrm{X}}_{k-1}$ are the contact poses (i.e., the test data labels) at time steps $k$ and $k-1$, and $\boldsymbol\psi_{k} \sim \mathcal{N} \left(\boldsymbol{\mathrm{0}}, \boldsymbol\Sigma_{\boldsymbol\psi} \right)$ is a Gaussian noise perturbation applied at time step $k$ (applied as a left-perturbation in the global tangent space), which represents simulated noise in the state dynamics. We specify the noise perturbation covariance, $\boldsymbol\Sigma_{\boldsymbol\psi} \in \mathbb{R}^{6 \times 6}$, as a diagonal matrix:
\begin{equation}
	\label{eqn:se3_filter_exp_state_trans_cov}
	\boldsymbol\Sigma_{\boldsymbol\psi}
	=
	\sigma_{\psi}^{2}
	\begin{bmatrix}
		1 & & & & \cdots & 0 \\
		& 1 & & & & \vdots \\
		& & 1 & & & \\
		& & & \left(\frac{\pi}{180}\right)^{2} & & \\
		\vdots & & & & \left(\frac{\pi}{180}\right)^{2} & \\
		0 & \cdots & & & & \left(\frac{\pi}{180}\right)^{2}
	\end{bmatrix}
\end{equation}

\begin{table*}[h]
	\small\sf\centering
	\caption{Pose component MAEs for 10 GDN models followed by Bayesian filter with different state dynamics noise levels, $\sigma_{\psi}$ (mean values $\pm$ standard deviation across 10 models). The lowest mean MAE values are highlighted in bold.
	\label{tab:bayes_filter_results}}
	\begin{tabular}{cccccc}
	\toprule
	Metric & $\sigma_{\psi} = \infty$ & $\sigma_{\psi} = 10.0$ & $\sigma_{\psi} = 1.0$ & $\sigma_{\psi} = 0.1$ & $\sigma_{\psi} = 0.01$\\
	\midrule
	MAE $v_{x}$ (mm/s) & 0.4259 $\pm$ 0.0054 & 0.4224 $\pm$ 0.0055 & 0.3599 $\pm$ 0.0068 & 0.1603 $\pm$ 0.0042 & \textbf{0.0624} $\pm$ 0.0040\\	
	MAE $v_{y}$ (mm/s) & 0.4224 $\pm$ 0.0051 & 0.4215 $\pm$ 0.0055 & 0.3522 $\pm$ 0.0062 & 0.1608 $\pm$ 0.0047 & \textbf{0.0648} $\pm$ 0.0048\\	
	MAE $v_{z}$ (mm/s) & 0.1230 $\pm$ 0.0015 & 0.1231 $\pm$ 0.0015 & 0.1214 $\pm$ 0.0015 & 0.0978 $\pm$ 0.0014 & \textbf{0.0685} $\pm$ 0.0026\\	
	MAE $\omega_{x}$ (rad/s) & 0.0087 $\pm$ 0.0002 & 0.0086 $\pm$ 0.0002 & 0.0069 $\pm$ 0.0002 & 0.0032 $\pm$ 0.0001 & \textbf{0.0014} $\pm$ 0.0002\\	
	MAE $\omega_{y}$ (rad/s) & 0.0111 $\pm$ 0.0002 & 0.0109 $\pm$ 0.0002 & 0.0073 $\pm$ 0.0003 & 0.0032 $\pm$ 0.0002 & \textbf{0.0013} $\pm$ 0.0002\\	
	MAE $\omega_{z}$ (rad/s) & 0.0203 $\pm$ 0.0003 & 0.0200 $\pm$ 0.0003 & 0.0143 $\pm$ 0.0003 & 0.0053 $\pm$ 0.0002 & \textbf{0.0019} $\pm$ 0.0003\\
	\bottomrule
	\end{tabular}
\end{table*}

\begin{figure*}
	\centering
	\includegraphics[width=\textwidth]{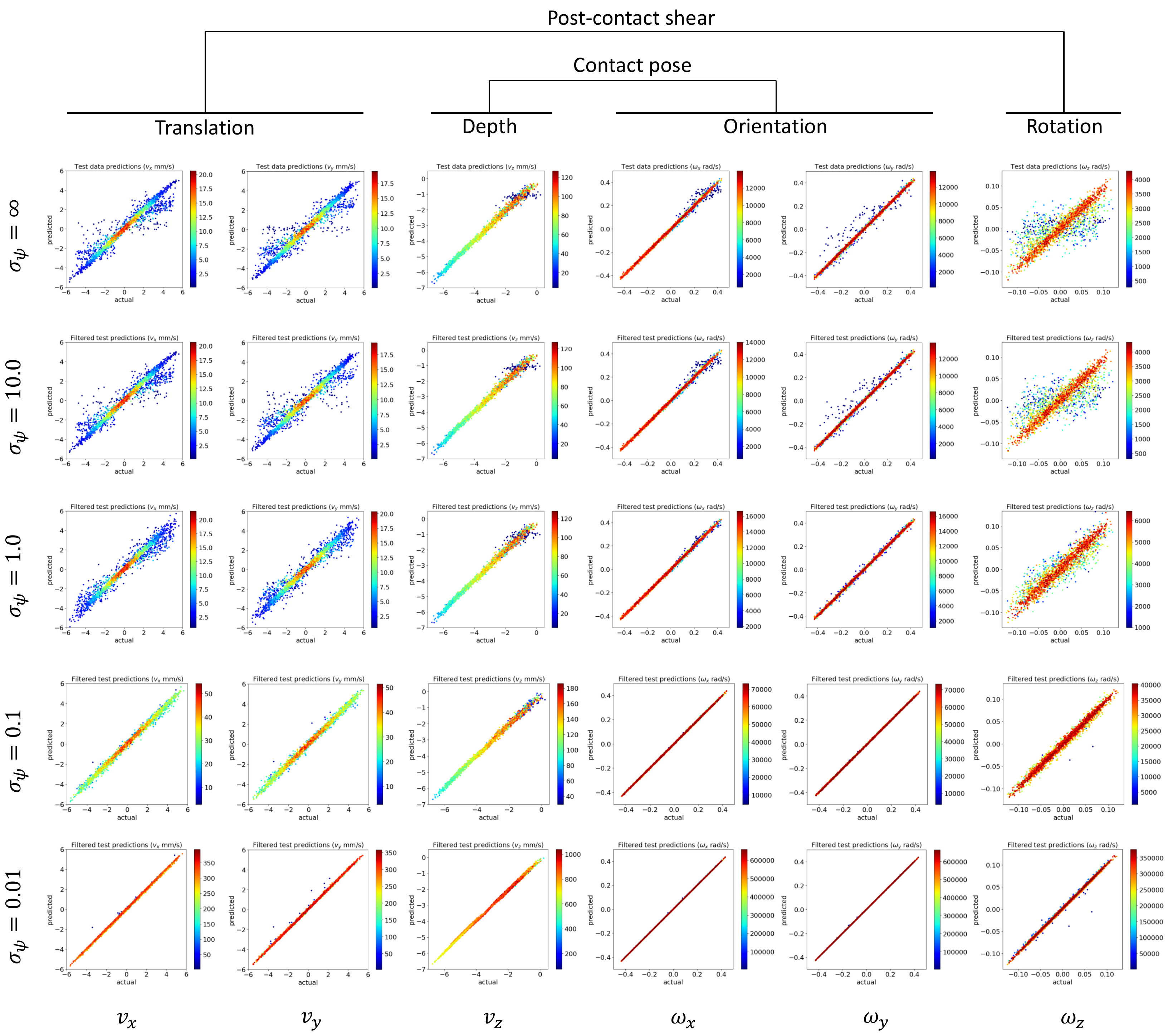}
	\caption{Distribution of test errors for best-performing (lowest loss) GDN model followed by Bayesian filter with different state dynamics noise levels, $\sigma_{\psi}$. Pose estimates (predicted vs. actual) are coloured according to their precision (i.e., reciprocal variance) estimated by the model (see attached colour bars).
	\label{fig:bayes_filter_performance}}
\end{figure*}

For $\sigma_{\psi} = 1$, this definition is equivalent to using a standard deviation (i.e., uncertainty) of 1 mm/s in the three translational components of $\boldsymbol\psi_{k}$ and 1 deg/s ($\frac{\pi}{180}$ rad/s) in the three rotational components.

We applied our discriminative Bayesian filter algorithm (Algorithm~\ref{alg:se3_bayes_filter}) to the sequence of GDN pose estimates generated for the sequence of test sensor images, using Equation~\ref{eqn:se3_filter_exp_state_trans_model} to compute the $SE(3)$ transformation used in the state dynamics model at each time step. We set the corresponding perturbation noise covariance, $\boldsymbol{\Sigma}_{\boldsymbol{\phi}}$, equal to the perturbation noise covariance defined in Equation~\ref{eqn:se3_filter_exp_state_trans_cov}, i.e., $\boldsymbol{\Sigma}_{\boldsymbol{\phi}} = \boldsymbol\Sigma_{\boldsymbol\psi}$.

As in the first experiment, we were keen to avoid statistical anomalies, so we applied our Bayesian filter to each of the 10 original GDN models we had trained from different random weight initializations and evaluated the mean and standard deviation component MAEs across all models for the test data set. We repeated the experiment for four different noise levels, $\sigma_{\psi}$, which were specified on a logarithmic scale between $\sigma_{\psi}=0.01$ and $\sigma_{\psi}=10.0$. For comparison, we also included the unfiltered results, which in the context of our model is equivalent to using an infinitely large state dynamics noise level, i.e., $\sigma_{\psi}=\infty$.

The results presented in Table~\ref{tab:bayes_filter_results} show that the filtered pose/shear estimates become more accurate as the noise levels in the real state dynamics and the state dynamics model become smaller. In the extreme case, when $\sigma_{\psi} = \infty$, the state dynamics model does not add any information when fusing pose information over consecutive time steps, so the best we can do in this situation is to use the unfiltered GDN pose estimate for the current time step (see Table~\ref{tab:nn_pose_shear_estimate_results}).

As in the previous experiment, we visualise the distribution of test sequence errors by plotting the filtered pose estimates against the actual poses for the best-performing GDN model and the different noise levels, $\sigma_{\psi}$
(Figure~\ref{fig:bayes_filter_performance}). With reference to these plots, we make the following observations. Firstly, the accuracy of the filtered estimates increases as the state dynamics noise level, $\sigma_{\psi}$, is decreased (the amount of scatter about the imaginary ground-truth line decreases). Secondly, the filtered uncertainty estimates become smaller as the state dynamics noise level, $\sigma_{\psi}$, is decreased (and the relative proportion of points coloured red increases and the proportion coloured blue decreases). Both of these observations are a consequence of the state dynamics model becoming more accurate as the noise levels become smaller. This allows more effective combination of consecutive pose estimates, which increases accuracy and reduces the associated uncertainty.

\subsection{Task 1: Object pose tracking}
\label{sec:obj_pose_tracking_exp}

In this experiment, we show how our tactile robotic system can be configured to track the pose of a moving object. We demonstrate this capability using two robot arms: the first arm (the \emph{leader robot}) moves an object around in 3D space, while a second arm fitted with a tactile sensor (the \emph{follower robot}) tracks the motion of the object using the tactile servoing controller described in Section~\ref{sec:servoing_controller}.

There are two parts to this experiment. In the first part, we show that the follower arm can track changes to individual pose components of a moving object. More specifically, we track translational motion along the $x$, $y$ and $z$ axes of the robot work frame, and the corresponding $\alpha$, $\beta$ and $\gamma$ rotational motion around these axes. In the second part of the experiment, we show that the follower arm can track simultaneous changes to all pose components while the leader arm moves the object in a complex periodic motion.

\begin{figure}
	\centering
	\includegraphics[width=\columnwidth]{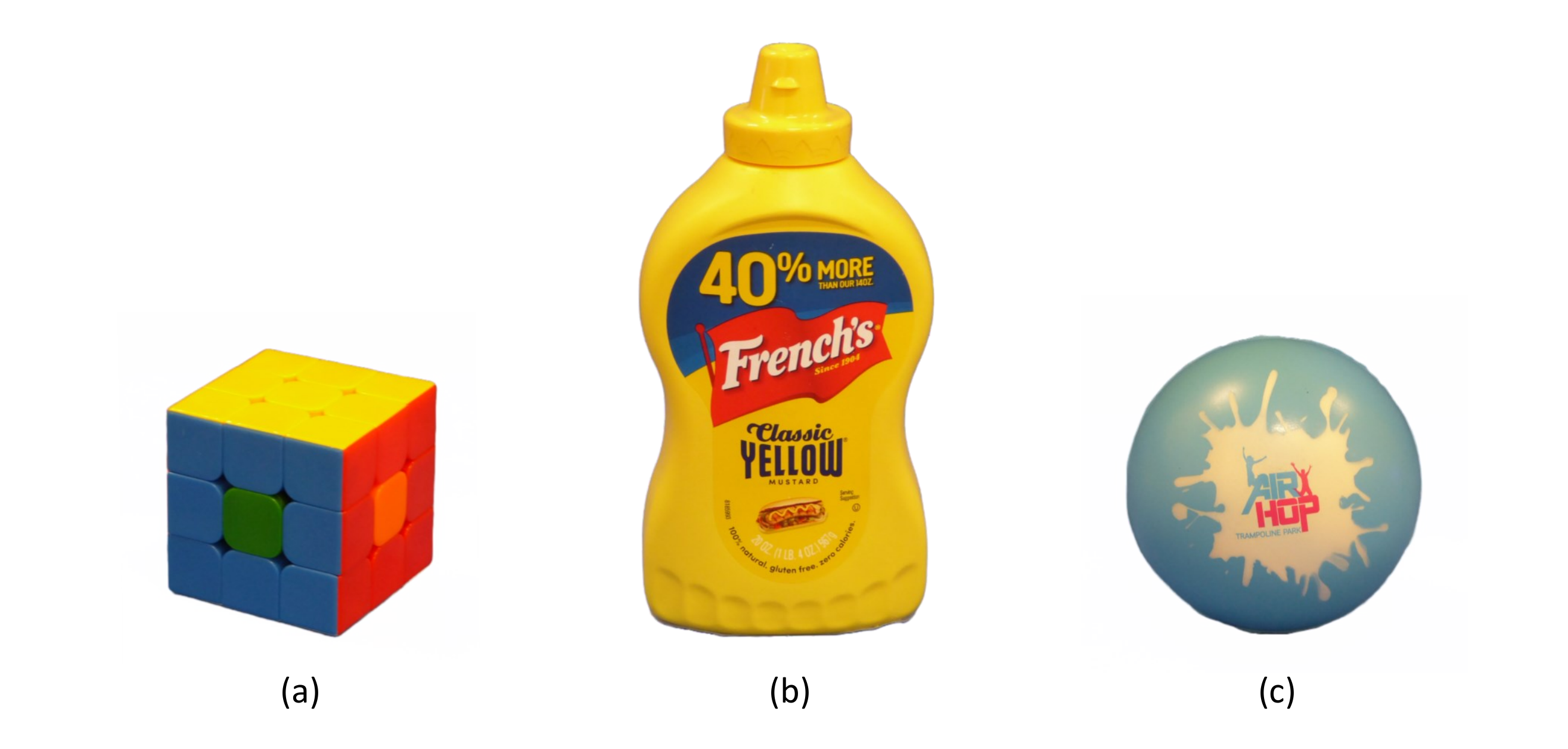}
	\caption{Everyday objects used in object tracking experiments. (a) Rubik's cube. (b) Mustard bottle. (c) Soft compliant foam ball.
	\label{fig:tracking_objects}}
\end{figure}

\begin{figure}
	\centering
	\includegraphics[width=\columnwidth]{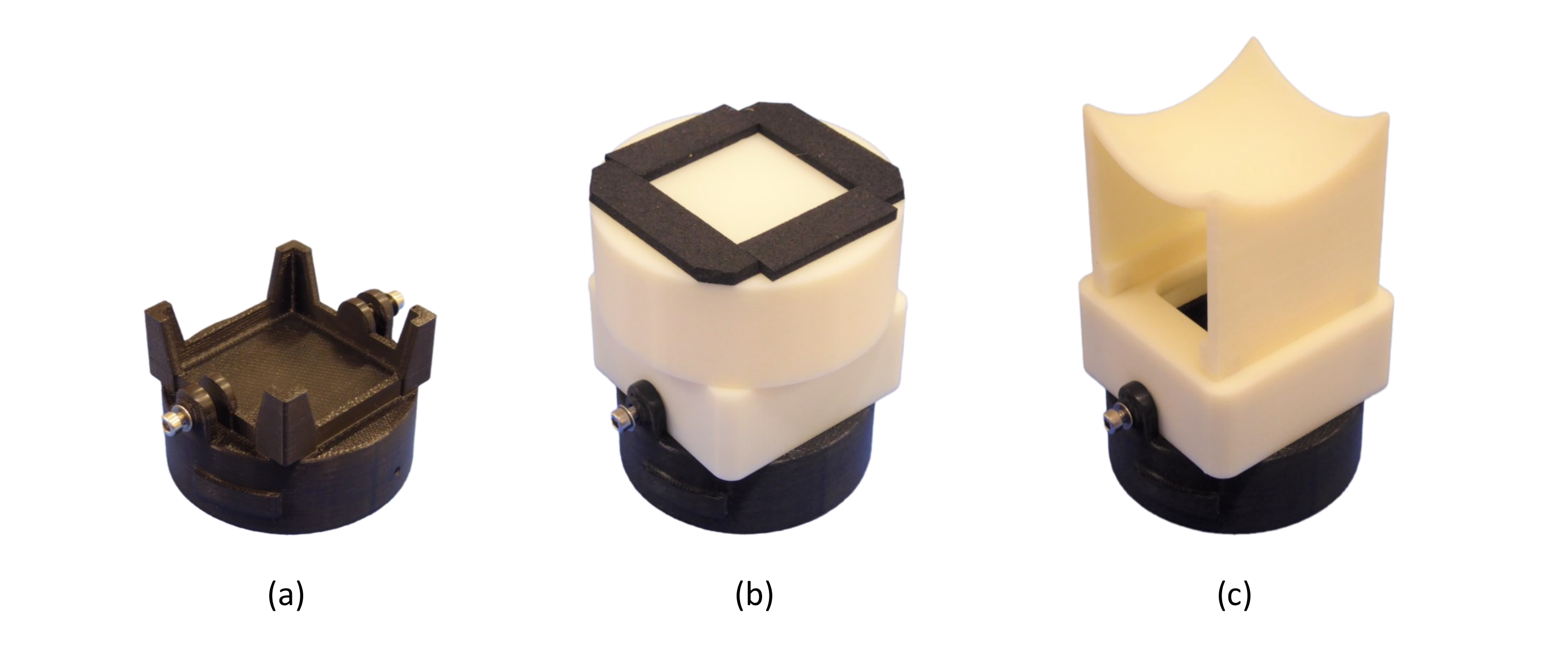}
	\caption{End-effector adaptor and surface mounts: (a) End-effector adaptor for attaching different surfaces, (b) Flat surface mount with non-slip foam pads (used for Rubik's cube and mustard bottle), (c) Concave curved surface mount (used for foam ball).
	\label{fig:end_effector_adaptor}}
\end{figure}

Another key difference between the two parts of the experiment is that in the first part we only track a flat surface attached to the end of the leader arm, whereas in the second part we also track several everyday objects that are held in position against the leader arm by the follower arm (Figure~\ref{fig:tracking_objects}). Hence, the second part of the experiment also demonstrates a form of dual-arm object manipulation. To prevent objects from sliding around or rolling off the end of the leader arm while they are being held against it by the follower arm, we developed an end-effector adapter for attaching a flat or concave curved surface to the end of the arm (Figure~\ref{fig:end_effector_adaptor}). The attached flat surface has foam rubber pads around the edges to reduce slippage while the second arm is holding an object against it.

For both parts of this experiment, we used the controller parameters listed in Table~\ref{tab:object_tracking_control_params}. The feedback reference pose specifies that the tactile sensor should be orientated normal to the contacted surface at a contact depth of 6 mm. Since the feedforward reference velocity is not required for object tracking tasks, it is set to zero.

\begin{table}[h]
	\small\sf\centering
	\caption{Controller parameters (gain matrices, clipping ranges, feedback reference pose and feedforward reference velocity) used in tactile servoing controller for follower robot arm in object tracking experiments. The feedback reference pose is expressed as a 6-vector, where the first three elements denote the 3D position and the last three elements denote the 3D rotation in extrinsic-$xyz$ Euler format.
	\label{tab:object_tracking_control_params}}
	\begin{tabular}{cc}
	\toprule
	Parameter & Value\\
	\midrule
	$\boldsymbol{\mathrm{K}}_{p}$ & $\mathrm{diag}(5,5,5,2,2,0)$\\	
	$\boldsymbol{\mathrm{K}}_{i}$ & $\mathrm{diag}(0.5,0.5,0.5,0.2,0.2,0.2)$\\	
	$\boldsymbol{\mathrm{K}}_{d}$ & $\mathrm{diag}(0.5,0.5,0.5,0.2,0.2,0.2)$\\
	Integral error clipping & Not used\\
	Feedback ref pose & $(0,0,6,0,0,0)$\\
	Feedforward ref velocity & $(0,0,0,0,0,0)$\\
	\bottomrule
	\end{tabular}
\end{table}

\subsubsection{Tracking changes to single pose components.}
\label{sec:object_pose_tracking_single_exp}

\begin{figure*}
	\centering
	\includegraphics[width=\textwidth]{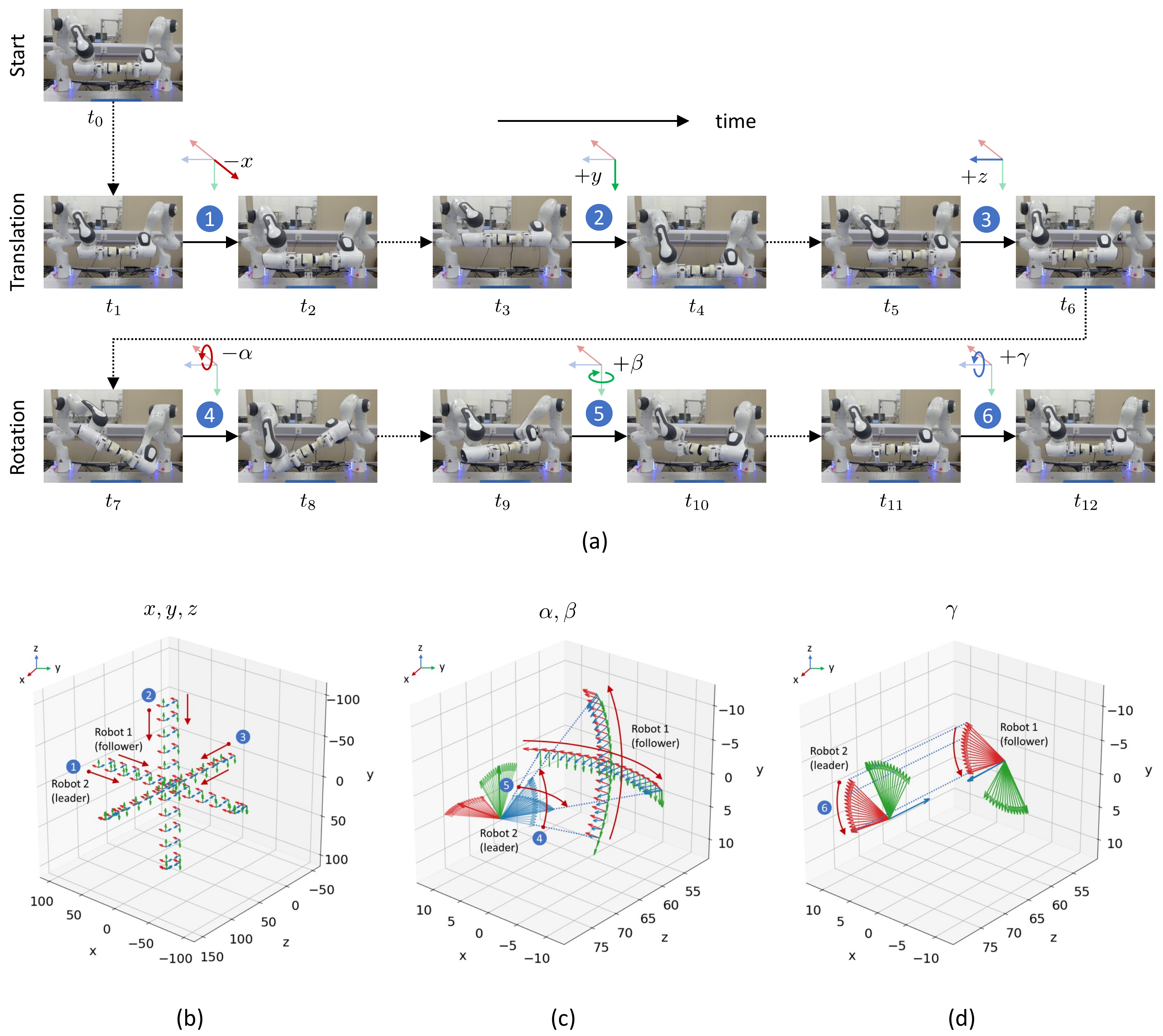}
	\caption{Using the follower arm to track changes to individual components of the leader arm pose. (a) Tracking sequence: translation along $-x \rightarrow y \rightarrow z$ axes (1-3), followed by $-\alpha \rightarrow \beta \rightarrow \gamma$ rotation around these axes (4-6). (b) Leader and follower arm pose trajectory as leader arm translates along $x$, $y$ and $z$ axes. (c) Leader and follower arm pose trajectory as leader arm rotates about $x$ and $y$ axes ($\alpha$ and $\beta$). (d) Leader and follower arm pose trajectory as leader arm rotates about $z$ axis ($\gamma$). The numbered points relate corresponding sections of the tracking sequence and pose trajectories.
	\label{fig:object_tracking_single}}
\end{figure*}

For the first part of the experiment, we initially positioned the follower arm tactile sensor in direct contact with the leader arm flat surface at a contact depth of approximately 6 mm so that its central axis was normal to the flat surface. We then used the leader robot to move the flat surface through a sequence of 200 mm translations along the $x$, $y$, and $z$ axes (of the robot work frame), and then through 60 degree $\alpha$, $\beta$ and $\gamma$ rotations about these axes (Figure~\ref{fig:object_tracking_single}(a)). During the tracking sequence, we recorded the end-effector poses and corresponding time stamps for both robot arms at the start of each control cycle. This allowed us to (approximately) match up the corresponding poses for the two arms and plot them in 3D for different points in the trajectory after the experiment had finished (Figure~\ref{fig:object_tracking_single}(b)-(d)).

For plots relating to translational pose components (Figure~\ref{fig:object_tracking_single}(b))), we removed any variation in the rotational components from the idealised (i.e., noise-free) response of the follower arm. Similarly, for plots relating to rotational pose components (Figures~\ref{fig:object_tracking_single}(c)-(d)), we removed any variation in the translational components from the idealised response of the follower arm. This allowed us to focus on individual pose components when evaluating how the follower arm responds to changes in leader arm pose. If we had not done this, but instead plotted the raw unaltered poses, it would have made it extremely difficult to compare individual pose components of the leader and follower arms at any point in time, particularly for the rotational components ($\alpha$, $\beta$ and $\gamma$). For example, in Figure~\ref{fig:object_tracking_single}(d), as well as rotating around the $z$-axis in response to a leader arm rotation about this axis, the follower arm also translates in the $xy$-plane to some extent. This is because it is extremely difficult to accurately align the central axis of the follower arm tactile sensor with the central axis of the leader arm flat surface and so there is some degree of translation of the tactile sensor as well as rotation. If we had included these translational components in the plotted follower arm poses it would have been extremely difficult to see the correspondence between the rotated poses for the leader and follower arms because the origins of the follower pose coordinate frames would not be coincident in the $xy$-plane. However, we should point out that we only needed to use this method of post-processing to visualise the results for the first part of the experiment. In the second part of the experiment, where we varied all components of the pose simultaneously, we found that we were able to plot the raw unaltered poses for both arms at different points in the trajectory and still see the correspondence between the two sets of poses.

The pose trajectory plots (Figures~\ref{fig:object_tracking_single}(b)-(d)) show that the follower arm can track changes to individual pose components of the leader arm while the leader arm follows a simple trajectory, translating along or around a single coordinate axis.

\subsubsection{Tracking simultaneous changes to all pose components.}
\label{sec:object_pose_tracking_multi_exp}

\begin{figure*}
	\centering
	\includegraphics[width=\textwidth]{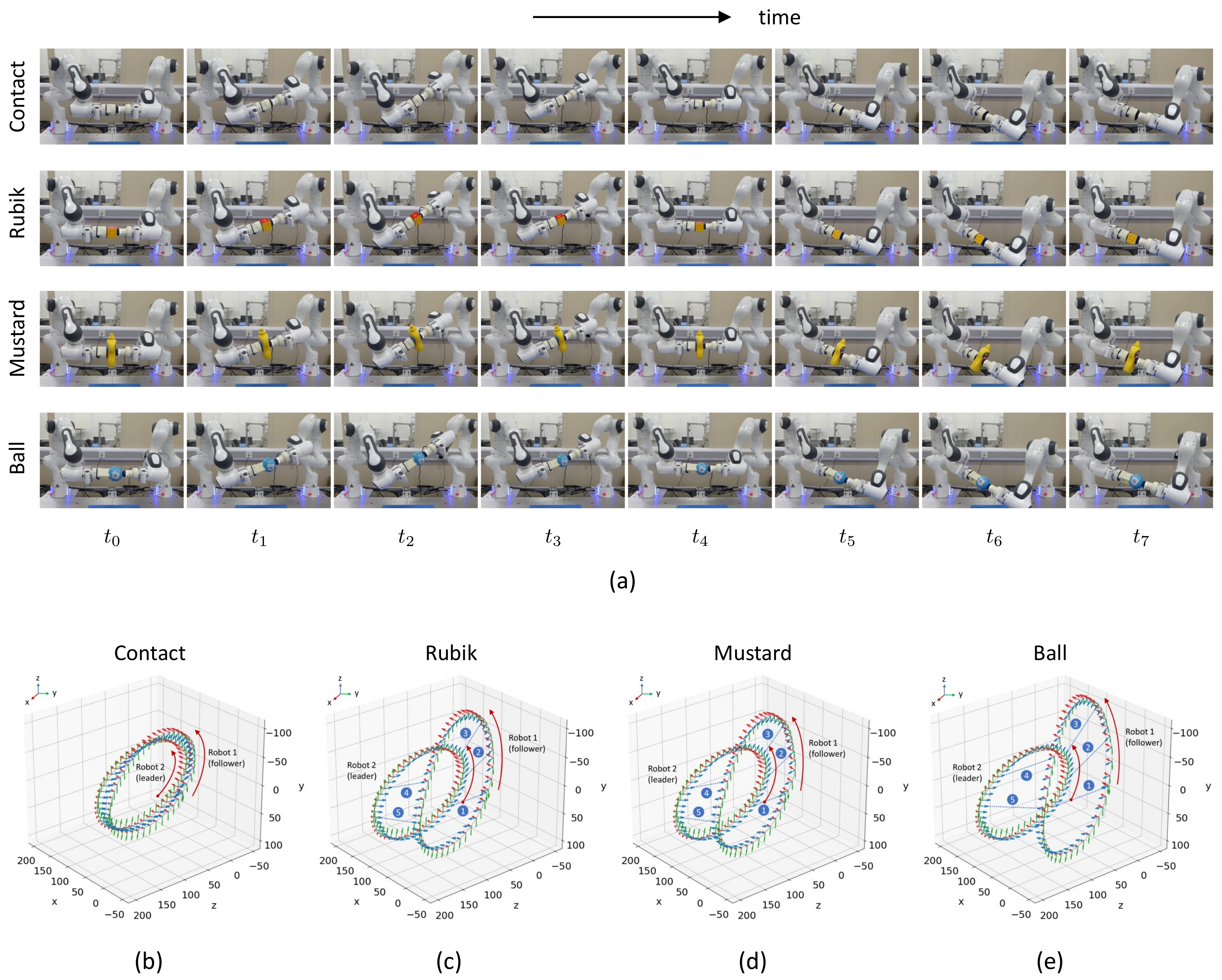}
	\caption{Using the follower arm to track simultaneous changes to all components of the leader arm pose. (a) Tracking sequences for direct (flat) surface contact, Rubik's cube, mustard bottle and soft foam ball. Leader and follower robot pose trajectory for a single period of the leader arm trajectory for, (b) direct surface contact, (c) Rubik's cube, (d) mustard bottle, and (e) soft foam ball. In (c)-(e), the numbered points indicate corresponding poses of the two robot arms at several points in the pose trajectory .
	\label{fig:object_tracking_multi}}
\end{figure*}

In the second part of the experiment, we moved the leader robot arm in a more complex velocity trajectory, $\boldsymbol{\mathrm{v}}(t)$, where all of the pose components were varied at the same time using the periodic function:
\begin{equation}
	\boldsymbol{\mathrm{v}}(t) = \frac{2 \pi \boldsymbol{\mathrm{b}}}{T} \odot \cos \left(\frac{2 \pi t}{T} + \boldsymbol\phi \right)
\end{equation}

Here, the amplitude $\boldsymbol{\mathrm{b}}=\left[ 75, 75, 75, \frac{25 \pi}{180}, \frac{25 \pi}{180}, \frac{25 \pi}{180} \right]^{\top}$, phase $\boldsymbol\phi=\left[ \frac{\pi}{2}, 0, 0, 0, 0, 0 \right]^{\top}$, and period $T=30$ s. The translational components of the amplitude $\boldsymbol{\mathrm{b}}$ have units of mm and the rotational components have units of radians. We tracked the velocity trajectory over three full periods ($3 \times 30 = 90$ s).

As well as tracking a flat surface attached to the end of the leader arm as we did in the first part of the experiment, in this second part we also tracked several objects (see Figure~\ref{fig:tracking_objects}) that were held between the two arms as they followed the leader arm trajectory. As in the first part of the experiment, we recorded the end-effector poses and corresponding time stamps for both robot arms at the start of each control cycle so that we could match up the corresponding poses from both robots and plot them in 3D after the experiment had finished (Figure~\ref{fig:object_tracking_multi}(b)-(e)).

The time-lapse photos and pose trajectory plots (Figure~\ref{fig:object_tracking_multi})) show that the follower arm can track simultaneous changes to all components of the leader arm pose as the leader arm follows a complex periodic trajectory. Moreover, the follower arm can also hold an object against the leader arm while it is following its trajectory, thereby implementing a simple form of 3D object manipulation.

\subsection{Task 2: Surface following}
\label{sec:surface_follow_exp}

\begin{figure}
	\centering
	\includegraphics[width=\columnwidth]{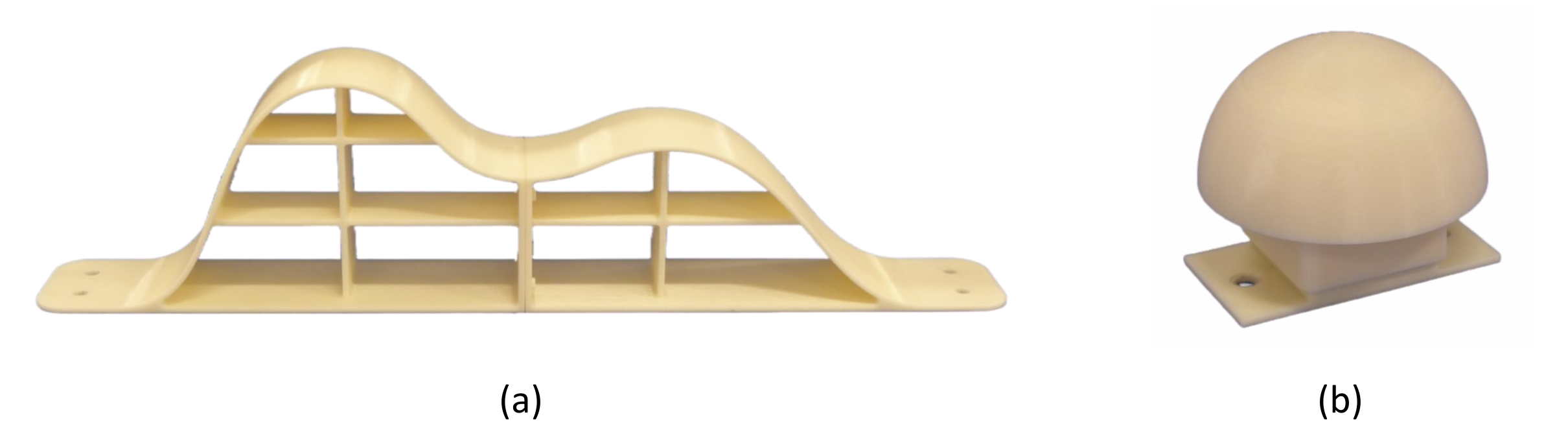}
	\caption{Surfaces used in surface following experiments: (a) curved ramp, (b) hemispherical dome.
	\label{fig:surface_follow_objects}}
\end{figure}

\begin{figure*}
	\centering
	\includegraphics[width=\textwidth]{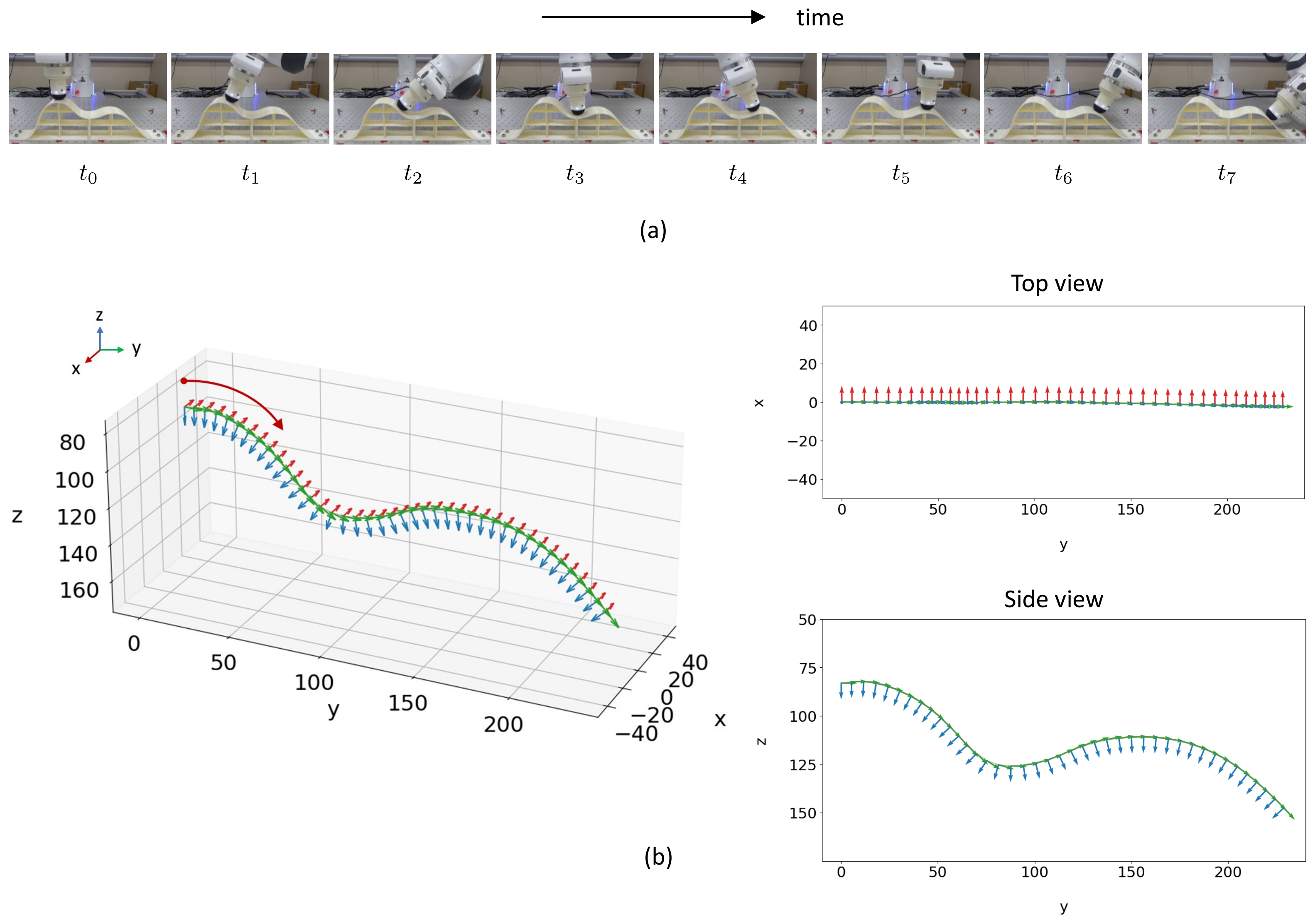}
	\caption{Using the robot arm to follow the surface of a curved ramp. (a) Time-lapse photos. (b) Robot arm (end-effector) pose trajectory.
	\label{fig:surface_follow_ramp}}
\end{figure*}

\begin{figure*}
	\centering
	\includegraphics[width=\textwidth]{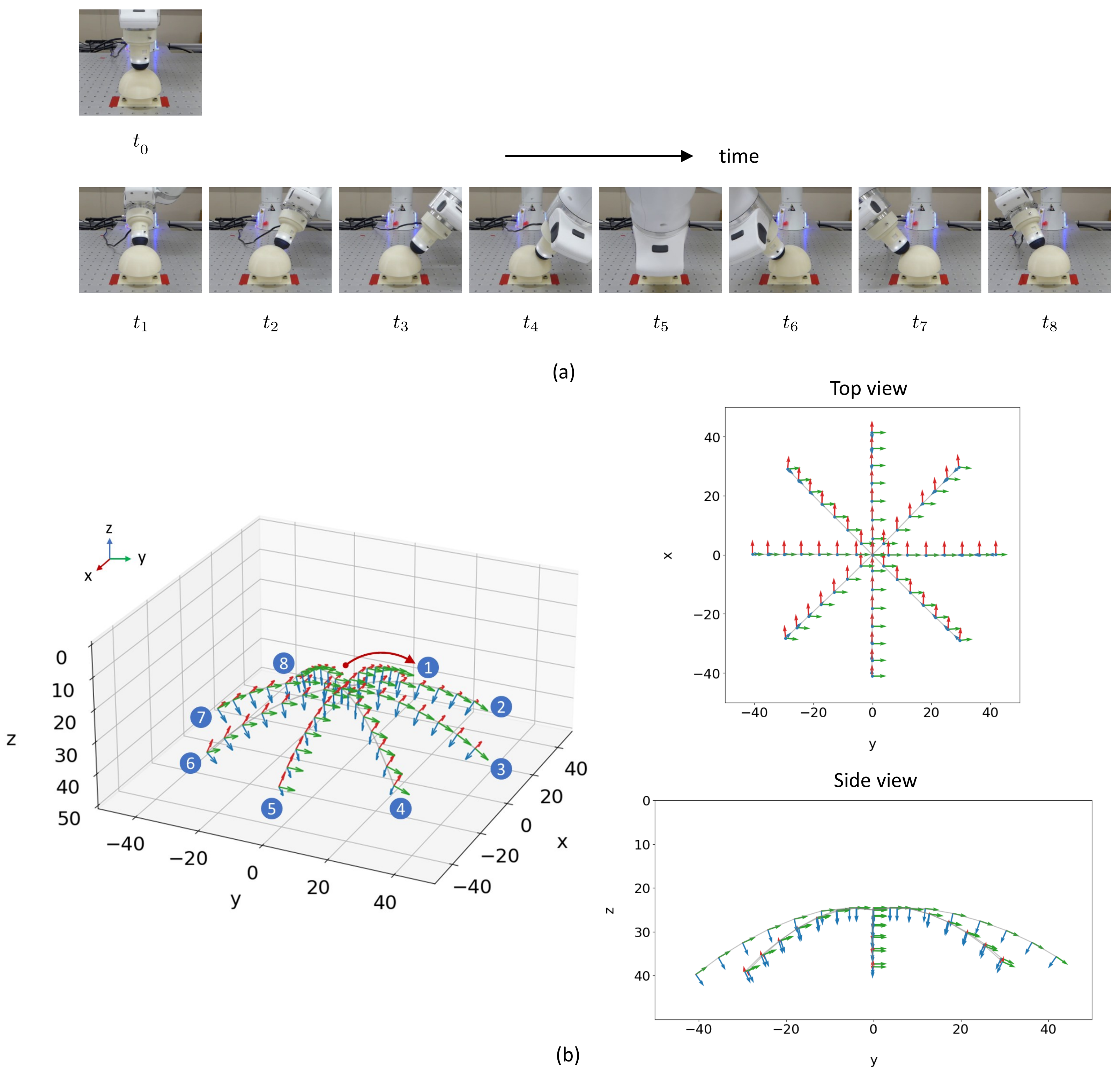}
	\caption{Using the robot arm to follow radial paths from the centre of the surface of a hemispherical dome. (a) Time-lapse photos. (b) Robot arm (end-effector) pose trajectory. In (b), the numbered points correspond to different radial paths over the surface.
	\label{fig:surface_follow_hemisphere}}
\end{figure*}

In this experiment, we show how our tactile robotic system can be configured for surface following tasks. In particular, we demonstrate how a single robot arm fitted with a tactile sensor can smoothly follow the surface of a curved object in a similar way to a human when it strokes its fingers over an object. This type of capability is useful for collecting tactile data to determine the extent of an object, map its shape, or determine its texture. We carry out this experiment for two scenarios: traversing a straight line projection on the surface of a curved ramp, and traversing a sequence of eight straight line projections outwards from the centre of a hemispherical dome at 45 degree intervals. The surfaces used in these two scenarios are shown in Figure~\ref{fig:surface_follow_objects}.

For both parts of this experiment, we use the tactile servoing controller described in Section~\ref{sec:servoing_controller} with the controller parameters listed in Table~\ref{tab:surface_follow_control_params}. The feedback reference pose specifies that the sensor should be orientated normal to the contacted surface at a contact depth of 3 mm. Since, the feedforward reference velocity depends on the particular surface following task being performed, it is specified in the subsections that describe each task.

\begin{table}[h]
	\small\sf\centering
	\caption{Controller parameters used in tactile servoing controller for surface following experiments. The feedback reference pose is expressed as a 6-vector, where the first three elements denote the position and the last three elements denote the rotation in extrinsic-$xyz$ Euler format.
	\label{tab:surface_follow_control_params}}
	\begin{tabular}{cc}
	\toprule
	Parameter & Value\\
	\midrule
	$\boldsymbol{\mathrm{K}}_{p}$ & $\mathrm{diag}(0,0,2,2,2,0)$\\	
	$\boldsymbol{\mathrm{K}}_{i}$ & $\mathrm{diag}(0,0,0.1,0.1,0.1,0)$\\	
	$\boldsymbol{\mathrm{K}}_{d}$ & $\mathrm{diag}(0,0,0.05,0.05,0.05,0)$\\
	Integral error clipping & $\left[ -25, 25 \right]$ all components\\
	Feedback ref pose & $(0,0,3,0,0,0)$\\
	Feedforward ref velocity & Task-dependent\\
	\bottomrule
	\end{tabular}
\end{table}

\subsubsection{Surface following on a curved ramp.}
\label{sec:surface_follow_ramp}

For this surface following task, we initially positioned the robot arm so that the tactile sensor made contact with the highest part of the curved ramp at a contact depth of approximately 3 mm, and with the $y$-axis of the sensor aligned with the $y$-axis of the robot work frame (i.e., so that the $y$-axis points along the length of the ramp). We set the (feedforward) reference velocity to 10 mm/s, directed along the $y$-axis of the (feedback) reference contact pose (i.e., so that $\boldsymbol{\mathrm{u}}_{s^{\prime}2}=(0,10,0,0,0,0)$ in Figure~\ref{fig:servoing_controller}). Specifying the reference contact pose and reference velocity in this way causes the sensor to move at 10 mm/s tangentially to the surface, while remaining at a contact depth of 3 mm. During the surface following sequence, we recorded the end-effector poses and corresponding time stamps at the start of each control cycle (Figure~\ref{fig:surface_follow_ramp}). 

For this surface following task, the time-lapse photos and pose trajectory plots show that the robot arm can successfully follow this type of gently curving surface while the sensor remains in contact with and orientated normal to the surface.

\subsubsection{Surface following on a hemispherical dome.}
\label{sec:surface_follow_hemisphere}

For this surface following task, we initially positioned the robot arm so that the tactile sensor made contact with the centre of the dome at a contact depth of approximately 3 mm, with the $y$-axis of the sensor aligned with the $y$-axis of the robot work frame. 

When following the $i$th radial path from the centre of the hemisphere, at angle $\theta_{i} = 0, \frac{\pi}{4}, \frac{\pi}{2}, \ldots, \frac{7\pi}{4}$ radians, we set the (feedforward) reference velocity, $\boldsymbol{\mathrm{u}}_{s^{\prime}2}$ in Figure~\ref{fig:servoing_controller}, equal to:
\begin{equation}
	\boldsymbol{\mathrm{u}}_{s^{\prime}2}
	\, = \,
	\left( u \cos \theta_{i}, u \sin \theta_{i}, 0, 0, 0, 0) \right)
\end{equation}
where the magnitude of the velocity, $u = 10$ mm/s. Specifying the reference pose and reference velocity in this way causes the sensor to move at 10 mm/s tangentially to the surface in direction $\theta_{i}$, while remaining at a contact depth of 3 mm. During the surface following sequence, we recorded the end-effector poses and corresponding time stamps at the start of each control cycle as the sensor moved along each of the radial paths (Figure~\ref{fig:surface_follow_hemisphere}).

For this second surface following task, the time-lapse photos and pose trajectory plots show that the robot arm can successfully follow this type of gently curving surface while the sensor remains in contact with and orientated normal to the surface.

\subsection{Task 3: Single-arm object pushing}
\label{sec:single_arm_object_pushing_exp}

In our first object pushing experiment, we demonstrate how our system can be used for single-arm object pushing tasks, similar to the ones we demonstrated in our earlier work (\cite{lloyd2021goal}). A notable improvement on our earlier work is that our new system is capable of pushing an object in a smooth continuous manner, rather than in a discrete, point-to-point motion as it did before. In terms of capability demonstration, we also show that our new system can push objects over harder medium-density fibreboard (MDF)) surfaces as well as the softer medium-density foam surfaces we used in previous work. Since the single-arm configuration forms the basis of our more complex dual-arm pushing configuration, we will discuss it first.

\begin{figure}
	\centering
	\includegraphics[width=\columnwidth]{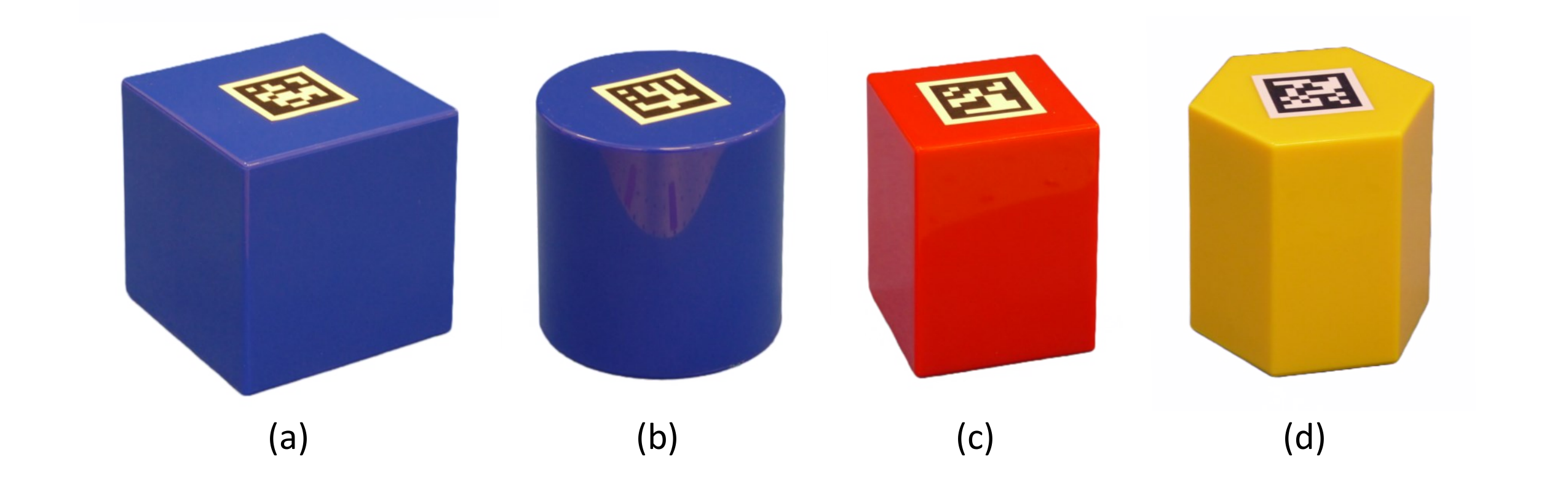}
	\caption{Regular geometric objects used in single-arm object pushing experiments: (a) large blue square prism (479 g), (b) blue circular prism (363 g), (c) small red square prism (264 g), (d) yellow hexagonal prism (310 g).
	\label{fig:single_arm_pushing_objects}}
\end{figure}

For the single-arm pushing configuration, we mounted the tactile sensor on the end of the robot arm, using a right-angle adapter (Figure~\ref{fig:exp_platform_configurations}(e)) so that it can be moved parallel to the surface during the pushing sequence without the arm getting snagged on the surface. At the start of each trial, we positioned the tactile sensor 45 mm above the surface so that its central axis lay parallel to the $y$-axis of the robot work frame, at position $(y=-250, z=100)$ in the $yz$-plane parallel to the surface. We then placed the object (approximately) centrally in front of the tactile sensor so that the contacted surface of the object was (approximately) normal to the sensor axis.

For this experiment, we pushed several regular geometric objects (Figure~\ref{fig:single_arm_pushing_objects}) across a medium-density, closed-cell, polyethylene foam surface and an MDF surface. These geometric objects are the same as the ones we used in previous work (\cite{lloyd2021goal}), but here we left out the yellow triangular prism because we wanted to be able to compare the results for the single-arm configuration with those for the dual-arm configuration. Since the sharp front apex of the triangular prism cannot be tracked by the follower arm of a dual-arm configuration using a pose estimation model that has been trained on a flat surface, it was not possible to use this object in the dual-arm configuration.

For each experimental trial, we used the robot arm to push the object towards the target at position $(y=0, z=375)$ while remaining in contact with the object. The location of the target relative to the object's starting pose means that the robot has to push the object around a corner to reach the target. To control the robot arm, we used the pushing controller described in Section~\ref{sec:pushing_controller}, with the parameters listed in Table~\ref{tab:pushing_control_params}.

\begin{table}[h]
	\small\sf\centering
	\caption{Pushing controller parameters used for single-arm and dual-arm (leader robot) object pushing experiments. The feedback reference pose is expressed as a 6-vector, where the first three elements denote the position and the last three elements denote the rotation in extrinsic-$xyz$ Euler format.
	\label{tab:pushing_control_params}}
	\begin{tabular}{cc}
	\toprule
	Parameter & Value\\
	\midrule
	PID 1 (MIMO)\\
	\midrule
	$\boldsymbol{\mathrm{K}}_{p}$ & $\mathrm{diag}(1,0,0,1,0,0)$\\	
	$\boldsymbol{\mathrm{K}}_{i}$ & $\mathrm{diag}(0.1,0,0,0.1,0,0)$\\	
	$\boldsymbol{\mathrm{K}}_{d}$ & $\mathrm{diag}(0.1,0,0,0.1,0,0)$\\
	Integral error clipping & $\left[ -25, 25 \right]$ all components\\
	Feedback ref pose & $(0,0,0,0,0,0)$\\
	Feedforward ref velocity & $(0,0,10,0,0,0)$\\
	\midrule
	PID 2 (SISO)\\
	\midrule
	$K_{p}$ & $0.9$\\	
	$K_{i}$ & $0.3$ (single-arm) / $0.5$ (dual-arm)\\	
	$K_{d}$ & $0.9$\\
	Integral error clipping & $\left[ -10, 10 \right]$\\
	Controller output clipping & $\left[ -15, 15 \right]$\\
	Ref bearing & $0$\\
	\bottomrule
	\end{tabular}
\end{table}

During the course of each trial, we recorded the end-effector poses and corresponding time stamps at the start of each control cycle. To avoid statistical anomalies, we repeated the trial five times for each object and computed the mean $\pm$ standard deviation final target error across all five trials (Table~\ref{tab:single_arm_pushing_results}). As in \cite{lloyd2021goal}, we define the final target error as the perpendicular distance from the target to the sensor-object contact normal on completion of the push sequence. This provides a measure of how close the pusher is able to approach the target with the object.

\begin{table}[h]
	\small\sf\centering
	\caption{Single-arm pushing final target error (mean $\pm$ standard deviation perpendicular distance from target to sensor-object contact normal on completion of push sequence). All statistics are computed over 5 independent trials.
	\label{tab:single_arm_pushing_results}}
	\begin{tabular}{ccc}
	\toprule
	Object & Foam surface & MDF surface\\
	\midrule
	Blue square & 0.50 $\pm$ 0.56 mm & 4.45 $\pm$ 0.78 mm\\	
	Blue circle & 6.93 $\pm$ 0.06 mm & 9.45 $\pm$ 0.13 mm\\	
	Red square & 1.88 $\pm$ 0.71 mm & 4.38 $\pm$ 0.77 mm\\	
	Yellow hexagon & 0.33 $\pm$ 1.32 mm & 2.87 $\pm$ 0.68 mm\\
	\bottomrule
	\end{tabular}
\end{table}

\begin{figure*}
	\centering
	\includegraphics[width=\textwidth]{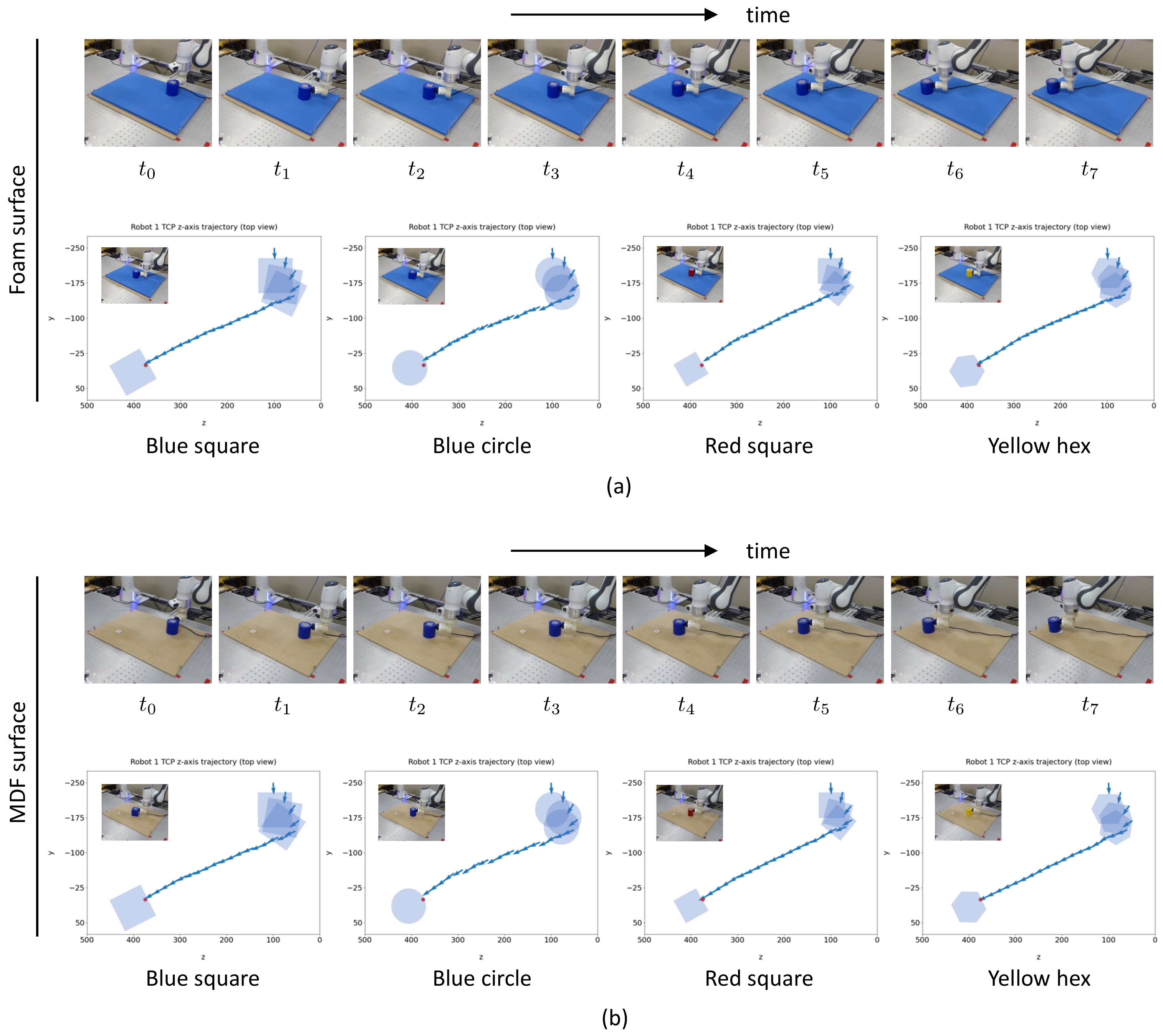}
	\caption{Using a single robot arm to push regular geometric objects across: (a) a medium density foam surface (time lapse photos show sequence for blue square prism), and (b) an MDF surface (time lapse photos show sequence for blue circular prism). In the 2D pose plots, the target is identified by a small red circle.
	\label{fig:single_arm_pushing}}
\end{figure*}

We visualised examples of the push sequences by plotting the end-effector poses in 2D and overlaid the approximate (i.e., indicative) poses of the pushed objects at the start and finish points of the trajectory (Figure~\ref{fig:single_arm_pushing}).

The results in Table~\ref{tab:single_arm_pushing_results} and Figure~\ref{fig:single_arm_pushing} show that our system can push these regular geometric objects over foam and MDF surfaces and approach the target to within less than 10 mm for the blue circular prism and to within less than 5 mm for the remaining objects.

\subsection{Task 4: Dual-arm object pushing}
\label{sec:dual_arm_object_pushing_exp}

In our second pushing experiment, we use a follower robot arm to constrain and stabilise the objects as they are pushed across a flat surface by the leader arm. In many ways, this is similar to the configuration used in the object tracking experiment (Section~\ref{sec:object_pose_tracking_multi_exp}), where we used the leader robot arm to move an object in a complex trajectory in 3D space, and the follower arm to track its motion while it holds the object against the first arm.

This experiment is split into two parts. In the first part, we use two robot arms to push the objects used in the previous single-arm experiment across foam and MDF surfaces (i.e., we essentially repeat the single-arm experiment but using two robot arms instead of one to push the objects). In the second part of the experiment, we replace the original set of geometric objects with a set of taller, double-height versions of these objects together with several taller everyday objects (e.g., bottles and containers). These taller objects cannot be pushed by a single robot arm without toppling over, so the second stabilising follower arm is essential to prevent this from happening. Even so, we were not able to push these taller objects across the foam surface without their leading edges catching on the surface, and so we could only perform this part of the experiment on the harder MDF surface.

For this dual-arm configuration, we mounted tactile sensors on both robot arms using right-angle adapters (see Figure~\ref{fig:exp_platform_configurations}(f)). At the start of each trial, we positioned the leader arm and object in the same way they were positioned at the start of each single-arm pushing trial. Then we positioned the follower arm so that its tactile sensor was approximately opposite the leader arm tactile sensor, and normal to the contacted opposite surface. During each trial, we used the leader robot arm to push the object towards the same target as before, at position $(y=0, z=375)$, while both tactile sensor end-effectors remained in contact with the object.

To control the leader robot arm, we used the same pushing controller and parameters as we used for the single-arm configuration. To control the stabilising follower arm, we used the tactile servoing controller described in Section~\ref{sec:servoing_controller}, with the parameters listed in Table~\ref{tab:stabiliser_control_params}.

\begin{table}[h]
	\small\sf\centering
	\caption{Tracking (stabilising) controller parameters used for dual-arm object pushing experiments. The feedback reference pose is expressed as a 6-vector, where the first three elements denote the position and the last three elements denote the rotation in extrinsic-$xyz$ Euler format.
	\label{tab:stabiliser_control_params}}
	\begin{tabular}{cc}
	\toprule
	Parameter & Value\\
	\midrule
	$\boldsymbol{\mathrm{K}}_{p}$ & $\mathrm{diag}(5,0,5,1,0,0)$\\	
	$\boldsymbol{\mathrm{K}}_{i}$ & $\mathrm{diag}(0.5,0,0.5,0.1,0,0)$\\	
	$\boldsymbol{\mathrm{K}}_{d}$ & $\mathrm{diag}(0.5,0,0.5,0.1,0,0)$\\
	Integral error clipping & $\left[ -200, 200 \right]$ all components\\
	Feedback ref pose & $(0,0,3,0,0,0)$\\
	Feedforward ref velocity & $(0,0,0,0,0,0)$\\
	\bottomrule
	\end{tabular}
\end{table}

During each trial for both parts of the experiment, we recorded the end-effector poses and corresponding time stamps at the start of each control cycle so that we could match up the corresponding poses at different points in the trajectory and plot them in 2D.

\subsubsection{Pushing regular geometric objects.}
\label{sec:dual_arm_short_object_pushing_exp}

\begin{figure*}
	\centering
	\includegraphics[width=\textwidth]{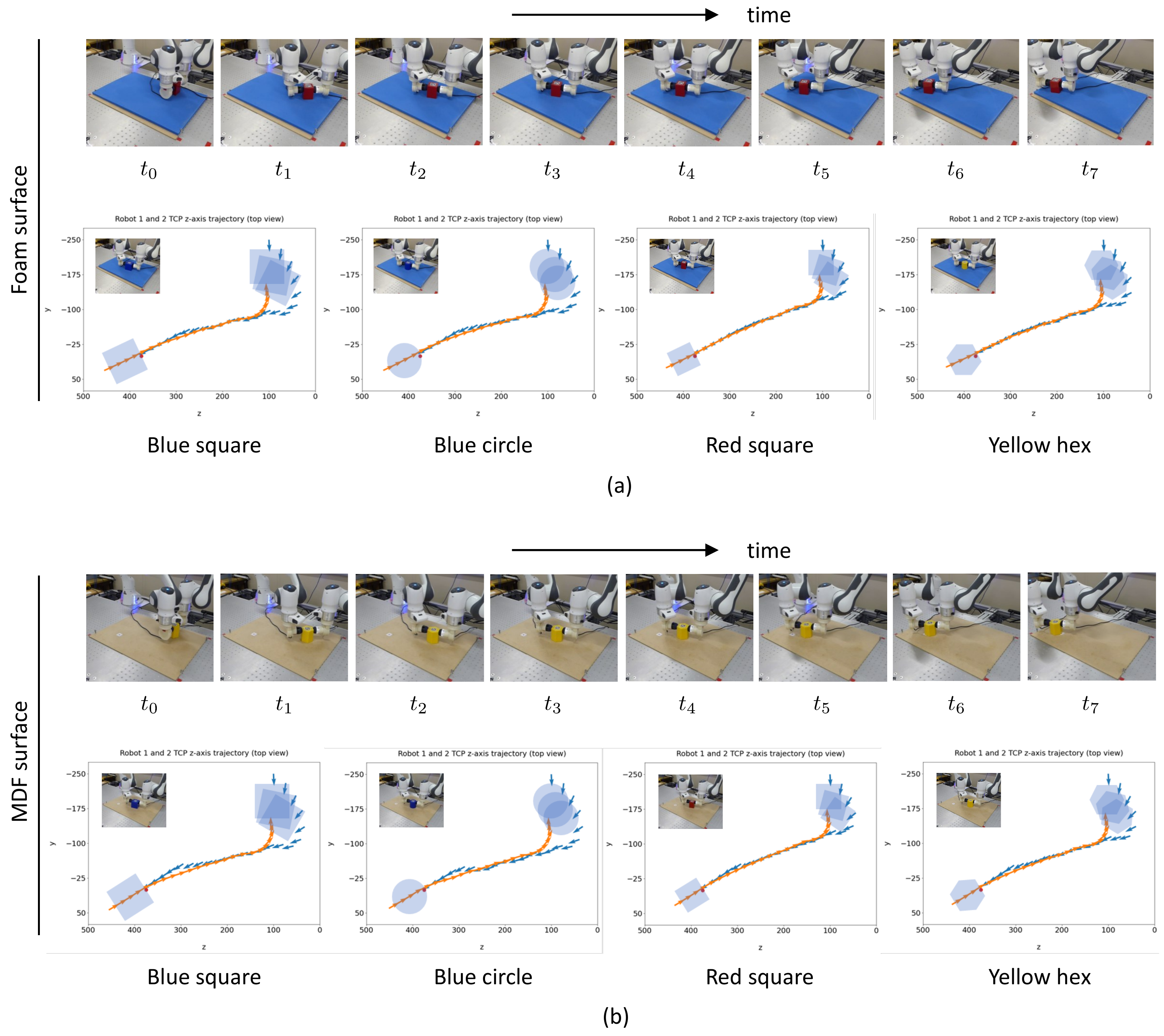}
	\caption{Using a leader and follower robot arm to push regular geometric objects across: (a) a medium density foam surface (time-lapse photos show sequence for red square prism), and (b) an MDF surface (time-lapse photos show sequence for yellow hexagonal prism). In the 2D pose plots, the target is identified by a small red circle.
	\label{fig:dual_arm_pushing}}
\end{figure*}

In the first part of the dual-arm experiment, we used two robot arms to push the same geometric objects we pushed in the single-arm experiment (Figure~\ref{fig:dual_arm_pushing}). We repeated the experiment five times for each object and then computed the mean $\pm$ standard deviation target error across all five trials (Table~\ref{tab:dual_arm_pushing_results}).

\begin{table}[h]
	\small\sf\centering
	\caption{Dual-arm pushing target error for short geometric objects (mean $\pm$ standard deviation perpendicular distance from target to sensor-object contact normal on completion of push sequence). All statistics are computed over 5 independent trials.
	\label{tab:dual_arm_pushing_results}}
	\begin{tabular}{ccc}
	\toprule
	Object & Foam surface & MDF surface\\
	\midrule
	Blue square & 4.43 $\pm$ 0.25 mm & 5.24 $\pm$ 0.24 mm\\	
	Blue circle & 4.46 $\pm$ 3.35 mm & 3.58 $\pm$ 2.04 mm\\	
	Red square & 4.97 $\pm$ 0.18 mm & 4.28 $\pm$ 0.15 mm\\	
	Yellow hexagon & 4.43 $\pm$ 0.31 mm & 4.44 $\pm$ 0.62 mm\\
	\bottomrule
	\end{tabular}
\end{table}

We visualised examples of the push sequences by plotting the end-effector poses of both robot arms in 2D and overlaid the approximate poses of the pushed objects at the start and finish points of the trajectory (Figure~\ref{fig:dual_arm_pushing}).

The results in Table~\ref{tab:dual_arm_pushing_results} and Figure~\ref{fig:dual_arm_pushing} show that our dual-arm system can push the regular geometric objects over foam and MDF surfaces, approaching the target to within less than 5 mm for all objects. In contrast to the results for the single-arm configuration, the accuracy achieved for the blue circular prism was comparable to that achieved for the other objects. In fact, for the MDF surface, the accuracy achieved for the circular prism was slightly better than for the other objects.

\subsubsection{Pushing tall objects that are prone to toppling.}
\label{sec:dual_arm_tall_object_pushing}

\begin{figure}
	\centering
	\includegraphics[width=\columnwidth]{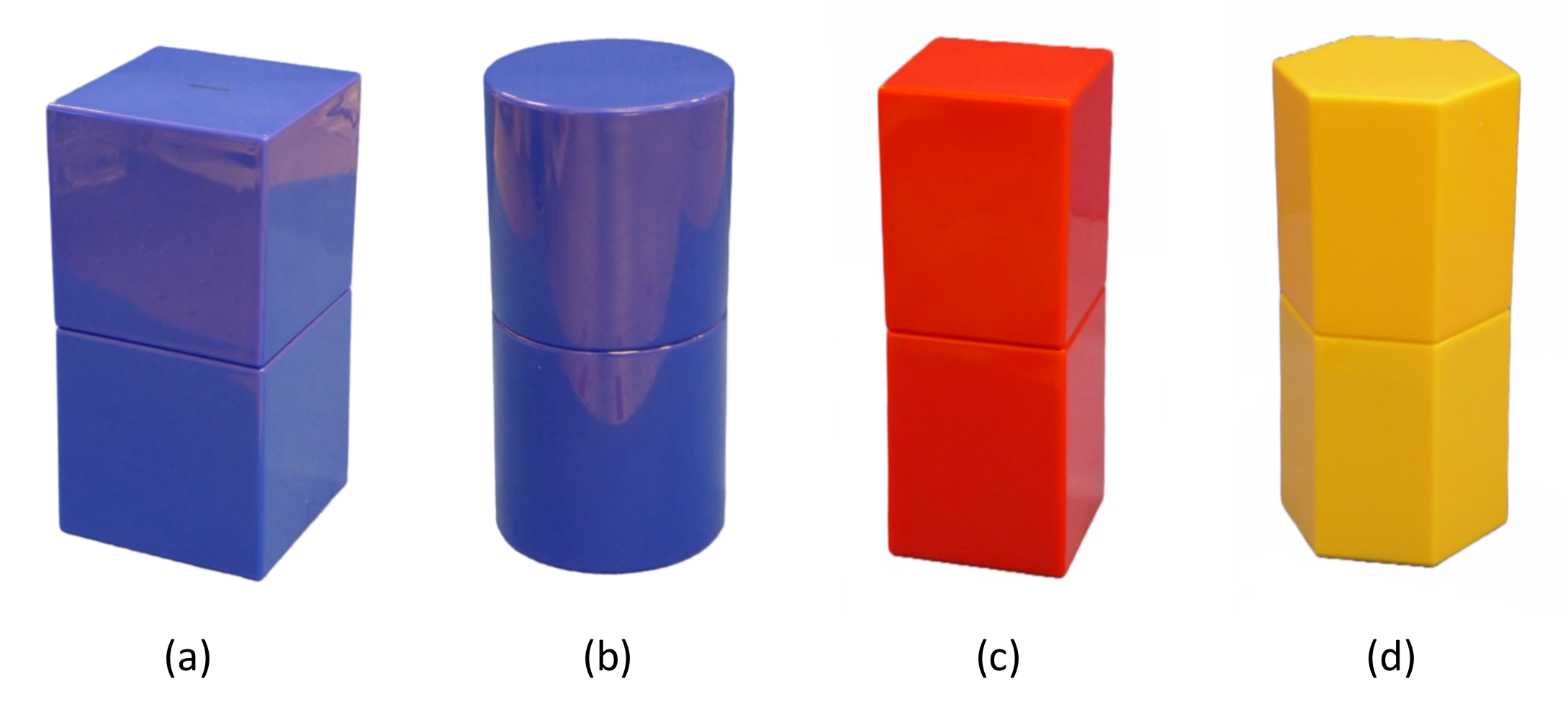}
	\caption{Tall (double height) geometric objects used in dual-arm object pushing experiments: (a) large blue square prism (566 g), (b) blue circular prism (436 g), (c) small red square prism (324 g), (d) yellow hexagonal prism (373 g).
	\label{fig:tall_geometric_pushing_objects}}
\end{figure}

\begin{figure}
	\centering
	\includegraphics[width=\columnwidth]{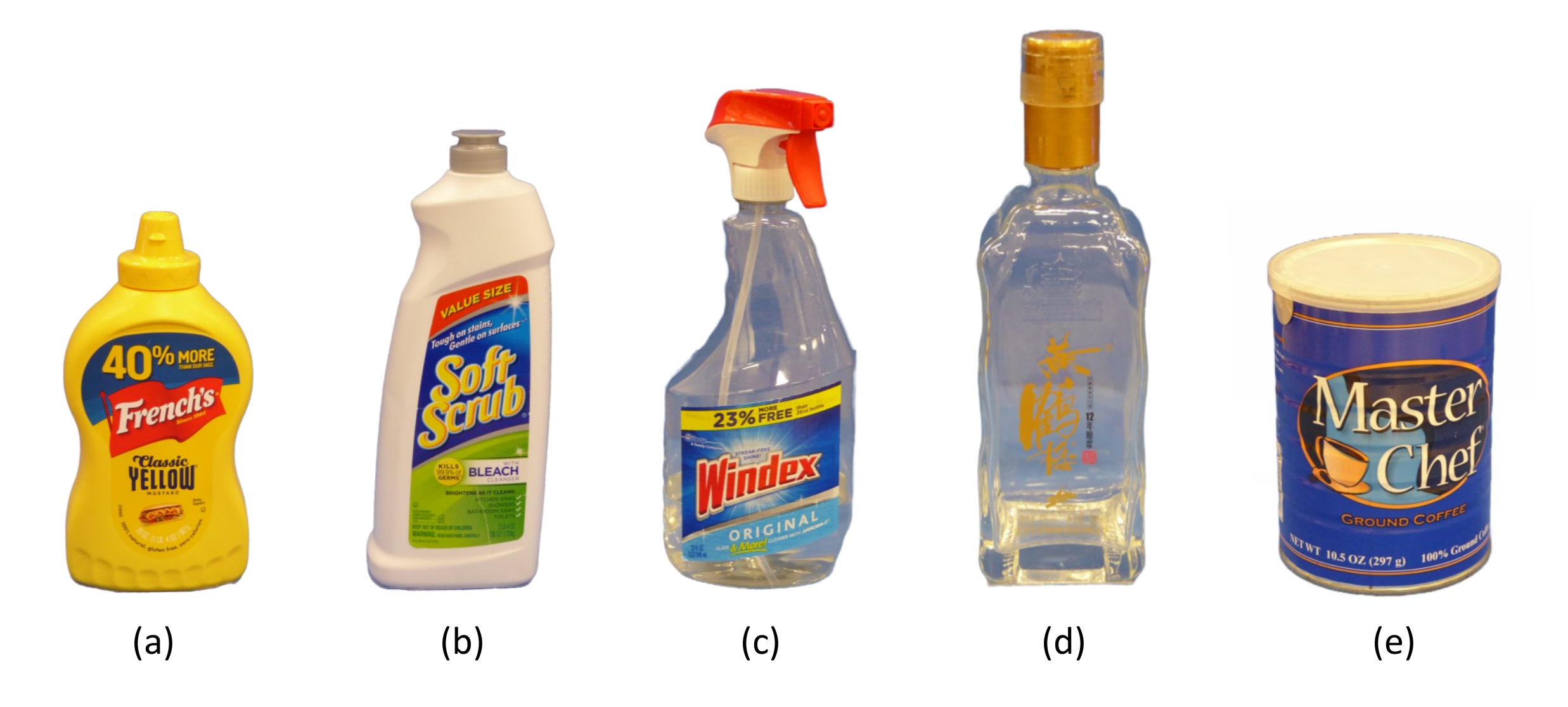}
	\caption{Tall everyday objects used in dual-arm object pushing experiments: (a) mustard bottle (237 g), (b) cream cleaner bottle (161 g), (c) Windex window cleaner spray bottle (339 g), (d) glass bottle (641 g), (e) large coffee tin (214 g).
	\label{fig:tall_everyday_pushing_objects}}
\end{figure}

\begin{figure*}
	\centering
	\includegraphics[width=\textwidth]{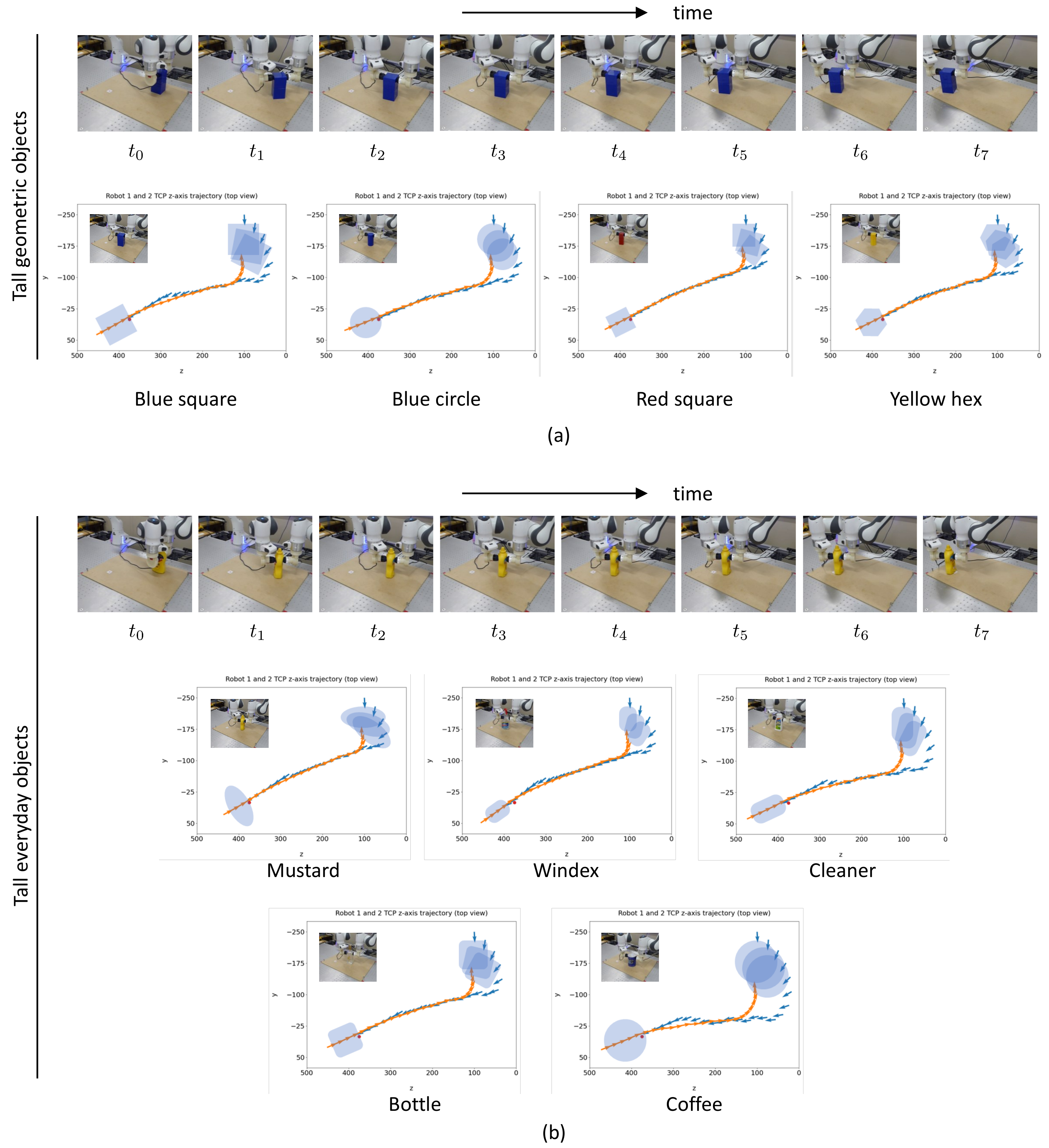}
	\caption{Using a leader and follower robot arm to push tall objects across an MDF surface: (a) tall geometric objects (time-lapse photos show sequence for tall blue square prism), and (b) tall everyday objects (time-lapse photos show sequence for mustard bottle). In the 2D pose plots, the target is identified by a small red circle.
	\label{fig:dual_arm_tall_object_pushing}}
\end{figure*}

In the second part of the dual-arm experiment, we used the two robot arms to push a set of taller (double-height) geometric objects (Figure~\ref{fig:tall_geometric_pushing_objects}) and tall everyday objects (Figure~\ref{fig:tall_everyday_pushing_objects}) across an MDF surface (Figure~\ref{fig:dual_arm_tall_object_pushing}).

For this part of the experiment, it was necessary to modify the (feedback) reference contact pose used in the pushing controller of the leader robot to $(0.5,0,0,0,0,0)$ , and the reference contact pose used in the servoing controller of the follower robot to $(-0.5,0,3,0,0,0)$. These modified poses only differ from the default ones listed in Table~\ref{tab:pushing_control_params} and Table~\ref{tab:stabiliser_control_params} in their first component. The effect of these changes is to apply a slight downward force on the pushed side of the object and slight upward force on the stabilised side of the object. This helps prevent these taller objects from catching their leading edges on the surface as they are being pushed. We repeated the experiment five times for each object, and computed the mean $\pm$ standard deviation target error across all five trials (Table~\ref{tab:dual_arm_pushing_tall_results}).

\begin{table}[h]
	\small\sf\centering
	\caption{Dual-arm pushing target error for tall objects (mean $\pm$ standard deviation perpendicular distance from target to sensor-object contact normal on completion of push sequence). All statistics are computed over 5 independent trials.
	\label{tab:dual_arm_pushing_tall_results}}
	\begin{tabular}{cc}
	\toprule
	Object & MDF surface\\
	\midrule
	Tall blue square & 3.62 $\pm$ 0.31 mm\\	
	Tall blue circle & 7.11 $\pm$ 2.18 mm\\	
	Tall red square & 4.94 $\pm$ 0.19 mm\\	
	Tall yellow hexagon & 4.73 $\pm$ 0.38 mm\\
	\midrule
	Mustard bottle & 6.24 $\pm$ 2.33 mm\\	
	Cleaner bottle & 7.02 $\pm$ 0.95 mm\\	
	Windex spray & 4.59 $\pm$ 1.36 mm\\	
	Glass bottle & 6.42 $\pm$ 0.39 mm\\
	Coffee tin & 4.81 $\pm$ 2.09 mm\\
	\bottomrule
	\end{tabular}
\end{table}

Once again, we visualised examples of the push sequences by plotting the end-effector poses of both robot arms in 2D and overlaid the approximate poses of the pushed objects at the start and finish points of the trajectory (Figure~\ref{fig:dual_arm_tall_object_pushing}).

The results in Table~\ref{tab:dual_arm_pushing_tall_results} and Figure~\ref{fig:dual_arm_tall_object_pushing} show that our dual-arm system can push the taller geometric and everyday objects over the MDF surface, approaching the target to within less than 7.5 mm for all objects.

\section{Summary and discussion}
\label{sec:discussion}

\subsection{Summary of results}
\label{sec:discussion_summary}

In this paper we have described, demonstrated and evaluated a tactile robotic system that uses contact pose and post-contact shear information for object tracking, surface following, and single-arm/dual-arm object pushing. Our new system is different from our previous work in two key ways: (a) it estimates and uses both contact pose \emph{and} post-contact shear, and (b) it allows smooth continuous control of the robot arm(s), as opposed to discrete, point-to-point motions. These features allow the robot arm to track objects in six degrees of freedom, either as a primary goal (e.g., as in object tracking) or as a secondary constraint on a primary goal (e.g., when moving an object along a desired trajectory while maintaining a given contact pose). In developing the underlying methods, we employed Lie group theory, specifically the use of tangent spaces, so that we could leverage established tools and techniques from probability and control theory that were originally developed for Euclidean vector spaces. This new perspective provides a more principled theoretical foundation than we have previous used.

\subsection{Successful estimation of contact pose and post-contact shear}
\label{sec:discussion_contact_pose_shear}

In Section~\ref{sec:nn_pose_shear_estimate}, we outlined a procedure for merging a normal contact pose with a post-contact shear motion to produce a unified surface contact pose. We showed that although multi-output regression CNNs are capable of estimating these surface contact poses from tactile sensor images, the estimates are not particularly accurate, even after hyperparameter tuning. To improve the accuracy, we modified the CNN architecture to estimate the uncertainty associated with its pose estimates, resulting in a model we refer to as a Gaussian density network (GDN). Our statistical evaluation of the GDN model showed that the average mean absolute error (MAE) for each pose component was lower than the corresponding MAE obtained for the CNN model, and a visual comparison of the estimation errors confirmed this improvement. Moreover, the GDN model allows us to employ a Bayesian filter to reduce noise and uncertainty in a sequence of estimates. In Section~\ref{sec:se3_bayes_filter}, we proposed a novel $SE(3)$ discriminative Bayesian filter that decreases the error and uncertainty of our GDN pose estimates. Our statistical evaluation confirmed that our filter improves the accuracy and reduces uncertainty, but the degree of improvement is contingent upon the accuracy of the state dynamics model used with the filter.

We believe that in this type of pose estimation task, most of the estimation error results from aliasing effects, whereby very similar sensor images become associated with very different contact pose labels. Specifically, when the sensor is sheared after lightly contacting a surface, it can glide across the surface and assume a comparable state to other contact trajectories, which may account for the higher errors observed for shear-related pose components. However, in practice, it is challenging, if not impossible, to prevent this type of aliasing. If we could prevent it, it would certainly reduce the complexity of the system since we could dispense with the Bayesian filter. Nonetheless, to accomplish this, we would need to restrict the contact depth to a very small range during data collection, and this would restrict the model's applicability to a limited range of contact types when it is deployed.

In this study, we concentrated on estimating contact poses with flat or gently curving surfaces. However, we believe it should be possible to expand this capability to estimate contact poses with other types of features, such as edges or corners. Examining the marker density plots in Figure~\ref{fig:marker_density_images}, we can see some unused degrees of freedom that could be used to represent other feature attributes. For instance, density variations within the low-density contact regions could represent edge orientation or edge offsets, enabling us to predict contact poses with respect to edge features. Clearly, if we did so, we would no longer be able to combine a contact pose having more than three degrees of freedom (DoF) with a post-contact shear with three DoF to form a single 6-DoF surface contact pose. However, we could still use a GDN model to simultaneously predict both the contact pose and the post-contact shear motion. In this scenario, we would need to combine the outputs of two feedforward-feedback pose controllers (one for the contact pose and one for post-contact shear). This extension for estimating contact poses with arbitrary features clarifies why using a tactile sensor for this type of problem is more potent than employing a 6-DoF force-torque sensor at the end of a robot arm. On the downside, incorporating more degrees of freedom in the model is likely to increase the estimation error and uncertainty due to aliasing and "feature entanglement" (see \cite{lepora2021pose}, and this would provide an even stronger requirement for some form of filtering.

\subsection{Smooth continuous motion using velocity-based control}
\label{sec:discussion_velocity_control}

In developing our new system, a primary objective was to achieve smooth and continuous motion of the robot arm using velocity-based control, in contrast to our previous position-based control system, which only gave the \emph{impression} of smooth motion through time-lapse videos. We achieved this goal by using a discrete-time controller that updates the velocity of the robot arm during each control cycle instead of its pose. Although this change has minimal impact on controller design for tasks such as object tracking and tactile servoing, it has some additional implications for object pushing. In our previous system (\cite{lloyd2021goal}), we steered the object towards its target by breaking contact with the object, moving around its perimeter and re-establishing contact to induce a turning moment on the object or to push it in a different direction. This action was possible due to the discrete, position-based movements. However, with the new velocity-based control system, it is challenging to implement such motion. Therefore, we modified the controller to maintain continuous frictional contact with the object and used tangential sideways motion to steer the object. Although both methods work well in their respective contexts, they do not behave in exactly the same way or have the same characteristics. For instance, the new velocity-based pushing controller relies to a much greater extent on a good initial contact position, as it cannot move around the perimeter of the object to find a better one. Thus, the new velocity-based pushing controller is in some ways less robust than the previous position-based one.

\subsection{Performance on servoing and manipulation tasks}
\label{sec:discussion_servo_manipulation_tasks}

We applied our robotic system to several tactile servoing and manipulation tasks, including some that we have previously demonstrated such as surface following and single-arm object pushing, and some new tasks such as object tracking and dual-arm object pushing. These new tasks were made possible by the system's ability to estimate post-contact shear motion. In Section~\ref{sec:obj_pose_tracking_exp}, we presented a successful demonstration of a follower robot arm tracking a flat surface attached to a leader robot arm. We also demonstrated a follower arm tracking various objects, such as a Rubik's cube, a mustard bottle, and a soft foam ball, while maintaining contact with the leader arm around a complex 3D trajectory. This task can be viewed as a form of 3D object manipulation, where one arm takes an active/leader role, and the other arm takes a passive/follower role. However, this is not the only possible configuration. In principle, there is no reason why the controllers of both arms could not be configured to that they follow an active reference signal, while at the same time maintaining a reference contact pose with an object.

In Section~\ref{sec:surface_follow_exp}, we demonstrated that a single robot arm can follow the surface of a curved ramp or a hemispherical dome while the sensor remains normal to the contacted surface at a fixed contact depth. To the best of our knowledge, this is the first time this task has been accomplished using smooth continuous motion in 3D. A potentially useful extension of this configuration might detect the amount of sensor shear experienced during the sliding action and engage a controller (e.g., a PID controller) to reduce the velocity until the sensor shear is returned to some safe maximum value, preventing the soft sensor from being torn by high-friction surfaces.

Section~\ref{sec:single_arm_object_pushing_exp} presented a successful demonstration of a single robot arm pushing several geometric objects to within approximately 7 mm of the target on a foam surface and 10 mm of the target on an MDF surface. However, these results are somewhat negatively skewed by the results achieved for the circular prism, since it was possible to push the other objects to within 2 mm of the target on the foam surface and 5 mm of the target on the MDF surface. Once again, to the best of our knowledge, this was the first time this operation has been achieved using smooth continuous motion. The current method has an advantage over our previous method as it detects shear in the vertical plane and adjusts the height of the sensor above the surface to maintain a desired reference value, preventing the sensor from colliding with the surface or drifting upwards when the physical surface is not quite parallel with the corresponding plane in the robot base frame.

In Section~\ref{sec:dual_arm_short_object_pushing_exp}, we demonstrated that two robot arms, working together in a leader-follower configuration, can push the geometric objects used in the single-arm pushing experiment to within approximately 5 mm of the target on both foam and MDF surfaces. In this case, using a second stabilizing arm helped to improve the performance, particularly when pushing the circular prism object. We also used this dual-arm configuration to push taller (double-height) geometric objects and tall everyday objects across an MDF surface. When we tried pushing these taller objects on the foam surface, the leading edges of the objects tended to catch, causing them to topple rather than move across the surface, so we abandoned using this type of surface for this part of the experiment. Even on the less compliant MDF surface, we needed to modify the vertical shear displacement on the leader and follower arms so that the objects were pushed slightly downwards on the leader side and lifted slightly upwards on the follower side. Once again, we should emphasise that this dual-arm capability has only been made possible by the system's ability to estimate post-contact tactile shear. In Section~\ref{sec:dual_arm_tall_object_pushing}, we demonstrated that we could use the dual-arm configuration to push the double-height geometric objects and everyday objects, such as tall bottles and containers, to within approximately 7 mm of the target.

\subsection{Limitations and further work}
\label{sec:discussion_further_work}

A limitation of the present system is the absence of a planning component, which hinders its ability to anticipate and prevent undesired situations, such as collisions between robot arms, or motion trajectories that approach or reach singularities. Implementing this planning capability would also be beneficial in scenarios where the system is unable to achieve a global task objective by following a local control objective, such as non-holonomic object pushing tasks where an object must be rotated to a target orientation as well as moved to a target position.

Furthermore, we would like to extend the system in a way that enables both robot arms to operate in an active configuration, where they are both functioning as leaders and followers to some degree. This would broaden their ability to collaborate on a more diverse set of tasks, particularly those related to more complex types of object manipulation.

Additional areas of future research encompass the use of a force/torque sensor to calibrate surface contacts using forces, torques, and post-frictional contact translational and rotational displacements, rather than based solely on poses. This approach would be useful in scenarios where control objectives or constraints are specified in such units.

\begin{acks}

This work was supported by an award from the Leverhulme Trust: “A Biomimetic Forebrain for Robot Touch” (RL-2016-39). The authors would also like to thank Andy Stinchcombe and
members of the Dexterous Robotics group at Bristol Robotics Laboratory.

\end{acks}

\bibliographystyle{SageH}
\bibliography{references.bib}

\begin{appendices}

\section{A numerically stable implementation of the softbound function}
\label{app:softbound implementation}

In Section~\ref{sec:gdn_model}, we defined the softbound function in terms of the softplus function, using:
\begin{multline}
	\label{eqn:softbound_def_app}
	\mathrm{softbound} \left( x \right)
	\, = \,
	x_{\mathrm{min}}
	+ \mathrm{softplus} \left( x - x_{\mathrm{min}} \right)\\
	- \mathrm{softplus} \left( x - x_{\mathrm{max}} \right)
\end{multline}
where
\begin{equation}
	\label{eqn:softplus_def_app}
	\mathrm{softplus} \left( x \right)
	\, = \,
	\ln \left( 1 + \exp \left( x \right) \right)
\end{equation}

In terms of numerical stability, there are two issues with this expression: firstly, the softplus function defined in Equation~\ref{eqn:softplus_def_app} must be implemented carefully to avoid round-off and overflow errors; and secondly, when $x_{\mathrm{min}} \ll 0$ and $x \gg x_{\mathrm{min}}$, round-off errors may occur when using a na\"ive implementation of Equation~\ref{eqn:softbound_def_app}.

The first of these problems is typically handled by libraries such as TensorFlow or PyTorch using a stable implementation of the softplus function such as:

\begin{equation}
	\label{eqn:softplus_stable_def_1}
	\mathrm{softplus} \left( x \right)
	\, = \,
	\begin{cases}
		x \text{, if $x >$ threshold},\\
		\ln \left( 1 + \exp \left( x \right) \right) \text{, otherwise.}\\
	\end{cases}
\end{equation}

Alternatively, it is possible to factor out the $\exp(x)$ term inside the $\ln(\cdot)$ term of Equation~\ref{eqn:softplus_def_app}, and use this expression to re-write the softplus function in a stabler form, as:
\begin{equation}
	\label{eqn:softplus_stable_def_2}
	\mathrm{softplus} \left( x \right)
	\, = \,
	\mathrm{max} \left(0, x \right) + \ln \left( 1 + \exp \left( - \left| x \right| \right) \right)
\end{equation}

Since the softplus function is a smooth approximation to $\mathrm{max} \left(0, x \right)$, the second term in this expression can be viewed as the approximation error. We can use this expression for the softplus function, to redefine the softbound function (Equation~\ref{eqn:softbound_def_app}) in a stabler form as:
\begin{multline}
	\label{eqn:softbound_stable_def}
	\mathrm{softbound} \left( x \right)\\
	\, = \,
	\mathrm{max} \left( x, x_{\mathrm{min}} \right)
	+ \mathrm{min} \left( x, x_{\mathrm{max}} \right)
	- x\\
	+ \ln \left( 1 + \exp \left( - \left| x - x_{\mathrm{min}} \right| \right) \right)\\
	- \ln \left( 1 + \exp \left( - \left| x - x_{\mathrm{max}} \right| \right) \right)
\end{multline}

Once again, since the softbound function is a smooth approximation to $\mathrm{max} \left( x, x_{\mathrm{min}} \right) + \mathrm{min} \left( x, x_{\mathrm{max}} \right) - x$, the remaining terms in this expression can be viewed as the approximation error. When implementing this function, we use a numerically stable implementation of $\ln \left( 1 + x \right)$, which is typically included in numerical libraries such as Numpy, TensorFlow and PyTorch.

\section{Probabilistic transformation and data fusion using linear-Gaussian models and Gaussian distributions}
\label{app:probabilistic_transform_fusion}

In this appendix, we derive expressions for probabilistic transformation using linear-Gaussian models, and probabilistic fusion of multivariate Gaussian random variables. As we did in the main text, we use lower-case, italic letters (e.g., $x$ or $y$) to represent continuous random variables, regardless of whether they are scalars, vectors, or $SE(3)$ objects, to avoid the notation becoming cumbersome.

\subsection{Probabilistic transformation using linear-Gaussian models}
\label{app:probabilistic_transform}

Starting with the general case of a PDF, $p(x)$, of a continuous random variable $x$, we express the probabilistic transformation of $x$ to another continuous random variable, $y$, using the conditional PDF $p(y|x)$. We then compute the PDF of $y$, $p(y)$, by marginalising the joint PDF $p(x,y)= p(y|x) \, p(x)$ over $x$: 
\begin{equation}
	\label{eqn:gen_probabilistic_transform}
	p(y)
	\, = \,
	\int p(y|x) \, p(x) \, dx
\end{equation}

In Bayesian filters, this type of probabilistic transformation is applied in a recursive manner during the prediction step. In this context, there are two desirable properties that a solution should possess: (a) $p(y)$ should have the same form of PDF as $p(x)$, so that the solution from one iteration can be substituted in the integrand in the following iteration; and (b) it should be computationally efficient to evaluate, preferably with a closed-form solution. The probabilistic transformations used in several popular Bayesian filters, including the Kalman filter (\cite{kalman1960new, kalman1961new}), extended Kalman filter (see \cite{gelb1974applied}) and unscented Kalman filter (\cite{julier1995new, julier1997new}), possess these two properties.

In general, the marginalisation integral in Equation~\ref{eqn:gen_probabilistic_transform} can be difficult and expensive to compute, and may require the use of Monte Carlo techniques, grid-based methods, or other types of simplifying approximation. However, in a linear-Gaussian model, the computation can be simplified by assuming that $x$ is a multivariate Gaussian random variable, which is linearly transformed and then summed with another multivariate Gaussian noise variable, $z$:
\begin{equation}
	\label{eqn:linear_gaussian_model}
	\begin{split}
		x & \sim \mathcal{N}(\mu_{x}, \Sigma_{x})\\
		z & \sim \mathcal{N}(\mu_{z}, \Sigma_{z})\\
		y & = Ax + z 
	\end{split}
\end{equation}

In this case, the integral in Equation~\ref{eqn:gen_probabilistic_transform} has a simple closed-form solution, which is also a Gaussian PDF (e.g., see Proposition 285, in Section 56.1.5 of \cite{taboga2012lectures}:
\begin{equation}
	\label{eqn:linear_gaussian_solution}
	\begin{split}
		y & \sim \mathcal{N}(\mu_{y}, \Sigma_{y})\\
		\mu_{y} & = A \, \mu_{x} + \mu_{z}\\
		\Sigma_{y} & = A \, \Sigma_{x} \, A^{\top} + \Sigma_{z}
	\end{split}
\end{equation}

It is relatively easy to extend this linear-Gaussian model to a non-linear function, by linearising the function about an appropriate operating point (e.g., see Section 2.2.7 in \cite{barfoot2017state}.

\subsection{Probabilistic fusion of Gaussian random variables}
\label{app:probabilistic_fusion}

Once again, we start with the general case for continuous random variables, before specialising to multivariate Gaussian PDFs.

We assume that we have two probabilistic estimates of a continuous random variable, $x$, which are based on two other variables, $y_{1}$ and $y_{2}$. For example, $x$ might represent the state of a robot, and $y_{1}$ and $y_{2}$ might represent two observations that we use to estimate the state. We represent the two probabilistic estimates using conditional PDFs, $p(x|y_{1})$ and $p(x|y_{2})$. We further assume that the variables $y_{1}$ and $y_{2}$ are conditionally independent given $x$. In other words, $p(y_{1}, y_{2} | x) = p(y_{1} | x) \, p(y_{2} | x)$. We can then use Bayes' rule to compute the conditional PDF of $x$, given both $y_{1}$ and $y_{2}$:
\begin{equation}
	\label{eqn:gen_probabilistic_fusion}
	\begin{split}
		p(x | y_{1}, y_{2})\\
		& \, = \,
		\frac{p(y_{1}, y_{2} | x) \, p(x)} {p(y_{1}, y_{2})}\\
		& \, = \,
		\frac{p(y_{1} | x) \, p(y_{2} | x) p(x)} {p(y_{1}, y_{2})}\\
		& \, = \,
		\frac{p(x | y_{1}) \, p(y_{1}) \, p(x | y_{2}) \, p(y_{2}) \, p(x)} {p(y_{1}, y_{2}) \, p(x)^{2}}\\
		& \, = \,
		\frac{1}{Z} \, \frac{p(x | y_{1}) \, p(x | y_{2})} {p(x)}
	\end{split}
\end{equation}

Here, we have used the conditional independence assumption in the second step, and a further application of Bayes' rule to compute $p(y_{1}|x)$ and $p(y_{2}|x)$ in the third step. The normalisation coefficient $Z$ is given by:
\begin{equation}
	\label{eqn:gen_probabilistic_fusion_normaliser}	
	Z
	\, = \,
	\int \frac{p(x | y_{1}) \, p(x | y_{2})} {p(x)} dx
	\, = \,
	\frac{p(y_{1}) \, p(y_{2})}{p(y_{1}, y_{2})}
\end{equation}

If $p(x)$ and $p(x|y_{i}), \; i=1,2$ are multivariate Gaussians with PDFs:
\begin{multline}
	p(x)
	\, = \,
	\frac {1} {\sqrt{ (2 \pi)^{M} \mathrm{det}( \Sigma_{x} ) } }\\
	\times
	\exp \left(
	-\frac{1}{2}
	(x - \mu_{x})^{\top}
	\Sigma_{x}
	(x - \mu_{x})
	\right)
\end{multline}
and
\begin{multline}
	p(x | y_{i})
	\, = \,
	\frac {1} {\sqrt{ (2 \pi)^{M} \mathrm{det}( \Sigma_{y_{i}} ) } }\\
	\times
	\exp \left(
	-\frac{1}{2}
	(x - \mu_{y_{i}})^{\top}
	\Sigma_{y_{i}}
	(x - \mu_{y_{i}})
	\right)
\end{multline}
then using a standard result for the normalised product of Gaussian PDFs (\cite{bromiley2003products, petersen2008matrix}) it can be shown that $p(x | y_{1}, y_{2})$, (Equation~\ref{eqn:gen_probabilistic_fusion}) is also Gaussian with PDF:
\begin{multline}
	\label{eqn:gauss_probabilistic_fusion}
	p(x | y_{i}, y_{2})
	=
	\frac {1} {\sqrt{ (2 \pi)^{M} \mathrm{det}( \Sigma^{\prime} ) } }\\
	\times
	\exp \left(
	-\frac{1}{2}
	(x - \mu^{\prime})^{\top}
	\Sigma^{\prime}
	(x - \mu^{\prime})
	\right)
\end{multline}
where the covariance and mean are given by:
\begin{equation}
	\label{eqn:gaussian_fusion_solution}
	\begin{gathered}
		\Sigma^{\prime} = \left( \Sigma_{y_{1}}^{-1} + \Sigma_{y_{2}}^{-1} - \Sigma_{x}^{-1} \right)^{-1}\\
		\mu^{\prime} = \Sigma^{\prime}
		\left(
		\Sigma_{y_{1}}^{-1} \mu_{y_{1}}
		+ \Sigma_{y_{2}}^{-1} \mu_{y_{2}}
		- \Sigma_{x}^{-1} \mu_{x}
		\right)
	\end{gathered}
\end{equation}

This result assumes that the expression used to calculate $\Sigma^{\prime}$ in Equation~\ref{eqn:gaussian_fusion_solution} produces a symmetric positive-definite covariance matrix.

If we also assume that $p(x)$ is approximately constant over its support (which would be the case if $p(x)$ is an uninformative, and possibly improper, prior), we can approximate $p(x | y_{1}, y_{2})$ as:
\begin{equation}
	\label{eqn:approx_probabilistic_fusion}
	p(x | y_{1}, y_{2})
	\, \approx \,
	\frac{1}{Z^{\prime}} \, p(x | y_{1}) \, p(x | y_{2})
\end{equation}
where $p(x)$ has been absorbed in the normalisation coefficient $Z^{\prime}$. This expression is easily recognizable as the normalised product of two conditional PDFs, which we use to perform probabilistic fusion in our Bayesian filter.

For multivariate Gaussian distributions, this normalised product of PDFs has the same form as Equation~\ref{eqn:gauss_probabilistic_fusion}, but with the covariance and mean now given by:
\begin{equation}
	\label{eqn:approx_gaussian_fusion_solution}
	\begin{gathered}
		\Sigma^{\prime} = \left( \Sigma_{y_{1}}^{-1} + \Sigma_{y_{2}}^{-1} \right)^{-1}\\
		\mu^{\prime} = \Sigma^{\prime}
		\left(
		\Sigma_{y_{1}}^{-1} \mu_{y_{1}}
		+ \Sigma_{y_{2}}^{-1} \mu_{y_{2}}
		\right)
	\end{gathered}
\end{equation}

Here, the mean, $\mu^{\prime}$, can be viewed as a normalised, inverse variance-weighted average of the two factor PDF means, $\mu_{y_{1}}$ and $\mu_{y_{2}}$. For the more general solution for $\mu^{\prime}$ given in Equation~\ref{eqn:gaussian_fusion_solution}, there is an additional negative term, $\Sigma_{x}^{-1} \mu_{x}$, which has the somewhat counter-intuitive effect of acting like a "repulsive force" on the solution for $\mu^{\prime}$ given in Equation~\ref{eqn:approx_gaussian_fusion_solution}.

As mentioned in the main body of the paper, if we assume a linear-Gaussian model with our more general Bayesian filter prediction and correction steps (Equation~\ref{eqn:bayes_filter_predict} together with Equation~\ref{eqn:bayes_filter_prob_fusion} or Equation~\ref{eqn:bayes_filter_prob_fusion_approx}), we obtain the prediction and correction equations for the Discriminative Kalman Filter (DKF) and robust DKF (\cite{burkhart2020discriminative}).

\section{Differential analysis of object pushing}
\label{app:pushing_differential_analysis}

\begin{figure}
	\centering
	\includegraphics[width=\columnwidth]{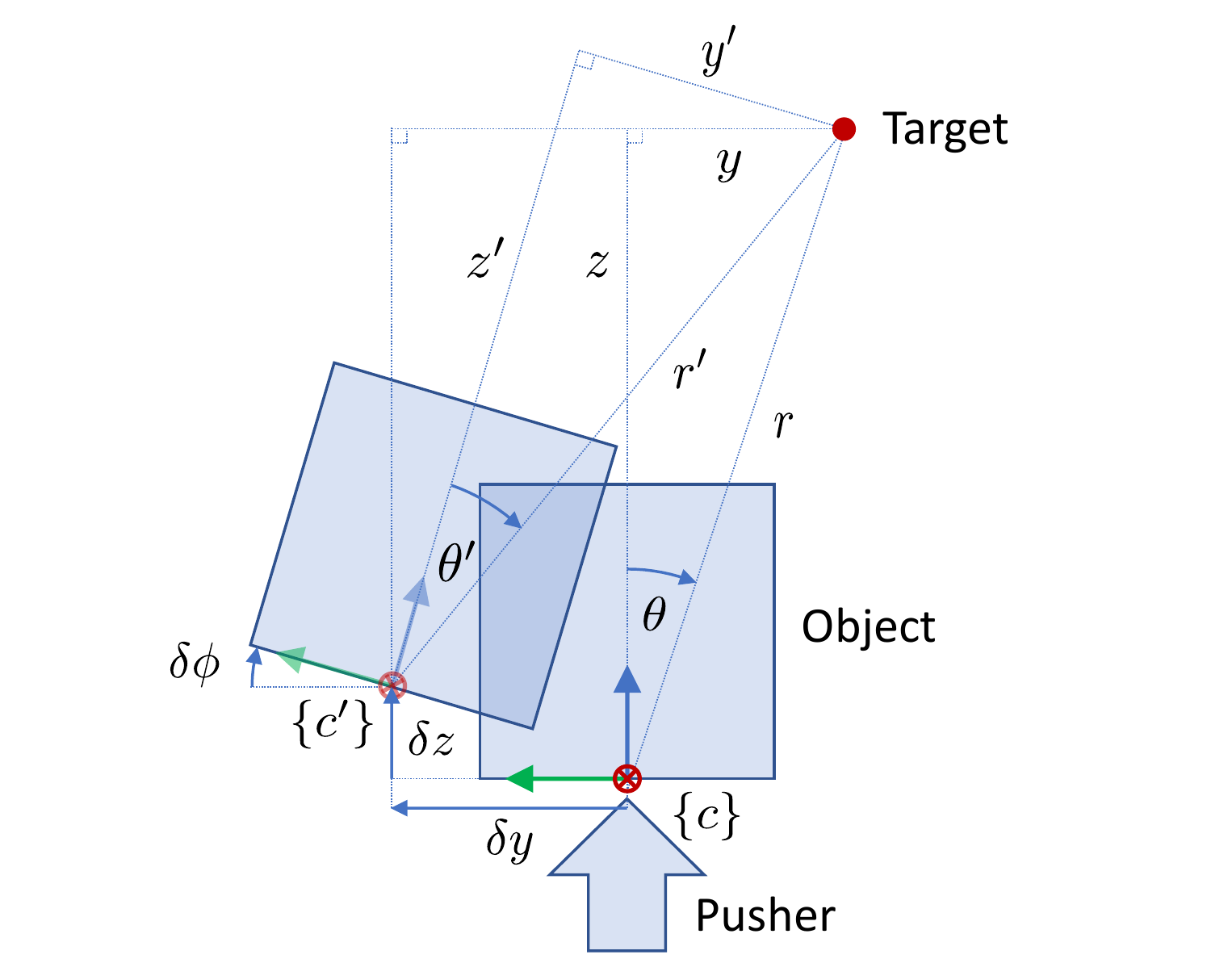}
	\caption{Differential analysis of object pushing. Note that, here, the target has a negative $y$-coordinate in the coordinate frame $\{ c \}$.
	\label{fig:pushing_differential_analysis}}
\end{figure}

In this appendix, we present a short differential analysis of object pushing, which is relevant to the scenarios covered in this paper. We begin by attaching a coordinate frame $\{ c \}$, to the pushed object at the point where it makes contact with the pusher (end-effector or sensor) (Figure~\ref{fig:pushing_differential_analysis}) so that it is normal to the contacting surfaces. In the figure, we use a square block to illustrate the situation, but the analysis also applies to objects with other shapes: it is the coordinate frames that are most relevant here.

In the coordinate frame, $\{ c \}$, the target bearing angle, $\theta$, can be computed as:
\begin{equation}
	\label{eqn:target_bearing_angle}
	\theta \, = \, \mathrm{atan2} \left( y, z \right) - \phi
\end{equation}
where $\phi$ specifies the rotation of the object in the $\{ c \}$-frame (initially, $\phi = 0$). During a small time interval, $\delta t$, the pusher moves the object so that in the instantaneous frame, $\{ c \}$, the frame is translated by $(\delta y, \delta z)$ and rotated by $\delta \phi$ to a new position $\{ c^{\prime} \}$ (the original variables are denoted using primed variables in the new frame). We can approximate the change in target bearing angle, $\delta \theta$, in the $\{ c \}$ frame using a first-order approximation:
\begin{equation}
	\label{eqn:first_order_approx}
	\delta\theta
	\, \approx \,
	\frac{\partial \theta}{\partial y} \delta y
	\, + \, \frac{\partial \theta}{\partial z} \delta z
	\, + \, \frac{\partial \theta}{\partial \phi} \delta \phi
\end{equation}
where:
\begin{equation}
	\label{eqn:partial_deriv_y}
	\begin{split}
		\frac{\partial \theta}{\partial y}
		\, = \,
		\frac{\partial} {\partial y} \left( \arctan \left( {\frac{y} {z}} \right) - \phi \right)
		\, = \,
		\frac{z}{y^2 + z^2}
		\, = \,
		\frac{z}{r^2}
	\end{split}
\end{equation}
\begin{equation}
	\label{eqn:partial_deriv_z}
	\begin{split}
		\frac{\partial \theta}{\partial z}
		\, = \,
		\frac{\partial} {\partial z} \left( \arctan \left( {\frac{y} {z}} \right) - \phi \right)
		\, = \,
		-\frac{y}{y^2 + z^2}
		\, = \,
		-\frac{y}{r^2}
	\end{split}
\end{equation}
\begin{equation}
	\label{eqn:partial_deriv_phi}
	\frac{\partial \theta}{\partial \phi}
	\, = \,
	\frac{\partial} {\partial \phi} \left( \arctan \left( {\frac{y} {z}} \right) - \phi \right)
	\, = \, -1
\end{equation}

Here, we have used the fact that, where these partial derivatives exist, $\mathrm{atan2}(y,z) = \arctan (y / z)$, except for a constant. Hence,
\begin{equation}
	\label{eqn:simplified_approx}
	\delta\theta
	\, \approx \,
	\frac{z}{r^2} \, \delta y
	\, - \, \frac{y}{r^2} \, \delta z
	\, - \, \delta \phi
\end{equation}

The angular change, $\delta \phi$, is independent of the \emph{position} of the target, $(y, z)$, but dependent on the \emph{change in position}, $(\delta y, \delta z)$. For example, in the simple case where the angular change, $\delta \phi$, is assumed proportional to the tangential motion, $\delta y$, of the pusher in the $y$-direction around the centre-of-friction (CoF), and the CoF does not move significantly in the $y$-direction, we might try and describe the relationship using an approximation of the form, $\delta \phi \approx \frac{1} {r_{0}} \delta y$. Here, $r_{0}$ is the distance from the CoF to the point of contact at the origin of $\{ c \}$. However, in reality, the CoF \emph{does} shift in the $y$-direction when a tangential motion occurs due to a frictional contact force, and hence the \emph{relative} tangential motion is somewhat reduced. Hence, a better approximation for $\delta \phi$ might be something of the form:
\begin{equation}
	\label{eqn:phi_approx}
	\delta \phi
	\, \approx \,
	\frac{\alpha} {r_{0}} \, \delta y
\end{equation}
where $0 < \alpha \leq 1$, depending on the shape and dimensions of the object, and its frictional properties with the surface it is pushed over. If we further assume that $\left| z \delta y \right| \gg \left| y \delta z \right|$ and $z \approx r$, which is true if the target bearing is small and $\delta y$ and $\delta z$ are of the same order of magnitude, then substituting in Equation~\ref{eqn:simplified_approx}, we obtain:
\begin{equation}
	\label{eqn:theta_approx}
	\delta \theta
	\, \approx \,
	\left( \frac{1} {r} - \frac{\alpha} {r_{0}} \right) \delta y
\end{equation}

Using this approximation, we can see that when the object is a long way from the target, $r$ is large and hence the first term inside the brackets will be negligible. In this situation, the change in bearing angle, $\delta \theta$, is roughly proportional to $\delta y$ and its sign is negative. However, as the object is pushed closer towards the target, the first term inside the brackets starts to play a greater role, effectively reducing the amount of rotation produced by $\delta y$ (i.e., it reduces the "effective gain" of the system). At some point, when $\frac{1} {r} = \frac{\alpha} {r_{0}}$, the rotation, $\delta \phi$ produced in response to a sideways motion, $\delta y$, becomes zero and then changes sign, effectively changing the polarity of the response in the dynamics model. This is like trying to steer a car toward a target, where the degree to which the vehicle responds as you turn the wheel decreases as you approach the target. Then at some point the car turns in the opposite direction to the one you expect it to go in.

Note that in carrying out this analysis we have \emph{not} assumed any particular form of controller. It applies equally to a simple feedback controller (e.g., PID), a linear-quadratic regulator, or a controller learned using reinforcement learning. The implication seems to be that, since the point at which this dynamic behaviour switching occurs depends on complicated properties of the object, surface and contact point, it is not possible to pre-define or learn a control policy that avoids this problem altogether unless the controller is also capable of (a) somehow detecting the point at which this happens for each type of object, and (b) responding appropriately (presumably, it would need to be able to detect which type of object it was pushing in order to apply the correct model).

In this work and previous work on pushing, we have used a very simple but crude method for mitigating the effects of this problem: we simply switch off the target alignment controller when we are within a pre-defined distance from the target. However, it raises an interesting question as to whether we can do better than this in future.

\end{appendices}

\end{document}